\documentclass[USenglish,onecolumn]{article}

\pdfoutput=1

\usepackage[utf8]{inputenc}
\usepackage[big]{dgruyter}
 
\usepackage[final]{microtype}
\usepackage[T1]{fontenc}
\usepackage{amsmath}
\usepackage{amsfonts}
\usepackage{amssymb}
\usepackage{amsthm}
\usepackage{mathtools}
\usepackage{boldline}
\usepackage{multirow,bigdelim}
\usepackage{colortbl}
\usepackage{adjustbox}
\usepackage{hyperref}
\usepackage{nameref}

\usepackage{comment}

\usepackage{bm}
\usepackage{enumitem}
\setlist[itemize]{itemsep=0mm}

\usepackage{caption}
\usepackage{float}

\usepackage[labelformat=simple]{subcaption}

\makeatletter
\def\Cline#1#2{\@Cline#1#2\@nil}
\def\@Cline#1-#2#3\@nil{%
  \omit
  \@multicnt#1%
  \advance\@multispan\m@ne
  \ifnum\@multicnt=\@ne\@firstofone{&\omit}\fi
  \@multicnt#2%
  \advance\@multicnt-#1%
  \advance\@multispan\@ne
  \leaders\hrule\@height#3\hfill
  \cr}
\makeatother

\makeatletter
\let\save@mathaccent\mathaccent
\newcommand*\if@single[3]{%
  \setbox0\hbox{${\mathaccent"0362{#1}}^H$}%
  \setbox2\hbox{${\mathaccent"0362{\kern0pt#1}}^H$}%
  \ifdim\ht0=\ht2 #3\else #2\fi
  }
\newcommand*\rel@kern[1]{\kern#1\dimexpr\macc@kerna}
\newcommand*\widebar[1]{\@ifnextchar^{{\wide@bar{#1}{0}}}{\wide@bar{#1}{1}}}
\newcommand*\wide@bar[2]{\if@single{#1}{\wide@bar@{#1}{#2}{1}}{\wide@bar@{#1}{#2}{2}}}
\newcommand*\wide@bar@[3]{%
  \begingroup
  \def\mathaccent##1##2{%
    \let\mathaccent\save@mathaccent
    \if#32 \let\macc@nucleus\first@char \fi
    \setbox\z@\hbox{$\macc@style{\macc@nucleus}_{}$}%
    \setbox\tw@\hbox{$\macc@style{\macc@nucleus}{}_{}$}%
    \dimen@\wd\tw@
    \advance\dimen@-\wd\z@
    \divide\dimen@ 3
    \@tempdima\wd\tw@
    \advance\@tempdima-\scriptspace
    \divide\@tempdima 10
    \advance\dimen@-\@tempdima
    \ifdim\dimen@>\z@ \dimen@0pt\fi
    \rel@kern{0.6}\kern-\dimen@
    \if#31
      \overline{\rel@kern{-0.6}\kern\dimen@\macc@nucleus\rel@kern{0.4}\kern\dimen@}%
      \advance\dimen@0.4\dimexpr\macc@kerna
      \let\final@kern#2%
      \ifdim\dimen@<\z@ \let\final@kern1\fi
      \if\final@kern1 \kern-\dimen@\fi
    \else
      \overline{\rel@kern{-0.6}\kern\dimen@#1}%
    \fi
  }%
  \macc@depth\@ne
  \let\math@bgroup\@empty \let\math@egroup\macc@set@skewchar
  \mathsurround\z@ \frozen@everymath{\mathgroup\macc@group\relax}%
  \macc@set@skewchar\relax
  \let\mathaccentV\macc@nested@a
  \if#31
    \macc@nested@a\relax111{#1}%
  \else
    \def\gobble@till@marker##1\endmarker{}%
    \futurelet\first@char\gobble@till@marker#1\endmarker
    \ifcat\noexpand\first@char A\else
      \def\first@char{}%
    \fi
    \macc@nested@a\relax111{\first@char}%
  \fi
  \endgroup
}
\makeatother

\usepackage{tikz}
\usepackage{tikz-cd}
\usetikzlibrary{arrows,arrows.meta,fit}
\usetikzlibrary{patterns}
\usetikzlibrary{matrix,decorations.markings,decorations.pathmorphing,decorations.pathreplacing,calc,positioning}

\tikzstyle{var}=[circle,draw=black,fill=white,thick,minimum size=20pt,inner sep=0pt]
\tikzstyle{exvar}=[circle,draw=black,fill=lightgray,thick,minimum size=20pt,inner sep=2pt]
\tikzstyle{ivar}=[rectangle,draw=black,fill=white,thick,minimum size=20pt,inner sep=0pt]
\tikzstyle{arr}=[->,>=stealth',draw=black,line width=1pt]
\tikzstyle{biarr}=[<->,>=stealth',draw=black,line width=1pt]

\tikzset{
  symbol/.style={
    draw=none,
    every to/.append style={
      edge node={node [sloped, allow upside down, auto=false]{$#1$}}}
  }
}

\newcommand\ellipsebyfoci[4]{
  let \p1=(#1), \p2=(#2), \p3=($(\p1)!.5!(\p2)$)
  in \pgfextra{
    \pgfmathsetmacro{\angle}{atan2(\y2-\y1,\x2-\x1)}
    \pgfmathsetmacro{\focal}{veclen(\x2-\x1,\y2-\y1)/2/1cm}
    \pgfmathsetmacro{\lentotcm}{\focal*2*#3}
    \pgfmathsetmacro{\axeone}{(\lentotcm - 2 * \focal)/2+\focal}
    \pgfmathsetmacro{\axetwo}{sqrt((\lentotcm/2)*(\lentotcm/2)-\focal*\focal}
  }
  (\p3) ellipse (\axeone cm and #4*\axetwo cm);
}

\newcommand{\gto}{\mathrel{\tikz{\draw[-{Straight Barb[angle'=90,scale=0.75]},line width=0.8](0,0)--(0.85em,0);}}}

\newcommand{\got}{\mathrel{\tikz{\draw[{Straight Barb[angle'=90,scale=0.75]}-,line width=0.8](0,0)--(0.8em,0);}}}

\newcommand{\goto}{\mathrel{\tikz{\draw[{Straight Barb[angle'=90,scale=0.75]}-{Straight Barb[angle'=90,scale=0.8]},line width=0.75](0,0)--(0.8em,0);}}}

\newcommand{\gtod}{\mathrel{\tikz{\draw[-{Straight Barb[angle'=90,scale=0.75]},dashed,dash pattern=on 1mm off 0.5mm,line width=0.8](0,0)--(0.8em,0);}}}

\newcommand{\gotd}{\mathrel{\scalebox{-1}[1]{\tikz{\draw[-{Straight Barb[angle'=90,scale=0.75]},dashed,dash pattern=on 1mm off 0.5mm,line width=0.8](0,0)--(0.8em,0);}}}}

\newcommand\Prb{\mathbb{P}}

\newcommand\given{\,|\,}

\newcommand\pa{\mathrm{pa}}

\newcommand\col{\mathrm{col}}

\newcommand\NN{\mathbb N}

\newcommand\RN{\mathbb R}

\newcommand\B[1]{\bm{#1}}
\newcommand\C[1]{\mathcal{#1}}  
\newcommand\BC[1]{\B{\C{#1}}}

\newcommand\intervene{\mathrm{do}}

\DeclareMathOperator*{\CI}{{\,\perp\mkern-12mu\perp\,}}

\DeclareMathOperator*{\SEP}{\perp}

\newcommand\CDM{SDCM}
\newcommand\CDMfull{structural dynamical causal model}
\newcommand\CDMFull{Structural dynamical causal model}

\newcommand\SDE{dynamic SE}
\newcommand\SDEfull{dynamic structural equation}

\newcommand{\Stephan}[1]{{\color{blue}{Stephan: #1}}}

\newcommand{\Joris}[1]{{\color{red}{Joris: #1}}}

\renewcommand{\Stephan}[1]{}

\newtheorem{theorem}{Theorem}[section]
\newtheorem{proposition}[theorem]{Proposition}
\newtheorem{corollary}[theorem]{Corollary}

\newtheorem{definition}[theorem]{Definition}

\newtheorem{lemma}[theorem]{Lemma}

\newtheorem{example}[theorem]{Example}

\newtheorem*{assumption*}{\assumptionnumber}
\providecommand{\assumptionnumber}{}
\makeatletter
\newenvironment{assumption}[2]
 {%
  \renewcommand{\assumptionnumber}{Assumption #1-$#2$}%
  \begin{assumption*}%
  \protected@edef\@currentlabel{#1}%
 }
 {%
  \end{assumption*}
 }
\makeatother
\newcommand{\asref}[2]{\ref{#1}-$#2$}

\begin{document}
 
\articletype{Research Article}

\author*[1]{S.\ Bongers}

\author[2]{T.\ Blom}

\author[3]{J.M.\ Mooij}

\affil[1]{Pattern Recognition Laboratory, Delft University of Technology; E-mail: s.r.bongers@tudelft.nl}

\affil[2]{Informatics Institute, University of Amsterdam; E-mail: t.blom2@uva.nl}

\affil[3]{Korteweg-de Vries Institute, University of Amsterdam; E-mail: j.m.mooij@uva.nl}

\title{\huge Causal Modeling of Dynamical Systems}

\runningtitle{Causal Modeling of Dynamical Systems}


\begin{abstract}
{Dynamical systems are widely used in science and engineering to model systems consisting of several interacting components.
Often, they can be given a \emph{causal} interpretation in the sense that they not only model the evolution of the states of the system's components over time, but also describe how their evolution is affected by external interventions on the system that perturb the dynamics.
We introduce the formal framework of \CDMfull s (\CDM s) that explicates the causal semantics of the system's components as part of the model. \CDM s represent a dynamical system as a collection of stochastic processes 
  and specify the basic causal mechanisms that govern the dynamics of each component as a structured system of random differential equations of arbitrary order.
  \CDM s extend the versatile causal modeling framework of structural causal models (SCMs), also known as structural equation models (SEMs), by explicitly allowing for time-dependence.
An \CDM\ can be thought of as the stochastic-process version of an SCM, where the static random variables of the SCM are replaced by dynamic stochastic processes and their derivatives. 
  We provide the foundations for a theory of \CDM s, by (i) formally defining \CDM s, their solutions, stochastic interventions, and a graphical representation; (ii) studying existence and uniqueness of the solutions for given initial conditions; (iii) providing Markov properties for \CDM s with initial conditions; (iv) discussing under which conditions \CDM s equilibrate to SCMs as time tends to infinity; (v) relating the properties of the \CDM\ to those of the equilibrium SCM. This correspondence enables one to leverage the wealth of statistical tools and discovery methods available for SCMs when studying the causal semantics of a large class of stochastic dynamical systems. 
The theory is illustrated with 
examples from different scientific domains.
}
\end{abstract}




\maketitle

\section{Introduction}
\label{sec:Intro}

Continuous dynamical systems consisting of differential equations are widely used in science and engineering to model the time-dependent behavior of certain phenomena.
A classical example is the modeling of the trajectory of a die that is thrown, by means of Newton's equations of motion.
Initial conditions or parameters of the dynamics may be stochastic, which can be modeled mathematically by making use of random differential equations (RDEs). These provide a natural extension of ordinary differential equations (ODEs) to the stochastic setting \citep{Bun72,Soo73,Sob90,NR13}.
For example, the initial position of the die is often not known, and varies from throw to throw, which leads to a probability distribution over the possible trajectories of the die (and eventually, to an uncertain outcome of the throw).

Many dynamical systems can be considered to consist of several interacting subsystems or components, for example, mass-spring systems in physics, predator-prey systems in biology, and mass-action law kinetics in chemistry.
These dynamical systems are often implicitly given a \emph{causal} interpretation in the sense that they are not only supposed to model the evolution of the state of the system over time, but also describe how the evolution of the system's components is affected by external interventions on the system that perturb the dynamics.
For example, when applying an external force to a particle, the change in the force term in Newton's second law of motion results in a changed acceleration, and hence a changed position, of the particle.
Another example is that hunting wolves may lead to an increase in the population of sheep.
The ensuing causal semantics of the system is usually only treated in an implicit and intuitive fashion, rather than that it is formally specified by (or derivable from) the mathematical model.
Indeed, a system of (random) differential equations simply expresses symmetric relations between the components, without any preferred order or asymmetry.
On the other hand, causal relations may be asymmetric, as they distinguish cause from effect.
Thus, while dynamical systems may describe how the state of a system consisting of several components evolves over time, by themselves they do not express the inherent ``causal structure'' of the system's components.

An apparently rather different modeling framework that allows to represent the causal semantics
of a system composed of components is provided by structural causal models (SCMs), 
also known as (non-parametric) structural equation models (SEMs)
\citep{Bol89,SGS00,PJS17,BPSM19}. First introduced in genetics by \citet{Wri21}, they became
popular over the years in econometrics \citep{Haa43}, the social sciences \citep{GD73, Dun75}, and more recently in AI \citep{Pea09}.
SCMs express causal relationships between variables corresponding to ``autonomous'' subsystems or components in the form of deterministic, functional relationships, 
and stochasticity is introduced through the assumption that certain variables are exogenous 
(latent) random variables. 
Their predictive power stems from the assumption that the equations of these models are organized in a structural way: each equation represents a distinct autonomous causal mechanism, where distinctness of the mechanisms means that they can be changed independently of one another by targeted interventions---at least in principle.
While SCMs explicate the causal semantics of a system composed of different components in this specific way, they have no built-in notion of time. 
A commonly used workaround for this limitation is to introduce multiple ``copies'' of the variables, corresponding to observations at different points (or intervals) in time.
This workaround only applies to discrete time, and SCMs cannot be used to model causal semantics of continuous-time systems without somehow discretizing time.

In this work, we propose the modeling framework of \emph{\CDMfull s} (\CDM s), which on the one hand explicates the causal relationships between components of continuous dynamical systems, and on the other hand extends structural causal models to explicitly allow for time-dependence.
\CDM s represent a dynamical system as a collection of stochastic processes (each one referring to a causally ``autonomous'' component) subject to a ``structured'' dynamics, which specifies the causal mechanisms that govern the dynamics of the components by means of random differential equations of arbitrary order.
An \CDM\ can be thought of as the stochastic-process version of an SCM, where the static (time-independent) random variables of the SCM are replaced by dynamic (time-dependent) stochastic processes and their derivatives. 
Our contributions can be considered as the first steps towards a theory of \CDM s. More specifically, we:
\begin{enumerate}[label=(\roman*)]
  \item formally define \CDM s, their solutions, stochastic interventions, and a graphical representation;
  \item study existence and uniqueness of the solutions for given initial conditions; 
  \item provide Markov properties for \CDM s with initial conditions;
  \item discuss under which conditions \CDM s equilibrate to SCMs as time tends to infinity;
  \item relate the properties of the \CDM\ to those of the equilibrium SCM.
\end{enumerate}
This correspondence between SCMs and equilibrated \CDM s enables one to leverage the wealth of statistical tools and discovery methods available for SCMs when studying the causal semantics of a large class of stochastic dynamical systems.  
We illustrate the theory with several well-known examples from different scientific domains.



\paragraph*{Related work}
Over the years, several efforts have been made to develop a notion of causality for stochastic processes, both in discrete and continuous time. 

For discrete time, Granger causality~\citep{Gra69,Whi06,Eic07,ED07}, simultaneous equation models~\citep{Fis70,LSRH08}, vector autoregressive (VAR) models~\citep{Sim80,Lut05} and dynamic Bayesian networks~\citep{DGH92,Gha98} have been studied extensively. More recently, there has been some work on learning difference-based causal models~\citep{VDD10} and structural equation models~\citep{PJS13}. In principle, all these models fit directly into the framework of SCMs by labeling the random variables with time.

For continuous time, there has been substantial work in the graphical modeling community \citep{Aal87,Did00,Did07,Did08,Did15} based on the concept of local independence, which was introduced by~\citet{Sch70}. However, none of these approaches explicitly takes into account that dynamical models are often based on differential equations. 
In parallel, several attempts have been made to arrive at causal interpretations of processes described by ordinary and stochastic differential equations. 
Many of these approaches start from the assumption of a first-order system of ODEs written in canonical form, and implicitly (or explicitly) attribute a causal interpretation to this \citep{IS94,MJS13,PBP19,BM21}.
The notion of causality in ODEs has also been studied using Simon's causal ordering algorithm \citep{IS94}. 
Relations between a certain class of causally interpreted ODEs and deterministic SCMs at equilibrium have been established under the strong assumption that all the solutions of the ODE converge to a single static equilibrium state \citep{MJS13}, independent of the initial condition. 
This assumption can be relaxed to allow for asymptotic dynamics \citep{RBSM18} such as periodic oscillations, but this still requires the assumption that the asymptotic dynamics does not depend on the initial condition.
Another way to relax the assumption of \citep{MJS13} is taken in the framework of causal constraints models \citep{BBM19}, which can model static equilibrium states as long as the dynamical system has a unique static equilibrium state corresponding to each initial condition, for every intervention. These models can give a more complete causal description of these static equilibrium states than SCMs can \citep{BBM19}, but this comes at the cost that they appear to be too ``flexible'' in general. Finally, several approaches in terms of stochastic differential equations, which are differential equations with an additive white noise term, have been developed over the years \citep{FF96,CG09,HS14,MMH18,PBP20}. The stochastic differential equations have the advantage that they can deal with ``instantaneous'' stochasticity in the dynamics, but solving them usually requires a considerable mathematical effort using It{\^o} calculus.

Compared with existing work, the framework of \CDMfull s that we propose here has the novel combination of features that it extends the semantics of continuous dynamical systems by formally encoding the causal structure into the model, it allows for stochasticity due to uncertainty over initial conditions or parameters of the dynamics without relying on strong stability assumptions, and it does not force one to consider time derivatives of processes as being ``causally independent'' of the processes themselves (that is, time derivatives of processes are considered to describe the same subsystem or component as the process itself).
Our framework reconciles the traditional intuitive treatment of causality in the context of deterministic dynamical systems as practiced in many exact sciences with the treatment of causality of stochastic systems that is nowadays very popular in AI, statistics and other scientific disciplines. An attractive feature is that it naturally accommodates many causally interpreted continuous dynamical systems that appear ``in the wild''~\citep[see e.g.,][]{FBKK16,NR13}.

\paragraph*{Contributions}

In this paper, we introduce the framework of \emph{\CDMfull s} (\CDM s),\footnote{Not to be confused with the \emph{dynamic causal models} of \citep{FHP03} or the \emph{dynamic structural causal models} of \citep{RBSM18}. The dynamic causal models of \citep{FHP03} have been developed to infer the causal relations between the activities of different brain regions, where each neuronal state is modeled by a first order differential equation. These much more restricted models could in principle be represented by \CDM s. The dynamic structural causal models of \citep{RBSM18} have been developed to model the asymptotic
behavior of an ordinary differential equation under non-constant interventions and assume that the asymptotic behavior does not depend on the initial condition.} 
which allows to model the causal semantics of stochastic processes for a large class of continuous dynamical systems by means of a ``structured'' system of random differential equations of arbitrary order (including zeroth-order). One can consider SCMs as special cases of \CDM s that only contain zeroth-order equations. The proposed modeling framework enables modeling of stochasticity, time-dependence and causality in a natural way. We study the existence and uniqueness of solutions of \CDM s, and propose a convenient graphical representation of the model structure for which we derive Markov properties. We define an idealized notion of stochastic interventions, and show that this yields a natural ``interventionist'' causal interpretation of the graph of an \CDM . We define a notion of equilibration of an SDCM to an SCM, which corresponds with letting a system converge towards equilibrium as time tends to infinity, and relate the properties of the \CDM\ to those of the equilibrium SCM. In the next paragraphs, we describe our contributions in more detail.

Intuitively, an \CDM\ can be thought of as an SCM where the notion of time is added to the structural equations by replacing the random variables of the SCM by stochastic processes and their (higher-order) derivatives. In the presence of these derivative processes, these equations, which we coin \emph{\SDEfull s}, can be read as random differential equations. The \SDEfull s have the property that they are organized in a structural way, similar to how the structural equations of an SCM are organized by associating a distinct causal mechanism to each observed variable. This distinguishes \CDM s from other ``non-causal'' (random) dynamical systems, and allows to define idealized stochastic interventions on these models, similarly to how this is usually done for SCMs.
The structure of the SCM can be expressed by its graph, which reflects the functional relationships between the components as encoded by the structural equations. Similarly, we define the graph of an \CDM\ to reflect the functional relationships between the components as encoded by the \SDEfull s.


The framework of \CDM s on the one hand allows one to specify the causal semantics of a system of RDEs, and on the other hand it enables temporal extensions for SCMs. In particular, we show when and how we can equilibrate an \CDM\ to an SCM, such that the static solutions of the SCM contain the equilibrium states of the \CDM. Our equilibration operation, inspired by the one of \citet{MJS13}, has the key property that it preserves the structure of the endogenous processes. Intuitively, the idea is that in the limit as time tends to infinity, the \SDEfull s converge to those equations for which the higher-order derivatives of the processes have been set to zero, yielding the structural equations of an SCM. This allows us to use SCMs to model the equilibrium states of dynamical systems, including cases that were previously considered to fall outside their scope, such as the price, supply and demand model in econometrics. In addition, we show that this equilibration operation commutes with intervention (as in Figure~\ref{fig:CommutingDiagramCDMSCM}), and naturally maps the graph of the \CDM\ to the graph of the SCM. This provides a different perspective on what \citet{Das05} calls the ``violation of the equilibration-manipulation commutability property''. 
Our formalism allows us to generalize the main result of \citet{MJS13}, which states that certain causally interpreted systems of ODEs can be equilibrated to SCMs, in several directions: (i) we replace the deterministic setting with a more general stochastic setting, that is, we can deal with randomness in the initial conditions and in the parameters, (ii) we allow the order of the equations of the dynamical model to be arbitrary, including zeroth-order, rather than restricting to first-order differential equations only, and (iii) we drop the strong assumption that the dynamical model needs to have a single static equilibrium that is independent of the initial condition. 

\begin{figure}
\begin{center}
\adjustbox{scale=0.90,center}{%
\begin{tikzcd}[/tikz/cells/.append style={nodes={draw, rounded corners}}, every label/.append style = {font = \small}] 
  \hyperref[def:SteadyCDM]{\begin{minipage}[c][3.4em][c]{4.8cm}\centering$\begin{array}{c} \text{steady \CDM} \\ \C{R} \end{array}$\end{minipage}} \arrow[dd,|->,shorten >=0.1cm,shorten <=0.1cm,line width=0.2mm,every label/.append style={xshift=0.2ex},"\text{(Def.~\ref{def:InterventionsCDM}})","{\text{\small $\intervene(I,\B{K}_I)$}}"'] \arrow[rrr,|->,shorten >=0.1cm,shorten <=0.1cm,line width=0.2mm,every label/.append style={yshift=-0.0ex},"{\text{\small $t \rightarrow \infty$}}","\text{(Def.~\ref{def:EquilibrationCDM})}"'] & & & \hyperref[def:SCM]{\begin{minipage}[c][3.4em][c]{4.8cm}\centering$\begin{array}{c} \text{SCM} \\ \C{M}_{\C{R}} \end{array}$\end{minipage}} \arrow[dd,|->,shorten >=0.1cm,shorten <=0.1cm,line width=0.2mm,every label/.append style={xshift=0.2ex},"\text{(Def.~\ref{def:InterventionsCDM})}","{\text{\small $\intervene(I,\B{K}_I^*)$}}"'] \\
    & & & \\
  \hyperref[def:InterventionsCDM]{\begin{minipage}[c][3.4em][c]{4.8cm}\centering$\begin{array}{c} \text{intervened steady \CDM} \\ \C{R}_{\intervene(I,\B{K}_I)} \end{array}$\end{minipage}} \arrow[rrr,|->,shorten >=0.1cm,shorten <=0.1cm,line width=0.2mm,every label/.append style={yshift=-0.0ex},"{\text{\small $t \rightarrow \infty$}}","\text{(Def.~\ref{def:EquilibrationCDM})}"'] & & & \hyperref[thm:InterventionCommutesWithEquilibrationCDMSCM]{\begin{minipage}[c][3.4em][c]{4.8cm}\centering$\begin{array}{c} \text{intervened SCM} \\ (\C{M}_{\C{R}})_{\intervene(I,\B{K}_I^*)} = 
\C{M}_{(\C{R}_{\intervene(I,\B{K}_I)})} \end{array}$\end{minipage}} 
  \end{tikzcd}
}
\end{center}
\vspace{-1\baselineskip}
\caption{This diagram shows that, under certain convergence assumptions,
  equilibration (left-to-right in the diagram) commutes with intervention (top-to-bottom in the diagram). The precise statement is made explicit in Theorem~\ref{thm:InterventionCommutesWithEquilibrationCDMSCM}.}
  \label{fig:CommutingDiagramCDMSCM}
\end{figure}

By no longer restricting to first-order dynamical systems, we arrive at a more natural causal interpretation of systems of higher-order RDEs, like the coupled harmonic oscillator. Thereby, we circumvent questions like “does position cause velocity, or does velocity cause position, or both?”. However, allowing for zeroth-order \SDEfull s leads to additional technical challenges that are absent when solving first-order RDEs. Indeed, the initial conditions of the solutions may be constrained by the zeroth-order \SDEfull s, and possibly even by additional ``hidden'' constraints. 
We provide sufficient conditions under which the existence and uniqueness of a solution of an \CDM\ with a given initial condition can be guaranteed. We also provide stronger conditions under which this still holds after certain interventions.

The existence and uniqueness of solutions of an \CDM\ are of key importance for obtaining Markov properties for \CDM s. 
By building on a powerful Markov property for SCMs \citep{FM17,BPSM19}, we derive a Markov property for \CDM s with initial conditions,
which enables one to read off (conditional) independencies between the stochastic processes that are solutions of the \CDM , provided the latter are uniquely defined. 
With a small extension, it can also be applied to the evaluation of the solutions at some specific point in time. 

Even if the existence and uniqueness of a solution of an \CDM\ can be guaranteed, not all solutions of an \CDM\ equilibrate, in general. For example, a coupled harmonic oscillator may oscillate indefinitely in the absence of friction. Moreover, the solutions that equilibrate may not always equilibrate to the same equilibrium state. For example, a freely moving particle subject to friction may end up anywhere, depending on its initial position and velocity. In other words, equilibrium states may depend on the initial condition. This is compatible with the recently proposed framework of cyclic SCMs of \citet{BPSM19}, which allows for the absence of (or, the presence of multiple) solutions of the structural equations. The intricate connection between the dependence of the equilibrium states of an \CDM\ on the initial conditions and the solvability properties of the equilibrated SCM sheds new light on the counterintuitive ``nonancestral'' causal effects in certain ``pathological'' cyclic SCMs with self-cycles that were first observed by \citet{Nea00}.

The scope of this paper is limited to establishing the framework of \CDM s and its bridge to SCMs at equilibrium. The importance of this bridge is that, although \CDM s can be used for modeling causal relationships between stochastic processes, inferring such causal models from data may pose certain difficulties. One significant practical drawback of using \CDM s for modeling systems with an unknown dynamics is that obtaining time series data with sufficiently high temporal resolution can be costly, impractical or even impossible.\footnote{For example, modern measurement techniques in biology, like RNA sequencing and mass cytometry, enable simultaneous measurements of multiple variables at once in single cells, but at the cost of destroying the cells during the measurement process. This means that it is impossible to obtain time-series measurements for individual cells, although one can take a ``snapshot'' of the internal states of many single cells at the same point in time.} The results of this work enable one to study the causal semantics of the equilibrium states of a large class of random dynamical models in terms of SCMs. In particular, this allows to infer properties of these dynamical models by employing the statistical tools and discovery methods available for static SCMs on equilibrium data. 

\paragraph*{Outline}
The paper is organized as follows: In Section~\ref{sec:Preliminaries}, we provide the necessary concepts of stochastic processes and random differential equations. In Section~\ref{sec:CDMs}, we introduce the class of \CDMfull s, define SCMs as special cases of \CDM s, define interventions, define the graph of an \CDM, discuss initial conditions, study existence and uniqueness of solutions, and derive a Markov property for \CDM s.
In Section~\ref{sec:EquilibrationOfCDMs}, we define the equilibration operation on steady \CDM s, define the graph of the equilibrated \CDM, describe the commutation of the intervention and the equilibration operation, 
study the inverse problem of finding steady \CDM s with non-trivial dynamics for which all the solutions equilibrate to solutions of the SCM,
and discuss subtleties in the causal interpretation of the graph of the equilibrated \CDM. 
We conclude with a discussion and some open problems in Section~\ref{sec:Discussion}. 
Proofs are provided in Appendix~\ref{app:Proofs}.

\section{Preliminaries}
\label{sec:Preliminaries}

We start off by defining some basic notation and terminology.

\subsection{Stochastic processes}\label{sec:StochasticProcesses}

In this subsection, we introduce the basic definitions and terminology for stochastic processes~\citep[see also][]{Bun72,NR13}. A \emph{stochastic process} is an $\RN^n$-valued function
$\B{X}:T\times\Omega\to\RN^n$, where $T$ is some index set, such that $\B{X}_t$ (which denotes $\B{X}(t,.)$, also sometimes denoted as $\B{X}(t)$) is for each $t\in T$ a random variable\footnote{Assuming the Borel $\sigma$-algebra $\mathfrak{B}(\RN^n)$ on $\RN^n$, that is, the smallest $\sigma$-algebra on $\RN^n$ that contains all open $n$-balls.} on a probability space $(\Omega,\C{F},\Prb)$. A random variable $\B{X}:\Omega\to\RN^n$ can itself be seen as a stochastic process that is constant in time, that is, as the process $\B{X}:T\times\Omega\to\RN^n$ defined by $\B{X}_t(\omega):=\B{X}(\omega)$. We always assume that there exists some background probability space $(\Omega,\C{F},\Prb)$ on which all random variables and processes are defined. Furthermore, we only consider processes where $T=[t_0,t_1]$ or $T=[t_0,\infty)$ for $t_0<t_1$ with $t_0,t_1\in\RN$, and the points of $T$ are thought of as representing \emph{time}. 
For each $\omega\in\Omega$ we have an $\RN^n$-valued function $T\to\RN^n$ mapping $t$ to $\B{X}_t(\omega)$, which is called a \emph{sample path}, or just a \emph{path}, of $\B{X}$. We call two stochastic processes $\B{X}$ and $\B{Y}$ \emph{a.s.\ equal} to each other, denoted by $\B{X} = \B{Y}$ a.s., if $\Prb$-almost surely all sample paths are equal, that is, if there exists a $\Prb$-null set\footnote{Let $(\Omega,\C{F},\Prb)$ be a probability space. A set $N\subseteq\Omega$ is called a \emph{$\Prb$-null set} if there exists a measurable set $\tilde{N}\in\C{F}$ with $N\subseteq \tilde{N}$ and $\Prb(\tilde{N})=0$.} $N\subseteq\Omega$ such that for all $\omega\in\Omega\setminus N$ and for all $t\in T$ we have $\B{X}_t(\omega) = \B{Y}_t(\omega)$. We consider stochastic processes, and random variables in particular, only up to a.s.\ equality.

A family $(\B{X}_i)_{i\in\C{I}}$ of stochastic processes for some finite index set $\C{I}$ is called \emph{independent} if for all $k\in\NN$ and all $k$-tuples $(t_1,\dots, t_k)$ of distinct elements of $T$ the family 
$$
(\tilde{\B{X}}_i)_{i\in\C{I}}
$$
of random variables $\tilde{\B{X}}_i:=((\B{X}_i)_{t_1}, \dots,(\B{X}_i)_{t_k})$ is independent.

We call a stochastic process $\B{X}$ \emph{continuous}, if its paths are continuous almost surely, that is, for $\Prb$-almost every $\omega\in\Omega$ and for all $t\in T$ we have
$$
\lim_{s\rightarrow t}\B{X}_s(\omega) = \B{X}_s(\omega) \,.
$$
We call a stochastic process $\B{X}$ \emph{differentiable}, if its paths are differentiable almost surely, that is, for $\Prb$-almost every $\omega\in\Omega$ and for all $t\in T$ the derivative
$$
  \B{X}'_t(\omega) := \frac{d\B{X}_t}{dt}(\omega) := \lim_{h\rightarrow 0}
\frac{\B{X}_{t+h}(\omega)-\B{X}_t(\omega)}{h}
$$
exists. The mapping $\B{X}':T\times\Omega\to\RN^n$ defines a stochastic process and is called the \emph{derivative} of $\B{X}$. Similarly, one can define, if it exists, the
\emph{$n^{\text{th}}$-order derivative} of $\B{X}$ as the derivative of the $(n-1)^{\text{th}}$-order derivative of $\B{X}$, which
we also write as $\B{X}^{(n)}$, where the zeroth-order derivative of $\B{X}$ is $\B{X}^{(0)}:=\B{X}$. 
We call a stochastic process $\B{X}$ \emph{continuously differentiable} or a \emph{$C^1$-stochastic process}, if its derivative $\B{X}'$ exists and is continuous. Similarly, we call $\B{X}$ a \emph{$C^m$-stochastic process}, if its derivatives $\B{X}'$, $\B{X}''$, \dots, $\B{X}^{(m)}$ exist and are continuous. In particular, $\B{X}$ is a $C^0$-stochastic process if it is continuous.

Consider a compact interval $T = [t_0, t_1] \subseteq \RN$.
The space $\C{C}^{m}(T,\RN^n)$ of $m$ times continuously differentiable functions $T \to \RN^n$, equipped with the $C^{m}$-norm 
$$\| \B{X} \|^{(m)} := \sum_{k=0}^{m} \sup_{t\in T} \|\B{X}^{(k)}(t)\|$$
 (where $\|\cdot\|$ is the Euclidean norm in $\RN^n$) is a Polish space, and with its Borel $\sigma$-algebra forms a standard measurable space~\citep{Kec95}.
A $C^{m}$-stochastic process $\B{X} : T \times \Omega \to \RN^n$ can also be seen as a random variable taking values in $\C{C}^{m}(T,\RN^n)$~\citep{Bor13}.
The following functionals (integration, differentiation and evaluation) are continuous, and hence measurable:
\begin{align*}
  \iota & : \RN^n \times \C{C}^{m}(T,\RN^n) \to \C{C}^{m+1}(T,\RN^n) : (\B{X}_{[0]},\B{X}) \mapsto \left(t \mapsto \B{X}_{[0]} + \int_{t_0}^t \B{X}(s)\,ds\right) \\
  \partial & : \C{C}^{m+1}(T,\RN^n) \to \C{C}^{m}(T,\RN^n) : \B{X} \mapsto \left(t \mapsto \B{X}'(t)\right) \\
  \pi & : \C{C}^m(T,\RN^n) \to \RN^n : \B{X} \mapsto \B{X}(t_1)\,.
\end{align*}
Furthermore, if we compose a process $\B{X} \in \C{C}^n(T,\RN^n)$ with a continuous function $\B{f} : \RN^n \to \RN^k$, we obtain a process $\B{f}(\B{X}) \in \C{C}^0(T,\RN^k)$.

\subsection{Clustered mixed graphs}

In this subsection, we introduce some graphical notions.

\label{sec:GraphTheory}
A \emph{mixed graph} is a pair $\C{G} = (\C{V},\C{E})$, where $\C{V}$ is a set of nodes and $\C{E}$ is a set of edges between the nodes of different types, in our case, $\gto,\got,\goto,\gtod,\gotd$. 
If $i \gto j$ or $i \gtod j$ in $\C{G}$, we call $i$ a \emph{parent of $j$} and denote with $\pa_{\C{G}}(j)$ the set of parents of $j$ (which may include $j$ itself in case $j \gtod j$ in $\C{G}$).
A mixed graph $\tilde{\C{G}}=(\tilde{\C{V}},\tilde{\C{E}})$ is a \emph{subgraph} of a mixed graph $\C{G}=(\C{V},\C{E})$ if $\tilde{\C{V}}\subseteq\C{V}$ and $\tilde{\C{E}}\subseteq\C{E}$.

A \emph{clustered mixed graph} is a triple $\C{G}=(\C{V},\C{E},\C{P})$ where $(\C{V},\C{E})$ is a mixed graph and $\C{P}$ is a partition of the nodes $\C{V}$, such that dashed edges $\gtod,\gotd$ only appear between nodes in the same element of $\C{P}$. Each element of $\C{P}$ is called a \emph{cluster} of the clustered mixed graph. A clustered mixed graph $\C{G}=(\C{V},\C{E},\C{P})$ induces a mixed graph $\col(\C{G})$ with nodes $\C{P}$, a directed edge $K \gto L$ for $K \ne L$ iff there is a directed edge $k \gto l$ in $\C{G}$ for some $k \in K, l \in L$, and a bidirected edge $K \goto L$ for $K \ne L$ iff there is a bidirected edge $k \goto l$ in $\C{G}$ for some $k \in K, l \in L$. This construction can be thought of as ``collapsing'' the clusters in the clustered mixed graph into nodes and subsequently removing self-cycles.

\subsection{Random differential equations}
\label{sec:RDEs}

In this subsection, we give a brief overview of some key aspects of random differential equations~\citep[for more details, see][]{Bun72,NR13}. Random differential equations (RDEs) are similar to ordinary differential equations (ODEs), but can deal with randomness in the initial conditions and in the parameters. Due to their close connection to ODEs they can be analyzed by use of methods that are analogous to those in the theory of ODEs~\citep{Bun72}. Their formalism is conceptually easier than the formalism of the white-noise driven stochastic differential equations (SDEs), while still being applicable to those systems via the generalized Doss-Sussmann correspondence~\citep[see][]{NR13,JK11}. They have been used for many years in a wide range of applications~\citep[see, for example,][]{Bun72,Soo73,Sob90,NR13,HK17,LLWS+20}. 

A stochastic process $\B{X}:T\times\Omega\to\RN^d$ is a \emph{solution} of a \emph{(first-order) random differential equation}
\begin{equation}
\label{eq:RDE}
\B{X}' = \B{f}(\B{X},\B{E}) \,,
\end{equation}
where $\B{f}:\RN^d\times\RN^e\to\RN^d$ is a measurable function and $\B{E}:T\times\Omega\to\RN^e$ a stochastic process, if for $\Prb$-almost every $\omega\in\Omega$ the (first-order) ordinary differential equation\footnote{These ordinary differential equations are also called \emph{explicit ordinary differential equations}~\citep{AP98}. Similarly, the random differential equations~\eqref{eq:RDE} are also called \emph{explicit random differential equations}.}
$$
\B{X}'_t(\omega) = \B{f}(\B{X}_t(\omega),\B{E}_t(\omega))
$$
holds for all $t\in T$. 
An \emph{initial condition} of the RDE~\eqref{eq:RDE} is a tuple $(t_0,\B{X}_{[0]})$ 
that specifies those solutions $\B{X}$ of the RDE~\eqref{eq:RDE} that satisfy for $\Prb$-almost every $\omega\in\Omega$
\begin{equation*}
  \B{X}_{t_0}(\omega)=\B{X}_{[0]}(\omega)
\end{equation*}
at the initial time $t_0$. Since every $n^{\text{th}}$-order ODE can be rewritten as a system of first-order ODEs, the general form of the random differential equation~\eqref{eq:RDE} can be used to express analogously all the \emph{$n^{\text{th}}$-order random differential equations}.\footnote{Furthermore, explicit time-dependence of $\B{f}$ can be incorporated by adding a dummy variable with $t$ with dynamics $t'=1$ and initial condition $t_{[0]}=0$.}

The inclusion of randomness in the equations can be classified into two basic types. The first type consists of \emph{randomness in the initial conditions}, that is, the initial conditions are not a.s.\ equal to a constant deterministic process. The second type consists of \emph{randomness in the parameters}, that is, the process $\B{E}$ is not a.s.\ equal to a deterministic stochastic process. Of course, a combination of both types can hold. In particular, an RDE together with an initial condition reduces to an \emph{initial value problem} for ODEs if it has no randomness in both the initial conditions and the parameters. 

If the stochastic process $\B{E}$ is continuous, sufficient conditions that guarantee the existence and uniqueness of solutions for any initial condition can be found in~\citet{Bun72} and \citet{KP92}. These results are similar to the uniqueness and existence theorems for ODEs~\citep{CL55}. 
\begin{example}[Two masses coupled by a spring]
\label{ex:TwoHarmonicOscillator}
Consider a one-dimensional system of two point masses $m_1$ and $m_2$ with positions $X_1$ and $X_2$ respectively that are coupled by an ideal spring with spring constant $\kappa_1>0$ and equilibrium length $L_1>0$ under influence of friction with friction coefficients $b_1,b_2\geq 0$ respectively (see Figure~\ref{fig:TwoMassSpringSystem} (left)). The equations of motion of this system, whose derivation can be found in physics textbooks, are given by the second-order random differential equations
\begin{equation*}
  \left\{\begin{aligned}
  X''_1 & = \frac{\kappa_1}{m_1} (X_2 - X_1 - L_1) -\frac{b_1}{m_1} X_1' \\
  X''_2 & = \frac{\kappa_1}{m_2} (X_1 - X_2 + L_1) -\frac{b_2}{m_2} X_2' \,. 
  \end{aligned}\right. 
\end{equation*}
Randomness may enter the system via the initial condition $(t_0,(X_1(t_0),X_1'(t_0),X_2(t_0),X_2'(t_0)))$ or via the parameters. For example, instead of assuming that the length $L_1$ has a fixed value, we can assume that it is an exogenous random variable distributed according to some distribution. The system of equations then forms an RDE.
\begin{figure}
  \begin{center}
    \begin{tikzpicture}[scale=0.6]
      \begin{scope}
        \node (Q1) at (0,0.6) {$m_1$};
        \node (Q2) at (3.5,0.6) {$m_2$};
        \begin{scope}[xshift=-1.5mm]
          \node (l1) at (1.75,-0.6) {$L_1$};
        \end{scope}
        \fill[black] (0,0) circle (.2);
        \fill[black] (3.5,0) circle (.2);
        \draw[thick,decorate,decoration={coil,aspect=0.7,amplitude=4,segment length=3mm}] (0,0) -- (3.5,0);
      \end{scope}
      \begin{scope}[xshift=10cm]
        \fill[pattern=north east lines,draw=none] (-0.75,-1) rectangle (-0,1);
        \draw (0,-1) -- (0,1);
        \node (Q1) at (0,0.6) {};
        \node at (0.6,0.6) {$m_1$};
        \node (Q2) at (3.5,0.6) {$m_2$};
        \begin{scope}[xshift=-1.5mm]
          \node (l1) at (1.75,-0.6) {$L_1$};
        \end{scope}
        \fill[black] (0,0) circle (.2);
        \fill[black] (3.5,0) circle (.2);
        \draw[thick,decorate,decoration={coil,aspect=0.7,amplitude=4,segment length=3mm}] (0,0) -- (3.5,0);
      \end{scope}
    \end{tikzpicture}
  \end{center}
  \caption{Two masses coupled by a spring, freely drifting in space (left, see Example~\ref{ex:TwoHarmonicOscillator}) and with one of the masses attached to a fixed point (right, see Example~\ref{ex:TwoHarmonicOscillatorIntervened}).}
  \label{fig:TwoMassSpringSystem}
\end{figure}
\end{example}


In this paper, we propose a modeling class that allows to model the causal semantics of stochastic processes with RDEs in an unambiguous way. The following example illustrates that modeling interventions on RDEs, and thereby grounding their causal semantics, is not a completely trivial matter.
\begin{example}[Two masses coupled by a spring, continued]
\label{ex:TwoHarmonicOscillatorIntervened}
  Consider again the RDE that describes the two masses coupled by an ideal spring from Example~\ref{ex:TwoHarmonicOscillator}. 
  These equations denote a symmetric relation, that is, for both equations $X_1$ can be expressed in terms of $X_2$, and vice versa. The causal relations between the processes $X_1$ and $X_2$ are not inherently implied by the form of the equations. For example, what happens to $X_2$ if we fix the mass $m_1$ to a fixed wall, say, at $X_1=0$ (see Figure~\ref{fig:TwoMassSpringSystem} (right))? The corresponding RDE for $X_1$ and $X_2$ is then given by 
\begin{equation*}\begin{split}
  \left\{\begin{aligned}
  X_1 & = 0 \\
  X''_2 & = \frac{\kappa_1}{m_2} (X_1 - X_2 + L_1) -\frac{b_2}{m_2} X_2' \,.
  \end{aligned}\right. 
\end{split}\end{equation*}
  In both cases we implicitly assumed that each mass has its own equation of motion, that is, the first and second equation determine the motion of the mass $m_1$ and $m_2$, in terms of the processes $X_1$ and $X_2$, respectively. Therefore, the intervention of fixing the mass $m_1$ to the wall is accomplished by changing only the equation for $m_1$ to the equation $X_1=0$. If instead we had changed the other equation to $X_1=0$, then, as one can easily verify, $X_2$ would always be fixed, which does not correspond to the expected physical behavior. This additional ``structure'' of knowing which RDE determines the dynamics of which process is not ``intrinsically'' defined by the RDE. Moreover, RDEs usually do not include zeroth-order equations (also referred to as ``algebraic equations''), such as $X_1=0$. Allowing for RDEs of arbitrary order, including zeroth-order, allows to model a wide range of interventions on these models. For example, instead of fixing the mass $m_1$ to the fixed wall at $X_1=0$ we could fix it to a wall that is driven by some external force, such as $X_1= A\sin(2\pi f t)$ for some $A,f>0$.
\end{example}

\section{\CDMFull s}
\label{sec:CDMs}

In this section, we introduce the class of \CDMfull s (\CDM s) that allows to formally specify causal semantics for any RDE of arbitrary order (including zeroth-order).
We organize the differential equations of the RDEs in a structural way, similar to how this is done for structural causal models, such that each differential equation expresses the causal mechanism that governs the dynamics of a single stochastic process (corresponding to a single component of the system). This allows us to model stochastic idealized interventions targeting certain components in dynamical models, similarly to how this is done for SCMs. 

We start in Section~\ref{sec:Notation} with introducing the notation and terminology that will be used throughout the paper. In Section~\ref{sec:DefCDMSol}, we formally define \CDM s and their solutions. In Section~\ref{sec:Interventions}, we formalize the causal semantics of \CDM s in terms of stochastic ``perfect'' interventions. In Section~\ref{sec:Graph}, we introduce and discuss a graphical representation for \CDM s. In Section~\ref{sec:InitialConditions}, we discuss the initial conditions and how these relate to the existence of solutions. In Section~\ref{sec:ExistenceAndUniquenessOfSolutions}, we state results about the existence and uniqueness of solutions of certain classes of \CDM s. We finish in 
Section~\ref{sec:MarkovPropertyCDM} by deriving a Markov property for \CDM s with initial conditions,
suitable for both the solutions of the \CDM\ and the evaluation of the solutions at any point in time.

\subsection{Notation and terminology}
\label{sec:Notation}

Let $\C{I}=\{1,\ldots,d\}$ be a finite index set and $\BC{X}=\prod_{i\in\C{I}}\C{X}_i$ the product of the domains of the components of a system, where domain $\C{X}_i=\RN^{d_i}$ encodes the range of possible values that the $i^{\text{th}}$ component can take. The stochastic process $\B{X}=(X_1,\ldots ,X_d):T\times\Omega\to\BC{X}$ has component processes $X_i:T\times\Omega\to\C{X}_i$. 

Let $i\in\C{I}$ and $n_i\in\NN_0$. If for the $i^{\text{th}}$ component $X_i$ the $n_i^{\text{th}}$-order derivative exists, then the \emph{complete $n_i^{\text{th}}$-order derivative} of $X_i$, defined as the stochastic process $\widebar{X_i}^{(n_i)}:=(X_i,X'_i,X''_i,\dots,X_i^{(n_i)}):T\times\Omega\to\C{X}_i^{n_i+1}$, is the tuple of all the derivatives of $X_i$ up to and   including order $n_i$. We adopt a similar notation for the values in $\C{X}_i^{n_i+1}$, that is, $\widebar{x}_i^{(n_i)}\in\C{X}_i^{n_i+1}$. Each component $X_i^{(k_i)}$ of $\widebar{X_i}^{(n_i)}$, or similarly $x_i^{(k_i)}$ of $\widebar{x}_i^{(n_i)}$, corresponds to an index $i^{(k_i)}$, which gives the index set $\bar{i}^{(n_i)}:=\{ i^{(k_i)} : 0\leq k_i \leq n_i\}$ for $\widebar{X}_i^{(n_i)}$, where the index $i^{(0)}$ is also written as $i$. 

Let $\B{n}=(n_1,\ldots,n_d)\in\NN_0^{\C{I}}$ be a tuple. If the $n_i^{\text{th}}$-order derivative of $X_i$ exists for every $i\in\C{I}$, then the \emph{$\B{n}^{\text{th}}$-order derivative} of $\B{X}$ is defined as the stochastic process $\B{X}^{(\B{n})}:=(X_1^{(n_1)},\ldots, X_d^{(n_d)}):T\times\Omega\to\BC{X}$ and the \emph{complete $\B{n}^{\text{th}}$-order derivative} of $\B{X}$ is defined as the stochastic process $\widebar{\B{X}}^{(\B{n})}:=(\widebar{X_1}^{(n_1)},\widebar{X_2}^{(n_2)},\ldots,\widebar{X_d}^{(n_d)}):T\times\Omega\to\BC{X}^{\B{n}+1}$, where $\BC{X}^{\B{n}+1}:=\prod_{i=1}^{d} \C{X}_i^{n_i+1}$. We adopt a similar notation for the values in $\BC{X}^{\B{n}+1}$, that is, $\widebar{\B{x}}^{(\B{n})}\in\BC{X}^{\B{n}+1}$. Similarly, each component $X_i^{(k_i)}$ of $\widebar{\B{X}}^{(\B{n})}$ corresponds to an index $i^{(k_i)}$ which gives the index set $\widebar{\C{I}}^{(\B{n})}:= \bigcup_{i\in\C{I}} \bar{i}^{(n_i)}$ for $\widebar{\B{X}}^{(\B{n})}$.  

For a subset $I:=\{i_1,\dots,i_k\}\subseteq\C{I}$ we will use the notation
$\B{n}_I:= (n_{i_1},\dots,n_{i_k})$ and write $\BC{X}_I = \prod_{i\in I} \C{X}_i$
and $\BC{X}^{\B{n}_I+1}_I = \prod_{i\in I} \C{X}_i^{n_i+1}$. For the $I^{\text{th}}$ components of the process $\B{X}$ and the complete $\B{n}^{\text{th}}$-order derivative $\widebar{\B{X}}^{(\B{n})}$, we write $\B{X}_I:=(X_{i_1},\dots,X_{i_k})$ and $\widebar{\B{X}}^{(\B{n}_I)}_I:=(\widebar{X}_{i_1}^{(n_{i_1})},\dots,\widebar{X}_{i_k}^{(n_{i_k})})$ respectively. Similarly, for the values
in $\BC{X}_I$ and $\BC{X}^{\B{n}_I+1}_I$, we write
$\B{x}_I:=(x_{i_1},\dots,x_{i_k})\in\BC{X}_I$ and
$\widebar{\B{x}}^{(\B{n}_I)}_I:=(\widebar{x}_{i_1}^{(n_{i_1})},\dots,\widebar{x}_{i_k}^{(n_{i_k})})
\in \BC{X}^{\B{n}_I+1}_I$ respectively.

In this notation, a stochastic process $X$ is a \emph{$C^n$-stochastic process}, if its complete $n^{\text{th}}$-order derivative $\widebar{X}^{(n)}$ exists and is continuous. Similarly, we call a stochastic process $\B{X}$ a \emph{$C^{\B{n}}$-stochastic process}, if its complete $\B{n}^{\text{th}}$-order derivative $\widebar{\B{X}}^{(\B{n})}$ exists and is continuous. We denote by $C^{\B{n}}(T,\BC{X})$ the space of $C^{\B{n}}$-stochastic processes. For $T=[t_0, t_1]\subseteq\RN$ compact, the space $C^{\B{n}}(T,\BC{X})$ forms a standard measurable space with Borel $\sigma$-algebra given by the $C^{\B{n}}$-norm $$\| \B{X} \|^{(\B{n})} := \sum_{i\in\C{I}} \sum_{k=0}^{n_i} \sup_{t\in T} \|X^{(k)}_i(t)\|
\,.$$

\subsection{\CDMFull s and their solutions}
\label{sec:DefCDMSol}

Informally, we think of an \CDM\ as an SCM where we replace the random variables of the SCM by stochastic processes and their derivatives, and where each structural equation of the SCM becomes a random differential equation of arbitrary order. 
This generalizes the class of SCMs to the continuous time domain and enables a causal semantics for a broad range of random dynamical models. In this paper, we closely follow the terminology of \citet{BPSM19} for SCMs and extend it to \CDM s.

\begin{definition}[\CDMFull]
  \label{def:CDM}
  A \emph{\CDMfull\ (\CDM)} is a tuple\footnote{We often use boldface for variables that have multiple components, that is, which take values in a Cartesian product.}
$$
  \C{R} := \langle \C{I}, \C{J}, \BC{X}, \BC{E}, \B{n}, \B{f}, \B{E} \rangle
$$
where 
\begin{itemize}
\item 
  $\C{I}$ is a finite index set for endogenous processes,
\item 
  $\C{J}$ is a disjoint finite index set for exogenous processes,
\item
  $\BC{X} = \prod_{i\in\C{I}} \C{X}_i$ is the product of the domains of the endogenous processes, where each domain $\C{X}_i = \RN^{d_i}$,
\item 
  $\BC{E} = \prod_{j\in\C{J}} \C{E}_j$ is the product of the domains of the exogenous processes, where each domain $\C{E}_j = \RN^{e_j}$,
\item 
  $\B{n} = (n_i)_{i\in\C{I}} \in \NN^{\C{I}}_{0}$ is the \emph{order tuple},
\item 
  $\B{f}:\BC{X}^{\B{n}+1} \times \BC{E} \to \BC{X}$ is a measurable function that specifies the \emph{dynamic causal mechanism},
\item 
  $\B{E}: T \times \Omega \to \BC{E}$ is an exogenous stochastic process with
  independent components, that is, $(E_j)_{j\in\C{J}}$ is independent.
\end{itemize}
\end{definition}
\Stephan{Remark: If we want to define the order tuple $\B{n}$ such that for each $i$ the order $n_i$ is the minimum number such that the \SDEfull s depend on the process $X_i^{(n_i)}$ (or functionally depend on $x_i^{(n_i)}$), then 
\begin{enumerate}
  \item we get a very ugly definition, by adding the condition that the causal mechanism should functionally depend on the highest-order variable $\B{x}_i^{(n_i)}$ for each $i\in\C{I}$. Functional relations are defined later in Definition~\ref{def:FunctionalAndIntegratedParents} in Section~\ref{sec:Graph}. Actually, if we decide to go this route, then to me it would make more sense to define the dynamic causal mechanism in terms of an augmented graph $\C{G}$, that is, $\B{f}:\BC{X}_{\pa_{\C{G}}(\C{I})}\times\BC{E}_{\pa_{\C{G}}(\C{J})}\to\BC{X}$. \Joris{Notation clash: $K$ was reserved for intervention processes.} For SCMs we also had our reasons to define the causal mechanism $\B{f}$ in terms of the spaces $\BC{X}$ and $\BC{E}$ rather than $\BC{X}_{\pa_{\C{G}}(\C{I})}$ and $\BC{E}_{\pa_{\C{G}}(\C{J})}$ for some augmented graph $\C{G}$. 
\item the definition of stochastic perfect interventions is not possible without the graph-based \CDM. Since for an intervention $\intervene(I,\B{K}_I)$  we need to update the order tuple components $\B{n}_I$ accordingly. Setting them to $\B{0}$ will not work generally, since there may exist a dependence on the higher-order derivatives of the intervened processes in the nonintervened components of $\B{f}$.
\end{enumerate}
If we do not let $\B{n}$ be such that each $n_i$ is a minimum number, as above, then we can take it arbitrary high for a given dynamic causal mechanism. But, then the nodes in the graph will also be arbitrary high, which seems a bit to arbitrary to me. One idea would be that we can define the graph only on the functionally dependent processes, which will always give the most sparse graph. 

However, I think it is best to just pick the order tuple $\B{n}$ wisely. Since picking an higher-order tuple has an effect on the smoothness assumption on the solution, that is, the complete $\B{n}^{\text{th}}$-order derivative $\widebar{\B{X}}^{(\B{n})}$ should exist. So if one want the least assumptions, then one need to take the order tuple $\B{n}$ as low as possible. Which then naturally gives a sparse graph.
}

The solutions of a \CDMfull\ in terms of stochastic processes are defined by the associated \SDEfull s.
\begin{definition}[Solution of an \CDM]
  \label{def:CDM_soln}
  A stochastic process $\B{X}:T \times \Omega \to \BC{X}$ is a \emph{solution} of the \emph{\SDEfull s} (\SDE s) associated to \CDM\ $\C{R}$,
\begin{equation*}
\B{X} = \B{f}(\widebar{\B{X}}^{(\B{n})},\B{E}) \,, 
\end{equation*}
if $\B{X}$ is a $C^{\B{n}}$-stochastic process,
and for $\Prb$-almost every $\omega\in\Omega$ the ordinary differential equations\footnote{These equations are called
  \emph{implicit ordinary differential equations} if the Jacobian matrix
  $\frac{\partial \B{f}(\widebar{\B{x}}^{(\B{n})},\B{e})}{\partial\B{x}^{(\B{n})}}$ is nonsingular for all its argument values in an appropriate
  domain, otherwise they are called \emph{differential-algebraic equations}~\citep{AP98}.}
\begin{equation*}
  \B{X}_t(\omega) = \B{f}\big(\widebar{\B{X}}^{(\B{n})}_t(\omega), \B{E}_t(\omega)\big)
\end{equation*}
hold for all $t\in T$.
\end{definition}
The value $n_i$ of the order tuple $\B{n}$ denotes the highest-order derivative of $X_i$ that may occur in the \SDEfull s.
Note that taking higher $n_i$'s will in general reduce the set of possible solutions, due to additional imposed smoothness constraints on the solutions. 
In contrast to the common way of writing RDEs (see equation~\eqref{eq:RDE}), the (higher-order) derivatives of the endogenous processes of an \CDM\ always appear on the right-hand side of the \SDE s.\footnote{\label{fn:RDE2CDM}For every RDE of the form $\B{X}'=\B{f}(\B{X},\B{E})$ with $\B{f}$ and $\B{E}$ continuous, there exists an \CDM\ with the same solutions: $\B{X}$ is a solution of the RDE if and only if it is a solution of the \CDM\ $\C{R}$ with the \SDE\ $\B{X}=\B{X}-\B{X}' + \B{f}(\B{X},\B{E})$, as long as $\B{n} = 1$ (since all solutions of the RDE must be continuously differentiable).} This notation explicitly allows us to model zeroth-order \SDEfull s, that is, equations that contain no derivatives of order one or higher, in other words, random algebraic equations.

In particular, if \emph{all} \SDEfull s are of zeroth order and the exogenous stochastic processes in the model are constant in time (that is, random variables), then the \CDMfull\ reduces to a structural causal model \citep[see][]{BPSM19}. In contrast to \citep{BPSM19}, we define an SCM here in terms of an exogenous random variable instead of an exogenous distribution.
\begin{definition}[Structural causal model]
\label{def:SCM}
A \emph{structural causal model (SCM)} is a tuple
$$
\C{M}:=\langle \C{I}, \C{J}, \BC{X}, \BC{E}, \B{f}, \B{E} \rangle \,,
$$
such that 
$\langle \C{I}, \C{J}, \BC{X}, \BC{E}, \B{0}, \B{f}, \B{E} \rangle$ is an \CDM\ with $\B{E}$ a random variable.
\end{definition}
That is, we can identify SCMs with certain special cases of \CDM s.
Similarly, we can identify the solutions of an SCM with the (constant) solutions of the corresponding \CDM.
The following definition is equivalent to Definition~\ref{def:CDM_soln} when the latter is applied to an SCM.
\begin{definition}[Solution of an SCM]
  \label{def:SCMSolutions}
A random variable $\B{X}:\Omega\to\C{X}$ is a \emph{solution} of the \emph{structural equations} associated to SCM $\C{M}$,
$$
  \B{X} = \B{f}(\B{X},\B{E}) \,,
$$
if for $\Prb$-almost every $\omega\in\Omega$
$$
\B{X}(\omega) = \B{f}(\B{X}(\omega),\B{E}(\omega)) 
$$
holds.
\end{definition}

Similar to the structural equations of an SCM \citep{Pea09,Woo03}, the \SDEfull s of an \CDM\ model the underlying causal mechanisms in a structural way, that is, each \SDEfull\ expresses a specific endogenous process (on the left-hand side) in terms of a dynamic causal mechanism depending on certain processes and their derivatives (on the right-hand side). It is this additional structure, which allows us to explicitly model the causal semantics, that distinguishes \CDMfull s from dynamical models such as ODEs and RDEs.\footnote{The importance of assigning a differential equation to an endogenous variable was already observed in \citep{MJS13}.} 
Allowing for zeroth and higher-order derivatives of $X_i$ in the \SDEfull s gives rise to a broad range of random dynamical models that can be described by an \CDM, ranging from ODEs (including first-order ODEs as in~\citep{MJS13}), RDEs (as in Section~\ref{sec:RDEs}) and more general random dynamical systems such as partially equilibrated systems (as in~\citep{IS94}).

\begin{example}[Damped coupled harmonic oscillator]
\label{ex:HarmonicOscillator}
Consider a one-dimensional system of $d$ point masses $m_i>0$
($i=1,\dots,d$) with positions $X_i \in \RN$, which are
coupled by ideal springs, with spring constants $\kappa_i>0$ and equilibrium lengths $L_i>0$
($i=1,\dots,d-1$), under influence of friction with friction coefficients $b_i\geq 0$ ($i=1,\dots,d$) (see Figure~\ref{fig:HarmonicOscillator} left).
\begin{figure}
  \begin{center}
    \begin{tikzpicture}[scale=0.6]
    \node (Q1) at (1.75,0.6) {$m_1$};
    \node (Q2) at (3.5,0.6) {$m_2$};
    \node (Q3) at (5.25,0.6) {$m_3$};
    \node (Q4) at (7,0.6) {$m_4$};
    \node (Q5) at (8.75,0.6) {$m_5$};
    \begin{scope}[xshift=-1.5mm]
      \node (l1) at (2.625,-0.6) {$L_1$};
      \node (l2) at (4.375,-0.6) {$L_2$};
      \node (l3) at (6.125,-0.6) {$L_3$};
      \node (l4) at (7.875,-0.6) {$L_4$};
    \end{scope}
    \fill[black] (1.75,0) circle (.2);
    \fill[black] (3.5,0) circle (.2);
    \fill[black] (5.25,0) circle (.2);
    \fill[black] (7,0) circle (.2);
    \fill[black] (8.75,0) circle (.2);
    \draw[thick,decorate,decoration={coil,aspect=0.7,amplitude=4,segment length=3mm}] (1.75,0) -- (3.5,0);
    \draw[thick,decorate,decoration={coil,aspect=0.7,amplitude=4,segment length=3mm}] (3.5,0) -- (5.25,0);
    \draw[thick,decorate,decoration={coil,aspect=0.7,amplitude=4,segment length=3mm}] (5.25,0) -- (7,0);
    \draw[thick,decorate,decoration={coil,aspect=0.7,amplitude=4,segment length=3mm}] (7,0) -- (8.75,0);
  \end{tikzpicture} 
  \hspace{1.5cm}
    \begin{tikzpicture}[scale=0.6]
    \fill[pattern=north east lines,draw=none] (1.25,-1) rectangle (2,1);
    \draw (2,-1) -- (2,1);
    \node (Q1) at (2,0.6) [xshift=1.8ex] {$m_1$};
    \node (Q2) at (4.25,0.6) {$m_2$};
    \node (Q3) at (6.5,0.6) {$m_3$};
    \node (Q4) at (8.75,0.6) {$m_4$};
    \node (Q5) at (11,0.6) [xshift=-1.8ex] {$m_5$};
    \begin{scope}[xshift=-1.5mm]
      \node (l1) at (3.125,-0.6) {$L_1$};
      \node (l2) at (5.375,-0.6) {$L_2$};
      \node (l3) at (7.625,-0.6) {$L_3$};
      \node (l4) at (9.875,-0.6) {$L_4$};
    \end{scope}
    \fill[black] (2,0) circle (.2);
    \fill[black] (4.25,0) circle (.2);
    \fill[black] (6.5,0) circle (.2);
    \fill[black] (8.75,0) circle (.2);
    \fill[black] (11,0) circle (.2);
    \draw (11,-1) -- (11,1);
    \fill[pattern=north east lines,draw=none] (11,-1) rectangle (11.75,1);
    \draw[thick,decorate,decoration={coil,aspect=0.7,amplitude=4,segment length=4mm}] (2,0) -- (4.25,0);
    \draw[thick,decorate,decoration={coil,aspect=0.7,amplitude=4,segment length=4mm}] (4.25,0) -- (6.5,0);
    \draw[thick,decorate,decoration={coil,aspect=0.7,amplitude=4,segment length=4mm}] (6.5,0) -- (8.75,0);
    \draw[thick,decorate,decoration={coil,aspect=0.7,amplitude=4,segment length=4mm}] (8.75,0) -- (11,0);
    \node at (2,1.3) {\scriptsize $X_1 = 0$};
    \node at (11,1.3) {\scriptsize $X_5 = L$};
  \end{tikzpicture}
\end{center}
\caption{Damped coupled harmonic oscillator model $\C{R}$ of Example~\ref{ex:HarmonicOscillator} (left) and the intervened model $\C{R}_{\intervene(\{1,5\},(0,L))}$ of Example~\ref{ex:HarmonicOscillatorIntervened} (right), both for $d=5$.}
\label{fig:HarmonicOscillator}
\end{figure}
This system can be modeled by the \CDM\footnote{We abuse notation here; more formally, we should use an index set for $\C{J}$ that is disjoint from $\C{I}$, for example, $\{\tilde{1},\dots,\tilde{d}-1\}$.} $\C{R} = \langle \{1,\dots,d\}, \{1,\dots,d-1\}, \RN^d, \RN^{d-1}, \B{n}, \B{f}, \B{E} \rangle$ with order tuple $\B{n}:=(2,\dots,2)$, where the exogenous process $\B{E}=\B{L}:=(L_1,\dots,L_{d-1})$ is constant in time (that is, a random variable), and the causal mechanism is specified by the \SDEfull s
\begin{equation*}
  \left\{\begin{aligned}
    X_1= f_1(\widebar{\B{X}}^{(\B{n})},\B{L}) & := X_2 - L_1 - \frac{b_1}{\kappa_1}X_1' - \frac{m_1}{\kappa_1}X_1'' \\
   X_i = f_i(\widebar{\B{X}}^{(\B{n})},\B{L}) & := \frac{\kappa_i}{\kappa_i + \kappa_{i-1}} (X_{i+1} - L_i) + \frac{\kappa_{i-1}}{\kappa_i + \kappa_{i-1}} (X_{i-1} + L_{i-1}) \\
              & \phantom{:=} -\frac{b_i}{\kappa_i + \kappa_{i-1}} X_i' - \frac{m_i}{\kappa_i + \kappa_{i-1}} X_i'' \quad\quad(i=2,\dots,d-1)\\
   X_d = f_d(\widebar{\B{X}}^{(\B{n})},\B{L}) & := X_{d-1} + L_{d-1} -\frac{b_d}{\kappa_{d-1}}X_d' - \frac{m_d}{\kappa_{d-1}}X_d'' \,.
  \end{aligned}\right. 
\end{equation*}
The motion of the masses, in terms of their positions $X_i$, velocities $X_i'$ and accelerations $X_i''$, is described by a separate equation of motion for each mass. For the case $d=2$, this \CDM\ $\C{R}$ has the same solutions as those described by the RDE in Example~\ref{ex:TwoHarmonicOscillator}.
\end{example}


The following example motivates why we only consider processes as solutions of \CDM s in case they satisfy the smoothness conditions.
\begin{example}[Sufficient smoothness of the solutions]
\label{ex:ContinuityOfSolutions}
Let $\C{R}=\langle \{1\},\emptyset, \C{X},\C{E}, n, f, E \rangle$ be the \CDM\ with $\C{X}=\RN$, $\C{E}$ the singleton $\{*\}$, $n=0$, the dynamic causal mechanism $f:\C{X}\times\C{E}\to\C{X}$ given by $f(x,e)=x-x^2+1$, and $E$ the trivial exogenous process. The zeroth-order \SDEfull\ associated to $\C{R}$ reads
$$
X = X - X^2 + 1 \,.  
$$
This \SDEfull\ does not depend on any exogenous process. The set of endogenous processes $X:T\times\Omega\to\RN$ that satisfy the \SDEfull\ consists of all stochastic processes in $\{-1,1\}^{T\times\Omega}$. Most of the stochastic processes in $\{-1,1\}^{T\times\Omega}$ are not continuous. The solutions of $\C{R}$ are exactly those processes in $\{-1,1\}^{T\times\Omega}$ that are $C^0$-stochastic processes. These are the processes that are constant in time, that is, the random variables of $\{-1,1\}^{\Omega}$. In particular, the solutions of the \CDM\ $\C{R}$ correspond exactly to the solutions of the SCM described by the above structural equation.
\end{example}


\subsection{Interventions}
\label{sec:Interventions}

Interventions on a \CDMfull\ can be modeled in different ways.
We consider here a stochastic version of \emph{perfect} interventions\footnote{These are also referred to as \emph{ideal}, \emph{hard}, \emph{structural}, \emph{surgical}, \emph{atomic}~\citep{ES07} or \emph{independent}~\citep{KHNA04} interventions.} on the endogenous processes~\citep{ES07}
that are analogous to stochastic perfect interventions in structural causal models~\citep{Pea09,Ebe14}. A stochastic perfect intervention on some endogenous process forces the intervened process to be equal to a given independent exogenous process. More generally, we model a stochastic perfect intervention on a subset $I:=\{i_1,\dots,i_k\}\subseteq\C{I}$ of the
endogenous processes by forcing those processes $\B{X}_I$ to be equal to the intervened processes $\B{K}_I$, by changing the model such that the corresponding dynamical structural equations become $\B{X}_I = \B{K}_I$. The process $\B{K}_I$ is treated 
as an independent exogenous process, such that all its components $K_i$ are 
mutually independent and independent from all the other exogenous processes that were already present in the model in the absence of the intervention. The dynamic causal mechanisms of the other endogenous processes $\C{I}\setminus I$ are untouched and 
their dynamics are still specified by the same \SDEfull s associated to those processes in the absence of the intervention, that is\footnote{For $I\subseteq\C{I}$ we adopt the notation $\setminus I$ for $\C{I}\setminus I$.}
$$
\B{X}_{\setminus I} = \B{f}_{\setminus I}(\widebar{\B{X}}^{(\B{n})}, \B{E}) \,.
$$
This yields the following formal definition of an intervened \CDMfull.
\begin{definition}[Stochastic perfect intervention on an \CDM]
\label{def:InterventionsCDM}
Let $\C{R} = \langle \C{I}, \C{J}, \BC{X}, \BC{E}, \B{n}, \B{f},
\B{E} \rangle$ be an \CDM, $I\subseteq\C{I}$ a subset, and $\B{K}_I:T\times\Omega\to\BC{X}_I$ a stochastic process such that $((K_i)_{i\in I},(E_j)_{j\in\C{J}})$ is independent. The
  \emph{stochastic perfect intervention $\intervene(I,\B{K}_I)$} maps $\C{R}$ to the \CDM\footnote{We abuse notation here; more formally, we should make a disjoint copy $\tilde{I} := \{\tilde{i} : i \in I\}$ and use $\tilde{I} \cup \C{J}$ as the new exogenous index set instead of $I \cup \C{J}$, to keep the endogenous indices $\C{I}$ and the exogenous indices $\tilde{I} \cup \C{J}$ disjoint.}
$$
\C{R}_{\intervene(I,\,\B{K}_I)} := \langle \C{I}, I \cup \C{J}, \BC{X}, \BC{X}_I\times\BC{E}, \B{n}, \tilde{\B{f}}, (\B{K}_I,\B{E}) \rangle \,,
$$
where the \emph{intervened causal mechanism} $\tilde{\B{f}}:\BC{X}^{\B{n}+1}\times(\BC{X}_I\times\BC{E})\to\BC{X}$ is given by
\begin{equation}
\label{eq:IntervenedMechanism}
  \tilde{f}_i(\widebar{\B{x}}^{(\B{n})},(\B{e}_I,\B{e}_{\C{J}})) = \begin{cases}
  f_i(\widebar{\B{x}}^{(\B{n})},\B{e}_{\C{J}}) \quad & i\in \C{I}\setminus I \\
  e_i \quad & i\in I \,.
  \end{cases}
\end{equation}
  We call a stochastic perfect intervention $\intervene(I,\B{K}_I)$ a \emph{perfect intervention} if $\B{K}_I$ is a deterministic stochastic process (that is, if it does not depend on $\omega$).
\end{definition}
This definition explicitly exposes a hitherto implicit but crucial modeling assumption: exogenous processes are not caused by endogenous processes. Indeed, no stochastic perfect intervention on any subset of the endogenous processes will lead to a change in any of the exogenous processes. 

\begin{example}
\label{ex:HarmonicOscillatorIntervened}
Consider the damped coupled harmonic oscillator represented by the \CDM\ $\C{R}$ of Example~\ref{ex:HarmonicOscillator}. Performing the perfect interventions on the masses $m_1$ and $m_d$ by fixing $m_1$ and $m_d$ to the walls at $X_1=0$ and $X_d=L>0$, respectively, (see Figure~\ref{fig:HarmonicOscillator} (right)) yields the model $\C{R}_{\intervene(\{1,d\},(0,L))}$ with the \SDEfull s
 $$
  \left\{\begin{aligned}
   X_1 &= 0 \\
     X_i &= \frac{\kappa_i}{\kappa_i + \kappa_{i-1}} (X_{i+1} - L_i) + \frac{\kappa_{i-1}}{\kappa_i + \kappa_{i-1}} (X_{i-1} + L_{i-1}) \\
                & \phantom{:=} -\frac{b_i}{\kappa_i + \kappa_{i-1}} X_i' - \frac{m_i}{\kappa_i + \kappa_{i-1}} X_i'' \quad\quad(i=2,\dots,d-1)\\
     X_d  &= L \,.
  \end{aligned}\right. 
  $$
\end{example}
It is clear from the definition that performing stochastic perfect interventions on disjoint subsets of the endogenous processes commutes. 
In case of overlap, the \SDEfull s of the overlapping intervention targets are determined by the most recent intervention applied to them.

As a special case, Definition~\ref{def:InterventionsCDM} reduces to the usual notion of (stochastic) perfect intervention on SCMs \citep[see][]{BPSM19}. 
\begin{definition}[Stochastic perfect intervention on an SCM]
\label{def:InverventionsSCM}
Let $\C{M}=\langle \C{I}, \C{J}, \BC{X}, \BC{E}, \B{f}, \B{E} \rangle$ be an SCM, $I\subseteq\C{I}$ a subset, and $\B{K}_I:\Omega\to\BC{X}_I$ a random variable such that $((K_i)_{i\in I},(E_j)_{j\in\C{J}})$ is independent. The
\emph{stochastic perfect intervention $\intervene(I,\B{K}_I)$} maps $\C{M}$ to the SCM 
$$
\C{M}_{\intervene(I,\,\B{K}_I)} := \langle \C{I}, I \cup \C{J}, \BC{X}, \BC{X}_I\times\BC{E}, \tilde{\B{f}}, (\B{K}_I,\B{E}) \rangle \,,
$$
where $\tilde{\B{f}}$ is defined by equation~\eqref{eq:IntervenedMechanism}.
\end{definition}

This provides \CDM s with a causal semantics that is analogous to that of SCMs.  The following example illustrates how this resolves the ambiguity of the causal interpretation of the RDE of Example~\ref{ex:TwoHarmonicOscillatorIntervened}.
\begin{example}[Ambiguous causal interpretation of RDEs]
\label{ex:TwoHarmonicOscillatorAmbiguousCausalInterpretation}
Consider the \CDM\ $\C{R}$ of Example~\ref{ex:HarmonicOscillator} for $d=2$, with \SDEfull s given by
\begin{equation*}
  \left\{\begin{aligned}
    X_1&= X_2 - L_1 - \frac{b_1}{\kappa_1}X_1' - \frac{m_1}{\kappa_1}X_1'' \\
    X_2&= X_1 + L_1 -\frac{b_2}{\kappa_1}X_2' - \frac{m_2}{\kappa_1}X_2'' \,.
  \end{aligned}\right. 
\end{equation*}
  The solutions of $\C{R}$ correspond exactly to the solutions of the RDE that describes the two masses attached to a spring in Example~\ref{ex:TwoHarmonicOscillator}. 
  Fixing the mass $m_1$ to the left wall at $X_1=0$ (see Figure~\ref{fig:TwoMassSpringSystem} (right)) by performing the stochastic perfect intervention\footnote{For convenience, we write $\intervene(i,K_i)$ for a stochastic perfect intervention $\intervene(I,\B{K}_I)$ whenever $I=\{i\}$ for some $i\in\C{I}$.} $\intervene(1,K_1)$ with $K_1=0$ on $\C{R}$ gives the intervened model $\C{R}_{\intervene(1,0)}$ with \SDEfull s given by 
$$
  \left\{\begin{aligned}
    X_1 & =  0 \\
    X_2 & =  X_1 + L_1 -\frac{b_2}{\kappa_1} X_2' - \frac{m_2}{\kappa_1} X_2'' \,.
  \end{aligned}\right. 
$$
The intervened model $\C{R}_{\intervene(1,0)}$ has exactly the same solutions as the RDE in Example~\ref{ex:TwoHarmonicOscillatorIntervened}.

Consider now the \CDM\ $\tilde{\C{R}}$ that is the same as $\C{R}$ except for its dynamic causal mechanism $\tilde{\B{f}}$, for which the associated \SDEfull s are given by 
$$
  \left\{\begin{aligned}
    X_1 & =  X_2 - L_1 +\frac{b_2}{\kappa_1} X_2' + \frac{m_2}{\kappa_1} X_2''\\
    X_2 & =  X_1 + L_1 +\frac{b_1}{\kappa_1} X_1' + \frac{m_1}{\kappa_1} X_1'' \,.
  \end{aligned}\right. 
$$
Both models $\C{R}$ and $\tilde{\C{R}}$ have the same solutions as those described by the RDEs in Example~\ref{ex:TwoHarmonicOscillator}. However, the intervened models $\C{R}_{\intervene(1,0)}$ and $\tilde{\C{R}}_{\intervene(1,0)}$ have different solutions. Only the model $\C{R}_{\intervene(1,0)}$ describes the expected physical behavior (see also Example~\ref{ex:TwoHarmonicOscillatorIntervened}).
\end{example}

Stochastic perfect interventions are only defined for the endogenous processes, but not for their higher-order derivatives. The higher-order derivative processes in an \CDM\ are always obtained by differentiation of the underlying endogenous processes and hence it suffices to define the stochastic perfect interventions only for those underlying endogenous processes. Allowing for stochastic perfect intervention on both the endogenous processes and some of their higher-order derivatives will generally lead to nonsensible causal behavior, as is illustrated in the following example.
\begin{example}[Modeling higher-order derivatives as separate endogenous processes]
\label{ex:HigherOrderDerivativeAsEndogenousProcess}
Suppose we model the velocities $X_i'$ of the positions $X_i$ of the masses between the walls in the damped coupled harmonic oscillator of Example~\ref{ex:HarmonicOscillatorIntervened} explicitly as separate endogenous processes $V_{i'}$. We could attempt to model this with an \CDM\ $\tilde{\C{R}}$ for which the \SDEfull s are given by $X_1 = 0$, $X_d = L$ and 
$$
  \left\{\begin{aligned}
     X_i &= \frac{\kappa_i}{\kappa_i + \kappa_{i-1}} (X_{i+1} - L_i) + \frac{\kappa_{i-1}}{\kappa_i + \kappa_{i-1}} (X_{i-1} + L_{i-1}) 
   -\frac{b_i}{\kappa_i + \kappa_{i-1}} V_{i'} - \frac{m_i}{\kappa_i + \kappa_{i-1}} V_{i'}' \\
     V_{i'} &= X_i'
 \end{aligned}\right.
$$
for $i=2,\dots,d-1$.
Performing a stochastic perfect intervention on both the position $X_i$ and the velocity $V_{i'}$ of one of the masses between the walls ($i\in\{2,\dots,d-1\}$) can lead to unphysical behavior. For example, the perfect intervention $\intervene(\{2,2'\},(0,1))$ gives an intervened \CDM\ with a solution that is physically impossible if we keep interpreting $X_i$ as the position and $V_{i'}$ the velocity of the $i^{\text{th}}$ mass.
\end{example}
This observation constitutes strong motivation for 
considering the higher-order derivatives $X_i^{(k_i)}$ (up to and including order $n_i$) to be aspects of the endogenous process $X_i$ rather than as ``causally independent'' processes. Thereby, we circumvent modeling velocity as the (instantaneous) cause of position~\citep[as in][]{IS94}, or the other way around. The resulting modeling framework appears more natural than that of \citep{MJS13}, which is explicitly limited to first-order dynamics and cannot accommodate systems like the damped harmonic oscillator as easily as \CDM s can, as it has to impose restrictions on the possible interventions to deal with this problem.

The higher-order derivatives $X_i^{(k_i)}$ do not always exist for a process $X_i$. For example, if we force the mass $m_1$ to follow a Brownian motion\footnote{A stochastic process $B$ on $T=[0,\infty)$ is called a \emph{Brownian motion} if: (i) $B_0 = 0$; (ii) $B$ has independent, stationary increments; (iii) $B_t \sim \C{N}(0,t)$ for all $t > 0$; (iv) $B$ is continuous. In particular, $B$ is not differentiable~\citep[see, for example, Theorem 21.17 in][]{Kle14}.
} $K_1$ in 
the spring model $\C{R}$ of Example~\ref{ex:TwoHarmonicOscillatorAmbiguousCausalInterpretation}, then the intervened model $\C{R}_{\intervene(1,K_1)}$ does not yield a solution (because $X_1''$ needs to exist and be continuous, which is not the case for $X_1 = K_1$).
In practice, we therefore only consider stochastic perfect interventions $\intervene(I,\B{K}_I)$ for which $\B{K}_I$ is a $C^{\B{n}_I}$-stochastic process.

\subsection{Graphs}
\label{sec:Graph}

We will now define a graphical representation of the structural properties of \CDM s that is inspired by the graphical representation of SCMs \citep{Pea09,BPSM19}. Where the graph of an SCM describes the functional relationships between the random variables encoded by the structural equations, the graph of an \CDM\ expresses the functional dependencies between the stochastic processes encoded by the \SDEfull s.


Typically, for $i \in \C{I}$, the component $f_i$ of the dynamic causal mechanism $\B{f}$ only depends on a subset of the (derivatives of the) endogenous and exogenous processes that we call the \emph{functional parents of $i$}.
\begin{definition}[Functional and integrated parents]
\label{def:FunctionalAndIntegratedParents}
Let $\C{R} = \langle \C{I}, \C{J}, \BC{X}, \BC{E}, \B{n}, \B{f}, \B{E} \rangle$ be an \CDM. For $k\in\widebar{\C{I}}^{(\B{n})}\cup\C{J}$ and $i\in\widebar{\C{I}}^{(\B{n})}$, we call
\begin{enumerate}
  \item $k$ a \emph{functional parent of} $i$ if and only if $i\in\C{I}$ and there does not exist a measurable function\footnote{For $\BC{X}^{\B{n}+1}=\prod_{i^{(k_i)}\in\widebar{\C{I}}^{(\B{n})}}\C{X}_i$, some subset $I\subseteq\widebar{\C{I}}^{(\B{n})}$ and $k\in\widebar{\C{I}}^{(\B{n})}$, we denote $(\BC{X}^{\B{n}+1})_{\setminus I}=\prod_{i^{(k_i)}\in\widebar{\C{I}}^{(\B{n})}\setminus I}\C{X}_i$ and $(\BC{X}^{\B{n}+1})_{\setminus k}=\prod_{i^{(k_i)}\in\widebar{\C{I}}^{(\B{n})}\setminus\{k\}}\C{X}_i$, and similarly for their elements.} $\tilde{f}_i:(\BC{X}^{\B{n}+1})_{\setminus k}\times\BC{E}_{\setminus k}\to\C{X}_i$ such that for all $\B{e}\in\BC{E}$ and for all $\widebar{\B{x}}^{(\B{n})}\in\BC{X}^{\B{n}+1}$
$$
    x_i = f_i(\widebar{\B{x}}^{(\B{n})},\B{e})  \quad\iff\quad x_i = \tilde{f}_i((\widebar{\B{x}}^{(\B{n})})_{\setminus k},\B{e}_{\setminus k}) \,;
$$
\item $k$ an \emph{integrated parent of} $i$ if and only if there exists an $\ell\in\C{I}$ such that $k=\ell^{(m_{\ell}-1)}$ and $i=\ell^{(m_{\ell})}$ for some $0< m_{\ell} \leq n_{\ell}$.
\end{enumerate}
\end{definition}

%
Exogenous processes have no functional and integrated parents by definition. The integrated parents denote the differential relationships that are satisfied by the endogenous processes. That is, for every $\ell\in\C{I}$ and $0<m_{\ell}\leq n_{\ell}$ we have that $\ell^{(m_{\ell}-1)}$ is an integrated parent of $\ell^{(m_{\ell})}$, which represents the differential relationship
$$
X_{\ell}^{(m_{\ell})} = \frac{d}{dt}X_{\ell}^{(m_{\ell}-1)} \,. 
$$
These differential relationships are absent for SCMs, because the endogenous variables are considered static. In contrast to \citep{IS94}, we express the differential relationships between the endogenous processes by the derivative operator, instead of the integration operator. In general, the integration operator of \citep{IS94} is not uniquely defined, since for a particular process there may exist several integrated processes differing by a (possibly random) integration constant. The derivative of a process, however, is always a.s.\ uniquely defined, if it exists. Hence, for a solution $\B{X}$ of an \CDM\ we can always derive the higher-order derivatives of $X_i$ up to order $n_i$ by repeatedly applying the derivative operator. In this way, we can consider the complete $n_i^{\text{th}}$-order derivative $\widebar{X}_i^{(n_i)}$ to encode aspects of the same endogenous process $X_i$.

The different parental relations can be expressed in a clustered mixed graph, where each cluster represents a complete $n_i^{\text{th}}$-order derivative.
\begin{figure}
  \begin{center}
   \adjustbox{scale=0.9}{%
    \begin{tikzpicture}
      \node[text=black!60] (Qn1) at (1.5,-3.55) {\small $\widebar{X}^{(n_1)}_1$};
      \node[text=black!60] (Qn2) at (3.0,-3.55) {\small $\widebar{X}^{(n_2)}_2$};
      \node[text=black!60] (Qn3) at (4.5,-3.55) {\small $\widebar{X}^{(n_3)}_3$};
      \node[text=black!60] (Qn4) at (6.0,-3.55) {\small $\widebar{X}^{(n_4)}_4$};
      \node[text=black!60] (Qn5) at (7.5,-3.55) {\small $\widebar{X}^{(n_5)}_5$};
      \node[var] (Q1) at (1.5,0) {$X_1$};
      \node[var] (Q2) at (3.0,0) {$X_2$};
      \node[var] (Q3) at (4.5,0) {$X_3$};
      \node[var] (Q4) at (6.0,0) {$X_4$};
      \node[var] (Q5) at (7.5,0) {$X_5$};
      \node[exvar] (L1) at (2.25,1.4) {$L_1$};
      \node[exvar] (L2) at (3.75,1.4) {$L_2$};
      \node[exvar] (L3) at (5.25,1.4) {$L_3$};
      \node[exvar] (L4) at (6.75,1.4) {$L_4$};
      \draw[draw=black!60] (1.75,1.9) rectangle (2.75,0.9);
      \draw[draw=black!60] (3.25,1.9) rectangle (4.25,0.9);
      \draw[draw=black!60] (4.75,1.9) rectangle (5.75,0.9);
      \draw[draw=black!60] (6.25,1.9) rectangle (7.25,0.9);
      \node[var] (dQ1) at (1.5,-1.4) {$X_1'$};
      \node[var] (dQ2) at (3.0,-1.4) {$X_2'$};
      \node[var] (dQ3) at (4.5,-1.4) {$X_3'$};
      \node[var] (dQ4) at (6.0,-1.4) {$X_4'$};
      \node[var] (dQ5) at (7.5,-1.4) {$X_5'$};
      \node[var] (ddQ1) at (1.5,-2.8) {$X_1''$};
      \node[var] (ddQ2) at (3.0,-2.8) {$X_2''$};
      \node[var] (ddQ3) at (4.5,-2.8) {$X_3''$};
      \node[var] (ddQ4) at (6.0,-2.8) {$X_4''$};
      \node[var] (ddQ5) at (7.5,-2.8) {$X_5''$};
      \draw[draw=black!60] (1.0,0.5) rectangle (2.0,-3.3);
      \draw[draw=black!60] (2.5,0.5) rectangle (3.5,-3.3);
      \draw[draw=black!60] (4.0,0.5) rectangle (5.0,-3.3);
      \draw[draw=black!60] (5.5,0.5) rectangle (6.5,-3.3);
      \draw[draw=black!60] (7.0,0.5) rectangle (8.0,-3.3);
      \draw[arr,dashed, bend left=10] (Q1) to (dQ1);
      \draw[arr,dashed, bend left=10] (Q2) to (dQ2);
      \draw[arr,dashed, bend left=10] (Q3) to (dQ3);
      \draw[arr,dashed, bend left=10] (Q4) to (dQ4);
      \draw[arr,dashed, bend left=10] (Q5) to (dQ5);
      \draw[arr,dashed, bend left=10] (dQ1) to (ddQ1);
      \draw[arr,dashed, bend left=10] (dQ2) to (ddQ2);
      \draw[arr,dashed, bend left=10] (dQ3) to (ddQ3);
      \draw[arr,dashed, bend left=10] (dQ4) to (ddQ4);
      \draw[arr,dashed, bend left=10] (dQ5) to (ddQ5);
      \draw[arr,bend left=15] (Q1) to (Q2);
      \draw[arr,bend left=15] (Q2) to (Q3);
      \draw[arr,bend left=15] (Q3) to (Q4);
      \draw[arr,bend left=15] (Q4) to (Q5);
      \draw[arr,bend left=15] (Q2) to (Q1);
      \draw[arr,bend left=15] (Q3) to (Q2);
      \draw[arr,bend left=15] (Q4) to (Q3);
      \draw[arr,bend left=15] (Q5) to (Q4);
      \draw[arr] (L1) to (Q2);
      \draw[arr] (L2) to (Q3);
      \draw[arr] (L3) to (Q4);
      \draw[arr] (L4) to (Q5);
      \draw[arr] (L1) to (Q1);
      \draw[arr] (L2) to (Q2);
      \draw[arr] (L3) to (Q3);
      \draw[arr] (L4) to (Q4);
      \draw[arr,dashed,bend left=10] (dQ1) to (Q1);
      \draw[arr,dashed,bend left=26] (ddQ1) to (Q1);
      \draw[arr,dashed,bend left=10] (dQ2) to (Q2);
      \draw[arr,dashed,bend left=26] (ddQ2) to (Q2);
      \draw[arr,dashed,bend left=10] (dQ3) to (Q3);
      \draw[arr,dashed,bend left=26] (ddQ3) to (Q3);
      \draw[arr,dashed,bend left=10] (dQ4) to (Q4);
      \draw[arr,dashed,bend left=26] (ddQ4) to (Q4);
      \draw[arr,dashed,bend left=10] (dQ5) to (Q5);
      \draw[arr,dashed,bend left=26] (ddQ5) to (Q5);
  \end{tikzpicture}}
  \hspace{1cm}
   \adjustbox{scale=0.9}{%
    \begin{tikzpicture}
      \node[text=black!60] (Qn1) at (1.5,-3.55) {\small $\widebar{X}^{(n_1)}_1$};
      \node[text=black!60] (Qn2) at (3.0,-3.55) {\small $\widebar{X}^{(n_2)}_2$};
      \node[text=black!60] (Qn3) at (4.5,-3.55) {\small $\widebar{X}^{(n_3)}_3$};
      \node[text=black!60] (Qn4) at (6.0,-3.55) {\small $\widebar{X}^{(n_4)}_4$};
      \node[text=black!60] (Qn5) at (7.5,-3.55) {\small $\widebar{X}^{(n_5)}_5$};
      \node[var] (Q1) at (1.5,0) {$X_1$};
      \node[var] (Q2) at (3.0,0) {$X_2$};
      \node[var] (Q3) at (4.5,0) {$X_3$};
      \node[var] (Q4) at (6.0,0) {$X_4$};
      \node[var] (Q5) at (7.5,0) {$X_5$};
      \node[var] (dQ1) at (1.5,-1.4) {$X_1'$};
      \node[var] (dQ2) at (3.0,-1.4) {$X_2'$};
      \node[var] (dQ3) at (4.5,-1.4) {$X_3'$};
      \node[var] (dQ4) at (6.0,-1.4) {$X_4'$};
      \node[var] (dQ5) at (7.5,-1.4) {$X_5'$};
      \node[var] (ddQ1) at (1.5,-2.8) {$X_1''$};
      \node[var] (ddQ2) at (3.0,-2.8) {$X_2''$};
      \node[var] (ddQ3) at (4.5,-2.8) {$X_3''$};
      \node[var] (ddQ4) at (6.0,-2.8) {$X_4''$};
      \node[var] (ddQ5) at (7.5,-2.8) {$X_5''$};
      \draw[draw=black!60] (1.0,0.5) rectangle (2.0,-3.3);
      \draw[draw=black!60] (2.5,0.5) rectangle (3.5,-3.3);
      \draw[draw=black!60] (4.0,0.5) rectangle (5.0,-3.3);
      \draw[draw=black!60] (5.5,0.5) rectangle (6.5,-3.3);
      \draw[draw=black!60] (7.0,0.5) rectangle (8.0,-3.3);
      \draw[arr,dashed, bend left=10] (Q1) to (dQ1);
      \draw[arr,dashed, bend left=10] (Q2) to (dQ2);
      \draw[arr,dashed, bend left=10] (Q3) to (dQ3);
      \draw[arr,dashed, bend left=10] (Q4) to (dQ4);
      \draw[arr,dashed, bend left=10] (Q5) to (dQ5);
      \draw[arr,dashed, bend left=10] (dQ1) to (ddQ1);
      \draw[arr,dashed, bend left=10] (dQ2) to (ddQ2);
      \draw[arr,dashed, bend left=10] (dQ3) to (ddQ3);
      \draw[arr,dashed, bend left=10] (dQ4) to (ddQ4);
      \draw[arr,dashed, bend left=10] (dQ5) to (ddQ5);
      \draw[arr,bend left=15] (Q1) to (Q2);
      \draw[arr,bend left=15] (Q2) to (Q3);
      \draw[arr,bend left=15] (Q3) to (Q4);
      \draw[arr,bend left=15] (Q4) to (Q5);
      \draw[arr,bend left=15] (Q2) to (Q1);
      \draw[arr,bend left=15] (Q3) to (Q2);
      \draw[arr,bend left=15] (Q4) to (Q3);
      \draw[arr,bend left=15] (Q5) to (Q4);
      \draw[biarr,bend left=45] (Q1) to (Q2);
      \draw[biarr,bend left=45] (Q2) to (Q3);
      \draw[biarr,bend left=45] (Q3) to (Q4);
      \draw[biarr,bend left=45] (Q4) to (Q5);
      \draw[arr,dashed,bend left=10] (dQ1) to (Q1);
      \draw[arr,dashed,bend left=26] (ddQ1) to (Q1);
      \draw[arr,dashed,bend left=10] (dQ2) to (Q2);
      \draw[arr,dashed,bend left=26] (ddQ2) to (Q2);
      \draw[arr,dashed,bend left=10] (dQ3) to (Q3);
      \draw[arr,dashed,bend left=26] (ddQ3) to (Q3);
      \draw[arr,dashed,bend left=10] (dQ4) to (Q4);
      \draw[arr,dashed,bend left=26] (ddQ4) to (Q4);
      \draw[arr,dashed,bend left=10] (dQ5) to (Q5);
      \draw[arr,dashed,bend left=26] (ddQ5) to (Q5);
  \end{tikzpicture}}
  \end{center}
  \caption{Augmented graph (left) and graph (right) of the damped coupled harmonic oscillator model $\C{R}$ of Example~\ref{ex:HarmonicOscillator} for $d=5$.}
  \label{fig:HarmonicOscillatorGraphs}
\end{figure}

\begin{definition}[Graph and augmented graph]
\label{def:GraphCDM}
Let $\C{R} = \langle \C{I}, \C{J}, \BC{X}, \BC{E}, \B{n}, \B{f}, \B{E} \rangle$ be an \CDM\ with order tuple $\B{n}$. We define:
  \begin{enumerate}
    \item the \emph{augmented graph} $\C{G}^a(\C{R})$ of $\C{R}$ as the clustered mixed graph with nodes $\widebar{\C{I}}^{(\B{n})}\cup\C{J}$ partitioned into \emph{clusters} $\bar{i}^{(n_i)}=\{ i^{(k_i)} : 0\leq k_i \leq n_i\}$ for $i\in\C{I}$ and clusters $\{j\}$ for $j\in\C{J}$, directed edges $k\gto l$ if and only if $k$ is functional parent of $l$ in a different cluster, dashed directed edges $k\gtod l$ if and only if $k$ is a functional or integrated parent of $l$ in the same cluster;
    \item the \emph{graph} $\C{G}(\C{R})$ of $\C{R}$ as the clustered mixed graph with nodes $\widebar{\C{I}}^{(\B{n})}$ partitioned into \emph{clusters} $\bar{i}^{(n_i)}=\{ i^{(k_i)} : 0\leq k_i \leq n_i\}$ for $i\in\C{I}$, directed edges $k\gto l$ if and only if $k$ is functional parent of $l$ in a different cluster, dashed directed edges $k\gtod l$ if and only if $k$ is a functional or integrated parent of $l$ in the same cluster, and bidirected edges $k\goto l$ if and only if there exists a $j\in\C{J}$ that is a functional parent of both $k$ and $l$.
  \end{enumerate}
\end{definition}

The augmented graph differs from the graph in that it gives an explicit representation of the exogenous processes rather than an implicit one using bidirected edges. The augmented graph contains no directed edge pointing towards an exogenous process node. 
The clusters $\bar{i}^{(n_i)}\in\widebar{\C{I}}^{(\B{n})}$ for $i \in \C{I}$ and $\{j\}$ for $j\in\C{J}$ of the (augmented) graph of an \CDM\ refer to the complete $n_i^{\text{th}}$-order derivative $\widebar{X}_i^{(n_i)}$ and $E_j$ respectively, and are represented by a box around the nodes of the cluster. The graph and augmented graph are illustrated\footnote{For visualizing the graphs we stick to the common convention of using stochastic processes and random variables with the index as a subscript, instead of using the indices themselves (even when no solutions are defined).} in Figure~\ref{fig:HarmonicOscillatorGraphs} for the damped coupled harmonic oscillator model of Example~\ref{ex:HarmonicOscillator}, where the white and gray nodes represent the endogenous and exogenous processes, respectively. Between the nodes of different clusters there are only functional parental relations. Within a cluster, the higher-order derivatives $i^{(k_i)}$ for $k_i>0$ of the endogenous processes $i\in\C{I}$ have no functional parents, but have only integrated parents. However, any node $i^{(k_i)}$ with $k_i > 0$ may be a functional parent of another node $j\in\C{I}$; see, for example, the graph of the \CDM\ $\tilde{\C{R}}$ in Example~\ref{ex:TwoHarmonicOscillatorAmbiguousCausalInterpretation}.\footnote{A more realistic example could be Faraday's law of induction. In terms of individual point charges: a moving point charge generates a magnetic field, which exerts a force on some other point charge that is proportional to the velocity of the moving point charge.}

In particular, this definition of the (augmented) graph of an \CDM\ reduces to the usual notion of the (augmented) graph of an SCM if we ignore the clusters. Indeed, the graph $\C{G}(\C{M})$ of an SCM $\C{M}=\langle \C{I}, \C{J}, \BC{X}, \BC{E}, \B{f}, \B{E} \rangle$ is a mixed graph with nodes $\C{I}$, directed edges $i\gto j$ if and only if $i$ is a functional parent of $j$ with $i\neq j$, dashed directed edge $i\gtod i$ if and only if $i$ is a functional parent of itself, and bidirected edges $i\goto j$ if and only if there exists a $k\in\C{J}$ that is a functional parent of both $i$ and $j$, where we apply Definition~\ref{def:FunctionalAndIntegratedParents} of a functional parent to $\C{M}$ (note that by definition, an SCM has no integrated parents). The augmented graph $\C{G}^{a}(\C{M})$ of an SCM $\C{M}$ is defined analogously, but the bidirected edges are replaced by exogenous nodes in $\C{J}$ with outgoing directed edges to their functional children.

On the graphs of an \CDM, the operation of a stochastic perfect intervention acts in a simple way.
\begin{proposition}[Graphs of the intervened \CDM]
  Let $\C{R}$ be an \CDM\ and $\intervene(I,\B{K}_I)$ a stochastic perfect intervention for $I\subseteq\C{I}$ a subset and $\B{K}_I$ an independent stochastic process. The graph $\C{G}(\C{R}_{\intervene(I,\B{K}_I)})$ of the intervened \CDM\ $\C{R}_{\intervene(I,\B{K}_I)}$ is the graph $\C{G}(\C{R})$, but without the edges that have an arrowhead pointing towards a node in the intervention target set $I$.
A similar statement holds for the augmented graph $\C{G}^a(\C{R}_{\intervene(I,\B{K}_I)})$.
\end{proposition}

\begin{figure}
  \begin{center}
  \adjustbox{scale=0.9}{%
    \begin{tikzpicture}
      \node[text=black!60] (Qn1) at (1.5,-3.55) {\small $\widebar{X}^{(n_1)}_1$};
      \node[text=black!60] (Qn2) at (3.0,-3.55) {\small $\widebar{X}^{(n_2)}_2$};
      \node[text=black!60] (Qn3) at (4.5,-3.55) {\small $\widebar{X}^{(n_3)}_3$};
      \node[text=black!60] (Qn4) at (6.0,-3.55) {\small $\widebar{X}^{(n_4)}_4$};
      \node[text=black!60] (Qn5) at (7.5,-3.55) {\small $\widebar{X}^{(n_5)}_5$};
      \node[var] (Q1) at (1.5,0) {$X_1$};
      \node[var] (Q2) at (3.0,0) {$X_2$};
      \node[var] (Q3) at (4.5,0) {$X_3$};
      \node[var] (Q4) at (6.0,0) {$X_4$};
      \node[var] (Q5) at (7.5,0) {$X_5$};
      \node[exvar] (L1) at (2.25,1.4) {$L_1$};
      \node[exvar] (L2) at (3.75,1.4) {$L_2$};
      \node[exvar] (L3) at (5.25,1.4) {$L_3$};
      \node[exvar] (L4) at (6.75,1.4) {$L_4$};
      \draw[draw=black!60] (1.75,1.9) rectangle (2.75,0.9);
      \draw[draw=black!60] (3.25,1.9) rectangle (4.25,0.9);
      \draw[draw=black!60] (4.75,1.9) rectangle (5.75,0.9);
      \draw[draw=black!60] (6.25,1.9) rectangle (7.25,0.9);
      \draw[arr] (L1) to (Q2);
      \draw[arr] (L2) to (Q3);
      \draw[arr] (L3) to (Q4);
      \draw[arr] (L2) to (Q2);
      \draw[arr] (L3) to (Q3);
      \draw[arr] (L4) to (Q4);
      \node[var] (dQ1) at (1.5,-1.4) {$X_1'$};
      \node[var] (dQ2) at (3.0,-1.4) {$X_2'$};
      \node[var] (dQ3) at (4.5,-1.4) {$X_3'$};
      \node[var] (dQ4) at (6.0,-1.4) {$X_4'$};
      \node[var] (dQ5) at (7.5,-1.4) {$X_5'$};
      \node[var] (ddQ1) at (1.5,-2.8) {$X_1''$};
      \node[var] (ddQ2) at (3.0,-2.8) {$X_2''$};
      \node[var] (ddQ3) at (4.5,-2.8) {$X_3''$};
      \node[var] (ddQ4) at (6.0,-2.8) {$X_4''$};
      \node[var] (ddQ5) at (7.5,-2.8) {$X_5''$};
      \draw[draw=black!60] (1.0,0.5) rectangle (2.0,-3.3);
      \draw[draw=black!60] (2.5,0.5) rectangle (3.5,-3.3);
      \draw[draw=black!60] (4.0,0.5) rectangle (5.0,-3.3);
      \draw[draw=black!60] (5.5,0.5) rectangle (6.5,-3.3);
      \draw[draw=black!60] (7.0,0.5) rectangle (8.0,-3.3);
      \draw[arr,dashed] (Q1) to (dQ1);
      \draw[arr,dashed, bend left=10] (Q2) to (dQ2);
      \draw[arr,dashed, bend left=10] (Q3) to (dQ3);
      \draw[arr,dashed, bend left=10] (Q4) to (dQ4);
      \draw[arr,dashed] (Q5) to (dQ5);
      \draw[arr,dashed] (dQ1) to (ddQ1);
      \draw[arr,dashed, bend left=10] (dQ2) to (ddQ2);
      \draw[arr,dashed, bend left=10] (dQ3) to (ddQ3);
      \draw[arr,dashed, bend left=10] (dQ4) to (ddQ4);
      \draw[arr,dashed] (dQ5) to (ddQ5);
      \draw[arr] (Q1) to (Q2);
      \draw[arr,bend left=15] (Q2) to (Q3);
      \draw[arr,bend left=15] (Q3) to (Q4);
      \draw[arr,bend left=15] (Q3) to (Q2);
      \draw[arr,bend left=15] (Q4) to (Q3);
      \draw[arr] (Q5) to (Q4);
      \draw[arr,dashed,bend left=10] (dQ2) to (Q2);
      \draw[arr,dashed,bend left=26] (ddQ2) to (Q2);
      \draw[arr,dashed,bend left=10] (dQ3) to (Q3);
      \draw[arr,dashed,bend left=26] (ddQ3) to (Q3);
      \draw[arr,dashed,bend left=10] (dQ4) to (Q4);
      \draw[arr,dashed,bend left=26] (ddQ4) to (Q4);
  \end{tikzpicture}}
  \hspace{1cm}
  \adjustbox{scale=0.9}{%
    \begin{tikzpicture}
      \node[text=black!60] (Qn1) at (1.5,-3.55) {\small $\widebar{X}^{(n_1)}_1$};
      \node[text=black!60] (Qn2) at (3.0,-3.55) {\small $\widebar{X}^{(n_2)}_2$};
      \node[text=black!60] (Qn3) at (4.5,-3.55) {\small $\widebar{X}^{(n_3)}_3$};
      \node[text=black!60] (Qn4) at (6.0,-3.55) {\small $\widebar{X}^{(n_4)}_4$};
      \node[text=black!60] (Qn5) at (7.5,-3.55) {\small $\widebar{X}^{(n_5)}_5$};
      \node[var] (Q1) at (1.5,0) {$X_1$};
      \node[var] (Q2) at (3.0,0) {$X_2$};
      \node[var] (Q3) at (4.5,0) {$X_3$};
      \node[var] (Q4) at (6.0,0) {$X_4$};
      \node[var] (Q5) at (7.5,0) {$X_5$};
      \node[var] (dQ1) at (1.5,-1.4) {$X_1'$};
      \node[var] (dQ2) at (3.0,-1.4) {$X_2'$};
      \node[var] (dQ3) at (4.5,-1.4) {$X_3'$};
      \node[var] (dQ4) at (6.0,-1.4) {$X_4'$};
      \node[var] (dQ5) at (7.5,-1.4) {$X_5'$};
      \node[var] (ddQ1) at (1.5,-2.8) {$X_1''$};
      \node[var] (ddQ2) at (3.0,-2.8) {$X_2''$};
      \node[var] (ddQ3) at (4.5,-2.8) {$X_3''$};
      \node[var] (ddQ4) at (6.0,-2.8) {$X_4''$};
      \node[var] (ddQ5) at (7.5,-2.8) {$X_5''$};
      \draw[draw=black!60] (1.0,0.5) rectangle (2.0,-3.3);
      \draw[draw=black!60] (2.5,0.5) rectangle (3.5,-3.3);
      \draw[draw=black!60] (4.0,0.5) rectangle (5.0,-3.3);
      \draw[draw=black!60] (5.5,0.5) rectangle (6.5,-3.3);
      \draw[draw=black!60] (7.0,0.5) rectangle (8.0,-3.3);
      \draw[arr,dashed] (Q1) to (dQ1);
      \draw[arr,dashed, bend left=10] (Q2) to (dQ2);
      \draw[arr,dashed, bend left=10] (Q3) to (dQ3);
      \draw[arr,dashed, bend left=10] (Q4) to (dQ4);
      \draw[arr,dashed] (Q5) to (dQ5);
      \draw[arr,dashed] (dQ1) to (ddQ1);
      \draw[arr,dashed, bend left=10] (dQ2) to (ddQ2);
      \draw[arr,dashed, bend left=10] (dQ3) to (ddQ3);
      \draw[arr,dashed, bend left=10] (dQ4) to (ddQ4);
      \draw[arr,dashed] (dQ5) to (ddQ5);
      \draw[arr] (Q1) to (Q2);
      \draw[arr,bend left=15] (Q2) to (Q3);
      \draw[arr,bend left=15] (Q3) to (Q4);
      \draw[arr,bend left=15] (Q3) to (Q2);
      \draw[arr,bend left=15] (Q4) to (Q3);
      \draw[arr] (Q5) to (Q4);
      \draw[biarr,bend left=45] (Q2) to (Q3);
      \draw[biarr,bend left=45] (Q3) to (Q4);
      \draw[arr,dashed,bend left=10] (dQ2) to (Q2);
      \draw[arr,dashed,bend left=26] (ddQ2) to (Q2);
      \draw[arr,dashed,bend left=10] (dQ3) to (Q3);
      \draw[arr,dashed,bend left=26] (ddQ3) to (Q3);
      \draw[arr,dashed,bend left=10] (dQ4) to (Q4);
      \draw[arr,dashed,bend left=26] (ddQ4) to (Q4);
  \end{tikzpicture}}
  \end{center}
  \caption{Augmented graph (left) and graph (right) of the intervened damped coupled harmonic oscillator model $\C{R}_{\intervene(\{1,5\},(0,L))}$ of Example~\ref{ex:HarmonicOscillatorIntervened} for $d=5$.}
  \label{fig:HarmonicOscillatorIntervenedGraphs}
\end{figure}
The graph and augmented graph of the damped coupled harmonic oscillator model of Example~\ref{ex:HarmonicOscillatorIntervened}, where we performed the perfect intervention of fixing the endpoint masses to the walls, are illustrated in Figure~\ref{fig:HarmonicOscillatorIntervenedGraphs}. Performing a stochastic perfect intervention on an endogenous process removes all the (bi-)directed edges that point towards the intervened process, including the dashed directed edges within the cluster. The dashed directed edges within the cluster that correspond to the integrated parents, that is, those pointing to a higher-order derivative, indicate that the higher-order derivatives of the intervened endogenous process need to exist for any solution of the model. Hence, we view a stochastic perfect intervention on an endogenous process as an intervention on the whole cluster of the intervened process. We say that there is a \emph{directed edge from cluster $I$ to cluster $J$} if there exists a directed edge from some $i\in I$ to some $j\in J$. Since a stochastic perfect intervention can be seen as an intervention on the entire associated cluster, the directed edges between the clusters express the direct causal relationships between the clusters.
We call a dashed directed edge $i\gtod i$ in the graph of an \CDM\  (that is, where $i$ is a functional parent of itself) a \emph{self-cycle at $i$}. An example of a model where a self-cycle arises is the well-known market equilibrium model from economics, which has been thoroughly discussed in the literature \citep[see, for example,][]{RichardsonRobins2014}.
\begin{example}[Price, supply and demand]
\label{ex:SupplyDemandModel}
Let $X_P$ denote the price, $X_S$ denote the supply and $X_D$ the demand of a quantity of a product. The following \SDEfull s specify an \CDM\ $\C{R}$ that describes how the
demanded and supplied quantities are determined by the price, and how price
adjustments occur in the market:
$$
  \left\{\begin{aligned}
    X_P  &= X_P + \lambda (X_D - X_S) - X_P' \\
    X_S  &= \beta_S X_P + E_S \\
    X_D  &= \beta_D X_P + E_D \,,
  \end{aligned}\right.
$$
where $\B{n}:=(n_P,n_S,n_D)=(1,0,0)$. Here, $E_S$ and $E_D$ are the exogenous influences on the supply and demand respectively, $\beta_S>0$ is the reciprocal of the slope of the supply curve, $\beta_D<0$ is the reciprocal of the slope of the demand curve, and $\lambda>0$ models how fast the price adjusts to market conditions. The graph of this model is depicted in Figure~\ref{fig:SupplyDemandExample} (left) and contains a self-cycle at $P$.
\begin{figure}
  \centering
  \adjustbox{scale=0.9}{%
    \begin{tikzpicture}[baseline={([yshift={-\ht\strutbox}]current bounding box.north)}]
      \node[var] (D) at (1.5,1.5) {$X_D$};
      \node[var] (S) at (-1.5,1.5) {$X_S$};
      \node[var] (P) at (0,1.5) {$X_P$};
      \node[var] (dP) at (0,0.1) {$X_P'$};
      \draw[draw=black!60] (-0.45,-0.35) rectangle (0.45,1.95);
      \draw[draw=black!60] (-1.95,1.05) rectangle (-1.05,1.95);
      \draw[draw=black!60] (1.05,1.05) rectangle (1.95,1.95);
      \draw[arr, bend left=20] (P) edge (S);
      \draw[arr, bend left=20] (P) edge (D);
      \draw[arr, bend left=20] (S) edge (P);
      \draw[arr, bend left=20] (D) edge (P);
      \draw[arr,dashed] (P)  to [out=60,in=120,looseness=5] (P);
      \draw[arr, dashed, bend left=15] (dP) to (P);
      \draw[arr,dashed, bend left=15] (P) to (dP);
      \node[text=black!60] at (0,-0.65) {$\widebar{X}^{(n_P)}_P$};
      \node[text=black!60] at (-1.5,0.75) {$\widebar{X}^{(n_S)}_S$};
      \node[text=black!60] at (1.5,0.75) {$\widebar{X}^{(n_D)}_D$};
      \node at (-2.9,1.5) {$\C{G}(\C{R})$:};
    \end{tikzpicture}}
  \hspace{1cm}
  \adjustbox{scale=0.9}{%
    \begin{tikzpicture}[baseline={([yshift={-\ht\strutbox}]current bounding box.north)}]
      \node[var] (D) at (1.5,1.5) {$X_D$};
      \node[var] (S) at (-1.5,1.5) {$X_S$};
      \node[var] (P) at (0,1.5) {$X_P$};
      \draw[arr, bend left=20] (P) edge (S);
      \draw[arr, bend left=20] (P) edge (D);
      \draw[arr, bend left=20] (S) edge (P);
      \draw[arr, bend left=20] (D) edge (P);
      \draw[arr,dashed] (P)  to [out=60,in=120,looseness=5] (P);
      \node at (-2.9,1.5) {$\C{G}(\C{M}_{\C{R}})$:};
    \end{tikzpicture}}
    \caption{Graphs of the price, supply and demand model $\C{R}$ (left) of Example~\ref{ex:SupplyDemandModel} and the corresponding equilibrated model $\C{M}_{\C{R}}$ (right) of Example~\ref{ex:SupplyDemandModelGraph}.}
  \label{fig:SupplyDemandExample}
\end{figure}
\end{example}

We already encountered several instances of \emph{linear} \CDM s (for example, in Examples~\ref{ex:HarmonicOscillator} and \ref{ex:SupplyDemandModel}).
\begin{definition}[Linear \CDM]
\label{def:linearCDM}
We call an \CDM\ $\C{R}$ \emph{linear}, if the dynamic causal mechanism $\B{f}:\BC{X}^{\B{n}+1}\times\BC{E}\to\BC{X}$ is of the form 
$$
  \B{f}(\widebar{\B{x}}^{(\B{n})},\B{e}) := B\widebar{\B{x}}^{(\B{n})} + \Gamma\B{e} \,,
$$
where $B\in\RN^{\C{I}\times\widebar{\C{I}}^{(\B{n})}}$ and $\Gamma\in\RN^{\C{I}\times\C{J}}$ are matrices.
\end{definition}
For a linear \CDM\ $\C{R}$, a nonzero coefficient $B_{ik}$ for $i,k\in\widebar{\C{I}}^{(\B{n})}$ such that $i\neq k$ corresponds to a directed edge $k\gto i$ in the graph $\C{G}(\C{R})$ (and augmented graph $\C{G}^a(\C{R})$) if $i$ lies in a different cluster than $k$, and a dashed directed edge $k\gtod i$ if $i$ lies in the same cluster as $k$. A coefficient $B_{ii}=1$ for $i\in\C{I}$ corresponds to a self-cycle $i\gtod i$. There is a bidirected edge $i\goto k$ in the graph $\C{G}(\C{R})$ for $i,k\in\C{I}$ with $i\neq k$ if and only if there exists a $j\in\C{J}$ for which $\Gamma_{ij} \neq 0$ and $\Gamma_{kj}\neq 0$.
In the augmented graph $\C{G}^a(\C{R})$, there is a directed edge $j \gto i$ for $i \in \C{I}$, $j \in \C{J}$ if and only if 
$\Gamma_{ij} \neq 0$.

\subsection{Initial conditions}
\label{sec:InitialConditions}

In contrast to RDEs, \CDM s allow for both zeroth and higher-order differential equations.
For this reason, the \SDE s of \CDM s admit problems that can be quite different from those of RDEs. For example, the order of the initial conditions for \CDM s does not directly relate to the order of the \CDM.
\begin{definition}[Initial condition]
\label{def:InitialCondition}
Let $\C{R}$ be an \CDM, $I \subseteq \C{I}$ a subset of the endogenous variables, $\B{m}_I=(m_i)_{i\in I}\in\NN_0^{I}$ an order tuple, $t_0\in T$ and $\widebar{\B{X}}^{(\B{m}_I)}_{I,[0]}$ a random variable taking values in $\BC{X}_I^{\B{m}_I+1}$. 
We say that a solution $\B{X}$ of $\C{R}$ has \emph{initial condition $(t_0,\widebar{\B{X}}^{(\B{m}_I)}_{I,[0]})$} if
$\widebar{\B{X}}_I^{(\B{m}_I)}(t_0)$ exists and satisfies
\begin{equation*}
  \widebar{\B{X}}_I^{(\B{m}_I)}(t_0)= \widebar{\B{X}}^{(\B{m}_I)}_{I,[0]}
\end{equation*}
almost surely. Here, $\B{m}_I$ is called the \emph{order} of the initial condition; for $I = \C{I}$ we also refer to the
initial condition as a \emph{full initial condition}, and for $I \subsetneq \C{I}$ as a \emph{partial initial condition}.
A solution $\B{X}$ of $\C{R}$ with initial condition $(t_0,\widebar{\B{X}}^{(\B{m}_I)}_{I,[0]})$ is called \emph{almost surely unique} if for every solution $\B{Y}$ of $\C{R}$ with initial condition $(t_0,\widebar{\B{X}}^{(\B{m}_I)}_{I,[0]})$ we have $\B{X}=\B{Y}$ a.s..
\end{definition}

For an \CDM\ for which the \SDE s can be rewritten into the form of a system of $n_i^{\text{th}}$-order RDEs (with all $n_i \ge 1$), the full initial conditions of order $\B{n}-1$ of the \CDM\ correspond exactly with the usually considered initial conditions of this system of RDEs. For example, the solutions of the damped coupled harmonic oscillator of Example~\ref{ex:HarmonicOscillator} can be a.s.\ uniquely determined by the full initial conditions of order $\B{n}-1$ (see also Corollary~\ref{cor:ExistenceAndUniquenessLinearCase}). In general, however, the solutions of an \CDM\ may not be a.s.\ uniquely determined by the full initial conditions of order $\B{n}-1$, as the following example illustrates.
\begin{example}[The order of the \CDM\ and of the initial conditions]
\label{ex:HigherOrderRandomInitialConditions}
Let $\C{R} = \langle \{1\},\emptyset, \C{X},\C{E}, n, f, E \rangle$ be the \CDM\ with $\C{X}=\RN$, $\C{E}=\{*\}$, $n=1$, the dynamic causal mechanism $f:\C{X}^2\times\C{E}\to\C{X}$ given by $f(\widebar{x}^{(1)},e)=x-x^2+(x')^2$, and $E$ the trivial exogenous process. The \SDEfull\ associated to $\C{R}$ reads
$$
X = X - X^2 + (X')^2 \,. 
$$
This \SDE\ cannot be written as a (first-order) RDE of the form~\eqref{eq:RDE}, since it cannot be a.s.\ uniquely solved for $X'$. ``Solving for'' $X'$ leads to two RDEs that are of the form~\eqref{eq:RDE}, namely
$$
X' = X \quad\text{or}\quad X'=-X \,. 
$$
  The solutions of these RDEs are given by $X_t = X_{[0]} e^t$ and $X_t = X_{[0]} e^{-t}$ respectively, where $(0,X_{[0]})$ denotes the initial condition for both RDEs. These processes are also solutions of the \CDM, and one can show that all (continuously differentiable) solutions of $\C{R}$ are of this form. Note that, in principle, we could well have taken the order $n$ arbitrarily high without restricting the set of solutions, because the solutions are $C^{\infty}$-stochastic processes. If we consider the solutions of $\C{R}$ with an initial condition $(0,\widebar{X}_{[0]}^{(0)})$ of order $0$, then there are always two solutions with this initial condition that are not a.s.\ equal to each other, unless $X^{(0)}_{[0]}=0$. For the initial condition $(0,\widebar{X}_{[0]}^{(1)})$ of order $1$, we can specify the solution $X$ a.s.\ uniquely, if it exists. Take for example $\widebar{X}^{(1)}_{[0]}=(X_{[0]},X_{[0]})$, then the solution $X$ with this initial condition is a.s.\ uniquely given by $X_t = X_{[0]} e^t$. However, an arbitrary initial condition $(0,\widebar{X}_{[0]}^{(m)})$ of order $m$ greater or equal to $1$ may well be inconsistent with the \SDEfull s. For example, the initial condition $\widebar{X}_{[0]}^{(1)} = (X_{[0]},2X_{[0]})$ will not have a solution for $X_{[0]} \ne 0$, since the initial condition $\widebar{X}_{[0]}^{(1)}:=(X_{[0]}^{(0)},X_{[0]}^{(1)})$ does not satisfy $(X_{[0]}^{(0)})^2=(X_{[0]}^{(1)})^2$. 
\end{example}
This example illustrates that an arbitrary imposed initial condition may well be inconsistent with the \SDEfull s.
\begin{definition}[Consistent initial condition]
\label{def:ConsistentInitialCondition}
Let $\C{R}$ be an \CDM\ and $\B{m}_I=(m_i)_{i\in I}\in\NN_0^{I}$ an order tuple for $I \subseteq \C{I}$. We call an initial condition $(t_0,\widebar{\B{X}}^{(\B{m}_I)}_{I,[0]})$ for $\C{R}$ \emph{consistent} if there exists a solution of $\C{R}$ with this initial condition.
\end{definition}

In other words, for an initial condition there only exists a solution if and only if the initial condition is consistent. In particular, zeroth-order \SDEfull s may constrain the initial conditions (of any order) for which a solution exists.
\begin{example}[Zeroth-order \SDEfull\ constraint]
\label{ex:InconsistentInitialConditions}
Consider the price, supply and demand model $\C{R}$ of Example~\ref{ex:SupplyDemandModel} that has order tuple $\B{n}=(n_P,n_S,n_D)=(1,0,0)$. The zeroth-order \SDEfull s of $\C{R}$ are those associated with the supply $X_S$ and the demand $X_D$ processes. Since the solutions of $\C{R}$ satisfy these zeroth-order \SDEfull s almost surely at every point in time, the consistent full initial conditions $(t_0, \widebar{\B{X}}^{(m)}_{[0]})$ also need to satisfy the zeroth-order \SDEfull s almost surely, that is, $X_{[0],S} = \beta_SX_{[0],P} + (E_S)_{t_0}$ and $X_{[0],D} = \beta_DX_{[0],P} + (E_D)_{t_0}$ almost surely.
\end{example}

By definition, the consistent full initial conditions always need to satisfy the zeroth-order \SDEfull s of the \CDM. Initial conditions of an order greater than or equal to the order of the \CDM\ need to satisfy the corresponding \SDEfull s of the \CDM, as we already saw in Example~\ref{ex:HigherOrderRandomInitialConditions}. Additionally, in general, \CDM s that have higher-order \SDEfull s may contain \emph{hidden} constraints\footnote{We refer the reader to the literature on differential-algebraic equations for more details on this, for example, \citep{AP98}.} as the following example illustrates.
\begin{example}[Hidden constraint]
\label{ex:HiddenConstraint}
Consider the \CDM\ $\C{R} = \langle \{1,2\}, \{3\}, \RN^2, \RN, \B{n}, \B{f}, E \rangle$ with $\B{n}=(0,1)$, the dynamic causal mechanism $\B{f}:\BC{X}^{\B{n}+1}\times\C{E}\to\BC{X}$ given by $f_1(\widebar{\B{x}}^{(\B{n})},e):=x'_2$ and $f_2(\widebar{\B{x}}^{(\B{n})},e):=e$, and $E:=E_3$ some exogenous process. The \SDEfull s associated to $\C{R}$ read
$$
  \left\{\begin{aligned}
    X_1 &= X'_2 \\
    X_2 &= E \,.
  \end{aligned}\right. 
$$
This model cannot be written as an RDE,\footnote{Observe that a higher-order RDE is of the form $\B{X}^{(\B{n})} = \B{g}(\widebar{\B{X}}^{(\B{n}-1)},\B{E})$ for some measurable function $\B{g}:\BC{X}^{\B{n}}\times\BC{E}\to\BC{X}$ and stochastic process $\B{E}:T\times\Omega\to\BC{E}$.} since the Jacobian matrix
$$
\frac{\partial\B{f}(\widebar{\B{x}}^{(\B{n})},\B{e})}{\partial\B{x}^{(\B{n})}} :=
  \begin{bmatrix}
    \frac{\partial f_1}{\partial x_1} & \frac{\partial f_1}{\partial x_2'} \\
    \frac{\partial f_2}{\partial x_1} & \frac{\partial f_2}{\partial x_2'}
  \end{bmatrix}
  =
  \begin{bmatrix}
    0 & 1 \\
    0 & 0
  \end{bmatrix}
$$
is singular everywhere. In order to solve the \SDE s we can differentiate the second equation with respect to time to get
$$
  X_1 = X_2' = E' \,. 
$$
This \CDM\ only has solutions if the derivative $E'$ exists. If it exists, then the solutions are given by $X_1= E'$ and $X_2=E$. Thus, the solutions satisfy not only the obvious constraint $X_2=E$, but also need to satisfy the ``hidden'' constraint $X_1=E'$.
That a solution of the model depends on a derivative of the exogenous variable $E$ cannot happen in a system of RDEs. These constraints imply that every consistent full initial condition $(t_0,\widebar{\B{X}}^{(\B{m})}_{[0]})$ of $\C{R}$ needs to satisfy $X_{[0],1}=E_{t_0}'$ and $X_{[0],2}=E_{t_0}$ almost surely. 
\end{example}

After performing a stochastic perfect intervention $\intervene(I,\B{K}_I)$ on an \CDM\ $\C{R}$, all consistent full initial conditions $(t_0,\widebar{\B{X}}_{[0]}^{(\B{m})})$ must satisfy $\widebar{\B{X}}_{[0],I}^{(\B{m}_I)} = \widebar{\B{K}}_I^{(\B{m}_I)}$ almost surely. For example, the consistent initial conditions $(t_0,(\widebar{X}^{(1)}_{[0],0},\dots,\widebar{X}^{(1)}_{[0],d}))$ for the \CDM\ $\C{R}$ in Example~\ref{ex:HarmonicOscillatorIntervened} need to satisfy $\widebar{X}^{(1)}_{[0],0}=(0,0)$ and $\widebar{X}^{(1)}_{[0],d}=(L,0)$ after the perfect intervention $\intervene(\{1,d\},(0,L))$ on the model. 

In summary, Examples~\ref{ex:HigherOrderRandomInitialConditions}, \ref{ex:InconsistentInitialConditions} and \ref{ex:HiddenConstraint} show that the initial (random) value problems associated to \SDE s of an \CDM\ behave differently compared to those of RDEs, as
not every initial condition is consistent, and the solutions may involve (higher-order) derivatives of the exogenous process $\B{E}$.
\subsection{Existence and uniqueness of the solutions}
\label{sec:ExistenceAndUniquenessOfSolutions}


For RDEs, there exist sufficient conditions for the existence and uniqueness of the solutions with an initial condition, which are similar to the existence and uniqueness theorems for initial value problems for ODEs~\citep{Bun72,KP92,CL55}. 
No similar theorem is known in such generality for \SDE s, although there are some weaker results of this type for differential-algebraic equations~\citep{AP98}. In this subsection, we provide sufficient conditions for the existence and uniqueness of solutions with a specified initial condition, both locally (considering only a subset of the stochastic processes) and globally.

We start with an assumption on the form of the \SDE s for a subset of endogenous processses $\C{O}\subseteq\C{I}$. This assumption entails that for some subset $I \subseteq \C{O}$, the \SDE s corresponding to $I$ can be written as an RDE, while the remaining \SDE s for the complement $\C{O}\setminus I$ can be solved uniquely for their corresponding endogenous processes in terms of the other processes appearing in these \SDE s.
Additionally, smoothness conditions are imposed on exogenous processes and on dynamical causal mechanisms to ensure the required smoothness of the solution.\footnote{The required smoothness of the solutions implies that we need to make assumptions about the smoothness of the exogenous
processes and the dynamical causal mechanisms in the model. The assumption we made here is still rather crude in the sense that it
suffices, but it is not at all necessary; if desired, one can arrive at weaker conditions by carefully tracing through the
graph how the required smoothness of the solution can be guaranteed by demanding certain smoothness of each exogenous process
and each dynamical causal mechanism individually.}

\begin{assumption}{1}{(I\subseteq\C{O})}
\label{ass:ExplicitlySolvable}
  For the \CDM\ $\C{R}$ and subsets $I\subseteq\C{O}\subseteq\C{I}$, writing $J:=\C{O}\setminus I$ and 
  $P := \pa_{\col(\C{G}^a(\C{R}))}(\C{O})\setminus\C{O}$ 
  with $\col(\C{G}^a(\C{R}))$ the ``collapsed'' graph,\footnote{We will abuse notation by using the notation 
  $\col(\C{G}^a(\C{R}))$ for the graph that is isomorphic to the ``collapsed'' mixed graph of $\C{G}^a(\C{R})$ where 
  the nodes are labeled by $\C{I}\cup\C{J}$ instead of $\{\widebar{i}^{(n_i)}: i\in\C{I}\}\cup\{\{j\} : j\in\C{J} \}$.} 
  the following both hold:
  \begin{enumerate}
    \item the order tuple $\B{n}_I\geq 1$;
    \item there exist continuous functions $\B{g}_I:\BC{X}_I^{\B{n}_I}\times\BC{X}_J\times\BC{X}_P^{\B{n}_P+1}\times\BC{E}_P\to\BC{X}_I$ and $\B{g}_J:\BC{X}_I^{\B{n}_I}\times\BC{X}_P^{\B{n}_P+1}\times\BC{E}_P\to\BC{X}_J$ such that\footnote{For a subset $P\subseteq\C{I}\cup\C{J}$, we use the convention that we write $\widebar{\B{X}}_P^{(\B{n}_P)}$ and $\B{E}_P$ instead of $\widebar{\B{X}}_{P\cap\C{I}}^{(\B{n}_{P\cap\C{I}})}$ and $\B{E}_{P\cap\C{J}}$ respectively, and adopt a similar notation for variables and their spaces.} for all $\B{e}\in\BC{E}$ and for all $\widebar{\B{x}}^{(\B{n})}\in\BC{X}^{\B{n}+1}$
$$
  \B{x}_I^{(\B{n}_I)} = \B{g}_I(\widebar{\B{x}}_I^{(\B{n}_I-1)},\B{x}_J, \widebar{\B{x}}_P^{(\B{n}_P)},\B{e}_P) \quad\iff\quad
\B{x}_I = \B{f}_I(\widebar{\B{x}}^{(\B{n})},\B{e}) 
$$
and
$$
  \B{x}_J = \B{g}_J(\widebar{\B{x}}_I^{(\B{n}_I-1)},\widebar{\B{x}}_P^{(\B{n}_P)},\B{e}_P) \quad\iff\quad
  \B{x}_J = \B{f}_J(\widebar{\B{x}}^{(\B{n})},\B{e}) \,.
$$
  \end{enumerate}
\end{assumption}
In particular, under Assumption~\asref{ass:ExplicitlySolvable}{(\C{I}\subseteq\C{I})} the \SDEfull s of $\C{R}$ are equivalent to an RDE. For an \CDM\ that satisfies Assumption~\asref{ass:ExplicitlySolvable}{(I\subseteq\C{I})} with $I$ a strict subset of $\C{I}$, we can eliminate the processes $\B{X}_{\C{I}\setminus I}$ by substitution, giving an RDE for the endogenous processes $I$ of the form
\begin{equation}
\label{eq:InducedRDE}
  \B{X}_I^{(\B{n}_I)} = \B{g}_I(\widebar{\B{X}}_I^{(\B{n}_I-1)},\B{g}_{\C{I}\setminus I}(\widebar{\B{X}}^{(\B{n}_I-1)}_I,\B{E}),\B{E}) \,. 
\end{equation}
Every solution of the original \CDM\ satisfies this RDE, and every solution of this RDE induces a solution of the \CDM, if it is sufficiently smooth. 
For $\C{O} \subsetneq \C{I}$ we can think of Assumption~\asref{ass:ExplicitlySolvable}{(I \subseteq\C{O})} as applying this assumption to the subsystem with endogenous processes $\C{O}$, treating the remaining endogenous processes in $\C{I} \setminus \C{O}$ as external inputs of the subsystem. This will turn out to be useful in Section~\ref{sec:MarkovPropertyCDM} for proving a Markov property.

\begin{example}
\label{ex:SupplyDemandModelRDE}
Consider the price, supply and demand model of Example~\ref{ex:SupplyDemandModel}. This model satisfies Assumption~\asref{ass:ExplicitlySolvable}{(I\subseteq\C{O})} for $I=\{P\}$ and $\C{O}=\{S,P,D\}$. Substituting the zeroth-order \SDEfull s into the first-order equation of $X_P$ yields the RDE
\begin{equation}
\label{eq:SupplyDemandModel}
X_P'= \lambda (\beta_D - \beta_S)X_P + \lambda (E_D - E_S) \,. 
\end{equation}
If instead we take $\C{O}=\{S,P\}$, then this yields the RDE
$$
  X_P'= \lambda (X_D - \beta_S X_P - E_S) \,,
$$
where now $X_D$ is treated as an external input of the subsystem $\C{O}$.
\end{example}

We formalize the notions of the existence and uniqueness of solutions of a subsystem of the \CDM\ as follows.
\begin{definition}[Unique solvability of an initial value problem]
\label{def:UniqueSolvabilityIVP}
Let $\C{R} = \langle \C{I}, \C{J}, \BC{X}, \BC{E}, \B{n}, \B{f}, \B{E} \rangle$ be an \CDM,
  $I \subseteq\C{O}\subseteq\C{I}$ be subsets such that $\B{n}_I\geq 1$, $J:=\C{O}\setminus I$ and $P := \pa_{\col(\C{G}^a(\C{R}))}(\C{O})\setminus\C{O}$.
We call the \emph{initial value problem} $\langle \C{R}, I, \C{O} \rangle$ \emph{(uniquely) solvable} if
for any partial initial condition $(t_0, \widebar{\B{X}}_{I,[0]}^{(\B{n}_I-1)})$  and any $C^{\B{n}_P}$-stochastic process $\B{X}_P$, there exists an (a.s.\ unique) $C^{\B{n}_{\C{O}}}$-stochastic process $\B{X}_{\C{O}}$ that is a solution of the \SDEfull s\footnote{These equations are equivalent to the \SDEfull s (see Definition~\ref{def:FunctionalAndIntegratedParents}).}
$$
  \B{X}_{\C{O}} = \B{f}_{\C{O}}(\widebar{\B{X}}_{\C{O}}^{(\B{n}_{\C{O}})},\widebar{\B{X}}_{P}^{(\B{n}_{P})}, \B{E}_P) \,,
$$
with partial initial condition $(t_0, \widebar{\B{X}}_{I,[0]}^{(\B{n}_I-1)})$.
\end{definition}
As a special case, we obtain the notion of unique solvability for SCMs \citep{BPSM19}, where initial values play no role.
\begin{definition}[Unique solvability of SCMs]
Let $\C{M}:=\langle \C{I}, \C{J}, \BC{X}, \BC{E}, \B{f}, \B{E} \rangle$ be an SCM,
  $\C{O}\subseteq\C{I}$ a subset and $P := \pa_{\col(\C{G}^a(\C{R}))}(\C{O})\setminus\C{O}$.
We say that $\C{M}$ is uniquely solvable w.r.t.\ $\C{O}$ if for any
value $\B{x}_P \in \BC{X}_P$ and any value $\B{e}_P \in \BC{E}_P$, there exists an a.s.\ unique solution $\B{x}_{\C{O}} \in \BC{X}_{\C{O}}$ of the structural equations
$$
  \B{x}_{\C{O}} = \B{f}_{\C{O}}(\B{x}_{\C{O}}, \B{x}_{P}, \B{e}_P) \,.
$$
\end{definition}
Note that this corresponds to unique solvability of the initial value problem $\langle \C{M}, \emptyset, \C{O} \rangle$.
Since \CDM s that satisfy Assumption~\asref{ass:ExplicitlySolvable}{(I\subseteq \C{O})} have the property that they determine an RDE on the subset $I$, we can apply the existence and uniqueness results of RDEs on this subsystem, which leads to the following result. 
\begin{lemma}
\label{lem:ExistenceAndUniquenessSolutions}
  Let $\C{R}$ be an \CDM\ that satisfies Assumption~\asref{ass:ExplicitlySolvable}{(I\subseteq\C{O})} for subsets $I\subseteq\C{O}\subseteq\C{I}$. Let $J:=\C{O}\setminus I$ and $P := \pa_{\col(\C{G}^a(\C{R}))}(\C{O})\setminus\C{O}$. If the following three conditions hold:
  \begin{enumerate}
  \item the exogenous process $\B{E}_P$ is continuous;
    \item the composition of $\B{g}_I$ with $\B{g}_J$ is uniformly Lipschitz in its $I$-input, that is, there exists a constant\footnote{This result can be weakened slightly by making $\kappa$ dependent on $t\in T$, $\omega\in\Omega$ and the parent processes $\widebar{\B{X}}_P^{(\B{n}_P-1)}$ and $\B{E}_P$ (see also Theorem~1.2 in \citet{Bun72} or Theorem~3.2 in \citet{NR13}).} $\kappa > 0$ such that for all $\widebar{\B{x}}^{(\B{n}_I-1)}_I,\widebar{\B{y}}^{(\B{n}_I-1)}_I\in\BC{X}^{\B{n}_I}_I$, for all $\widebar{\B{x}}_{P}^{(\B{n}_P)} \in \BC{X}_P^{\B{n}_P+1}$ and for all $\B{e}_P \in \BC{E}_P$ the condition
\begin{gather*}
 \big\|
  g_I\big(\widebar{\B{x}}^{(\B{n}_I-1)}_I,\B{g}_{J}(\widebar{\B{x}}^{(\B{n}_I-1)}_I,\widebar{\B{x}}_{P}^{(\B{n}_P)},\B{e}_P),\widebar{\B{x}}_{P}^{(\B{n}_P)},\B{e}_P\big)
  - g_I\big(\widebar{\B{y}}^{(\B{n}_I-1)}_I,\B{g}_{J}(\widebar{\B{y}}^{(\B{n}_I-1)}_I,\widebar{\B{x}}_{P}^{(\B{n}_P)},\B{e}_P),\widebar{\B{x}}_{P}^{(\B{n}_P)},\B{e}_P\big) \big\| \\ 
  \leq \kappa \| x_I^{(0)} - y_I^{(0)} \|
\end{gather*}
      is satisfied, where $\|\cdot\|$ denotes the Euclidean norm on $\BC{X}_I$;
  \item for each $j \in J$, either $n_j = 0$, or $g_j$ only depends on $\B{e}_P$ (that is,
$g_j(\widebar{\B{x}}_I^{(\B{n}_I-1)},\widebar{\B{x}}_P^{(\B{n}_P)},\B{e}_P) = \tilde{g}_j(\B{e}_P)$ for 
$\tilde{g}_j:\BC{E}_P\to\BC{X}_j$) and $g_j(\B{E}_P)$ is a $C^{n_j}$-stochastic process;
  \end{enumerate}
  then $\langle \C{R}, I, \C{O} \rangle$ is uniquely solvable. 
\end{lemma}
This lemma guarantees the existence and uniqueness of solutions for a large class of (subsystems of) \CDM s. 
Indeed, it states that for any partial initial condition $(t_0,\widebar{\B{X}}^{(\B{n}_I-1)}_{I,[0]})$ and any $C^{\B{n}_P}$-stochastic process $\B{X}_P$ there exists an a.s.\ unique solution $\B{X}_{\C{O}}$ of the \SDEfull s
$$
  \B{X}_{\C{O}} = \B{f}_{\C{O}}(\widebar{\B{X}}_{\C{O}}^{(\B{n}_{\C{O}})}, \widebar{\B{X}}_P^{(\B{n}_P)}, \B{E}_P) \,,
$$
with initial condition
$$
  \big(\widebar{\B{X}}^{(\B{n}_I-1)}_I(t_0),\B{X}_J(t_0)\big) = \Big(\widebar{\B{X}}_{I,[0]}^{(\B{n}_I-1)}, \B{g}_J\big(\widebar{\B{X}}^{(\B{n}_I-1)}_{I,[0]}, \widebar{\B{X}}^{(\B{n}_P)}_P(t_0),\B{E}_P(t_0)\big) \Big)
$$
at $t_0$. In particular, this provides a sufficient condition for an initial condition to be consistent (see Definition~\ref{def:ConsistentInitialCondition}). 

In general, Assumption~\asref{ass:ExplicitlySolvable}{(I\subseteq\C{O})} for an \CDM\ is not preserved under a stochastic perfect intervention.
Consider for example the \CDM\ $\tilde{\C{R}}$ in Example~\ref{ex:TwoHarmonicOscillatorAmbiguousCausalInterpretation} which satisfies Assumption~\asref{ass:ExplicitlySolvable}{(\C{I}\subseteq\C{I})}. Performing the intervention $\intervene(1,0)$ on this model yields a model that does not satisfy Assumption~\asref{ass:ExplicitlySolvable}{(I\subseteq\C{I})} for any $I\subseteq\C{I}$. Under the following stronger assumption the \CDM\ will satisfy Assumption~\asref{ass:ExplicitlySolvable}{(I\subseteq\C{I})} for some $I\subseteq\C{I}$ after every stochastic perfect intervention.
\begin{assumption}{2}{(I\subseteq\C{O})}
  \label{ass:StructExplicitlySolvable}
  For the \CDM\ $\C{R}$ and subsets $I\subseteq\C{O}\subseteq\C{I}$, writing $J:=\C{O}\setminus I$ and 
  $P := \pa_{\col(\C{G}^a(\C{R}))}(\C{O})\setminus\C{O}$,
  the following all hold:
  \begin{enumerate}
    \item the order tuple $\B{n}_I\geq 1$; 
    \item there exist continuous functions $g_i:\C{X}_i^{n_i}\times\BC{X}_{\C{O}\setminus i}\times\BC{X}_P^{\B{n}_P+1}\times\BC{E}_P\to\C{X}_i$ for all $i\in I$ and $g_j:\BC{X}_I\times\BC{X}_P^{\B{n}_P+1}\times\BC{E}_P\to\C{X}_j$ for all $j\in J$ such that for all $i \in I$, all $j \in J$, all $\B{e}\in\BC{E}$ and all $\widebar{\B{x}}^{(\B{n})}\in\BC{X}^{\B{n}+1}$,
$$
  x_i^{(n_i)} = g_i(\widebar{x}_i^{(n_i-1)},\B{x}_{\C{O}\setminus i},\widebar{\B{x}}_P^{(\B{n}_P)},\B{e}_P)
\quad\iff\quad 
x_i = f_i(\widebar{\B{x}}^{(\B{n})},\B{e}) 
$$
and 
$$
x_j = g_j(\B{x}_I,\widebar{\B{x}}_P^{(\B{n}_P)},\B{e}_P)
\quad\iff\quad 
x_j = f_j(\widebar{\B{x}}^{(\B{n})},\B{e}) \,.
$$
  \end{enumerate}
\end{assumption}
In particular, Assumption~\asref{ass:StructExplicitlySolvable}{(I\subseteq\C{O})} implies Assumption~\asref{ass:ExplicitlySolvable}{(I\subseteq\C{O})}.

\begin{proposition}[Assumption~\asref{ass:StructExplicitlySolvable}{(I\subseteq\C{O})} under stochastic perfect intervention]
\label{prop:StructuralExplicitSolvabilityIntervention}
  Let $\C{R}$ be an \CDM\ that satisfies Assumption~\asref{ass:StructExplicitlySolvable}{(I\subseteq\C{O})} for subsets $I \subseteq \C{O}\subseteq\C{I}$. Then, for a stochastic perfect intervention $\intervene(L,\B{K}_L)$ for $L\subseteq \C{O}$,
  the intervened \CDM\ $\C{R}_{\intervene(L,\B{K}_L)}$ satisfies Assumption~\asref{ass:StructExplicitlySolvable}{(I\setminus L\subseteq\C{O})}.
\end{proposition}
This proposition shows the usefulness of Assumption~\asref{ass:StructExplicitlySolvable}{(I\subseteq\C{O})}, in that it gives a guarantee that after any stochastic perfect intervention on a subset of $\C{O}$, Assumption~\asref{ass:ExplicitlySolvable}{(\tilde{I}\subseteq\C{O})} is satisfied for some $\tilde{I}\subseteq\C{O}$, and hence Lemma~\ref{lem:ExistenceAndUniquenessSolutions} can be applied.

\paragraph*{Linear \CDM s}

Observe that a linear \CDM\ that satisfies Assumption~\asref{ass:ExplicitlySolvable}{(I\subseteq\C{O})} is of the following form.
\begin{proposition}
\label{prop:LinearCDMExplicitlySolvable}
  Let $\C{R}$ be a linear \CDM, $I\subseteq\C{O}\subseteq\C{I}$ be subsets, and let $J:=\C{O}\setminus I$ and $P := \pa_{\col(\C{G}^a(\C{R}))}(\C{O})\setminus\C{O}$. Then $\C{R}$ satisfies Assumption~\asref{ass:ExplicitlySolvable}{(I\subseteq\C{O})} iff the dynamic causal mechanism $\B{f}_{\C{O}}$ of $\C{R}$ restricted to $\C{O}$ is of the form
$$
  \left\{\begin{aligned}
    \B{f}_I(\widebar{\B{x}}^{(\B{n})},\B{e}) &:= B_{I I^{(\B{n}_I)}} \B{x}^{(\B{n}_I)}_I + B_{I \widebar{I}^{(\B{n}_I-1)}}\widebar{\B{x}}^{(\B{n}_I-1)}_I + 
    B_{I J}\B{x}_J + B_{I \widebar{P}^{(\B{n}_P)}}\widebar{\B{x}}_P^{(\B{n}_P)} + \Gamma_{I P} \B{e}_P \\
    \B{f}_J(\widebar{\B{x}}^{(\B{n})},\B{e}) &:= B_{J \widebar{I}^{(\B{n}_I-1)}}\widebar{\B{x}}^{(\B{n}_I-1)}_I + B_{J J} \B{x}_J + \B{x}_J + B_{J \widebar{P}^{(\B{n}_P)}}\widebar{\B{x}}_P^{(\B{n}_P)} + \Gamma_{J P} \B{e}_P \,,
  \end{aligned}\right.
$$
where $B_{I I^{(\B{n}_I)}}$ and $B_{JJ}$ are invertible matrices.
\end{proposition}

In particular, for linear \CDM s, Lemma~\ref{lem:ExistenceAndUniquenessSolutions} gives the following useful corollary.
\begin{corollary}
\label{cor:ExistenceAndUniquenessLinearCase}
  Let $\C{R}$ be a linear \CDM, $I\subseteq\C{O}\subseteq\C{I}$ be subsets, and let $J:=\C{O}\setminus I$ and $P := \pa_{\col(\C{G}^a(\C{R}))}(\C{O})\setminus\C{O}$.
If
  \begin{enumerate}
    \item $\C{R}$ satisfies Assumption~\asref{ass:ExplicitlySolvable}{(I\subseteq\C{O})};
    \item $\B{E}_P$ is continuous; 
    \item for each $j \in J$, either $n_j = 0$, or 
$(B_{J J}^{-1})_{jJ} B_{J \widebar{I}^{(\B{n}_I-1)}} = \B{0}$, $(B_{J J}^{-1})_{jJ} B_{J \widebar{P}^{(\B{n}_P)}} = \B{0}$ and 
$(B_{J J}^{-1})_{jJ} \Gamma_{J P} \B{E}_P$ is a $C^{n_j}$-stochastic process;
  \end{enumerate}
then $\langle \C{R}, I, \C{O}\rangle$ is uniquely solvable.
\end{corollary}
Examples of linear \CDM s that satisfy Assumption~\asref{ass:ExplicitlySolvable}{(I\subseteq\C{I})} for some subset $I$ are the \CDM s $\C{R}$ of Example~\ref{ex:HarmonicOscillator} and $\C{R}_{\intervene(\{1,d\},(0,L))}$ of Example~\ref{ex:HarmonicOscillatorIntervened}, which satisfy Assumption~\asref{ass:ExplicitlySolvable}{(\C{I}\subseteq\C{I})} and \asref{ass:ExplicitlySolvable}{(\C{I}\setminus\{1,d\}\subseteq\C{I})}, respectively. As the other conditions in Corollary~\ref{cor:ExistenceAndUniquenessLinearCase} are fulfilled, they both have an a.s.\ unique solution for each respective partial initial condition.

In particular, for linear \CDM s that satisfy Assumption~\asref{ass:StructExplicitlySolvable}{(I\subseteq\C{O})} we have the following corollary.
\begin{corollary}
\label{cor:ExistenceAndUniquenessLinearCaseStructural}
  Let $\C{R}$ be a linear \CDM, $I\subseteq\C{O}\subseteq\C{I}$ be subsets, and let $J:=\C{O}\setminus I$ and $P := \pa_{\col(\C{G}^a(\C{R}))}(\C{O})\setminus\C{O}$.
  If 
  \begin{enumerate}
    \item $\C{R}$ satisfies Assumption~\asref{ass:StructExplicitlySolvable}{(I\subseteq\C{O})}, 
    \item $\B{E}_P$ is continuous; 
    \item for each $j \in J$, either $n_j = 0$, or 
$(B_{J J}^{-1})_{jJ} B_{J \widebar{I}^{(\B{n}_I-1)}} = \B{0}$, $(B_{J J}^{-1})_{jJ} B_{J \widebar{P}^{(\B{n}_P)}} = \B{0}$ and 
$(B_{J J}^{-1})_{jJ} \Gamma_{J P} \B{E}_P$ is a $C^{n_j}$-stochastic process;
  \end{enumerate}
  then $\langle \C{R}_{\intervene(L,\B{K}_L)}, I\setminus L, \C{O} \rangle$ is uniquely solvable for any
  stochastic perfect intervention $\intervene(L,\B{K}_L)$ with $L\subseteq\C{O}$ and $\B{K}_L$ a $C^{\B{n}_L}$-stochastic process.
\end{corollary}
Examples of linear \CDM s that satisfy Assumption~\asref{ass:StructExplicitlySolvable}{(I\subseteq\C{I})} for some subset $I$ are the damped coupled harmonic oscillator of Example~\ref{ex:HarmonicOscillator} and the price, supply and demand model of Example \ref{ex:SupplyDemandModel}. Hence, the existence of solutions is guaranteed for both models after any (sufficiently smooth) stochastic perfect intervention, and the solutions are a.s.\ uniquely determined by the respective partial initial conditions.

\paragraph*{Nonlinear \CDM s}

An example of an \CDM\ that is not linear but satisfies Assumption~\asref{ass:StructExplicitlySolvable}{(I\subseteq\C{O})} is the bathtub model discussed in \citep{IS94}. The existence and uniqueness conditions apply to this particular model.
\begin{example}[Bathtub model]
\label{ex:Bathtub}
Water enters a bathtub from the faucet at a certain rate $X_{Q_i}$ and exits
the bathtub via the drain at a rate $X_{Q_o}$. The drain has a diameter of $X_K$, the
depth of the water is $X_D$ and the pressure at the base of the drain is $X_P$.
\citet{IS94} propose to model this as a dynamical system with
(random) differential equations given by
\begin{equation}
  \label{eq:Bathtub}
  \left\{\begin{aligned}
    X_K &= k_0 \\
    X_{Q_i} &= q_0 \\
    X_P' &= \alpha_2(\alpha_4 X_D - X_P) \\
    X_{Q_o}' &= \alpha_3(\alpha_1 X_K X_P - X_{Q_o})  \\
    X_D' &= \alpha_0(X_{Q_i} - X_{Q_o}) \,,
  \end{aligned}\right.
\end{equation}
where $k_0,q_0\in\RN_{>0}$ and $\B{\alpha}=(\alpha_0,\alpha_1,\dots,\alpha_4)\in\RN_{>0}^5$ are some constants. We consider the dynamic causal mechanism
$$
  \left\{\begin{aligned}
    f_K(\widebar{\B{x}}^{(\B{n})},\B{e}) &:= e_K \\
    f_{Q_i}(\widebar{\B{x}}^{(\B{n})},\B{e}) &:= e_{Q_i} \\
    f_P(\widebar{\B{x}}^{(\B{n})},\B{e}) &:= \alpha_4 x_D - \alpha_2^{-1}x_{P'}\\
    f_{Q_o}(\widebar{\B{x}}^{(\B{n})},\B{e}) &:=  \alpha_1 x_K x_P - \alpha_3^{-1}x_{Q_o'} \\
    f_D(\widebar{\B{x}}^{(\B{n})},\B{e}) &:= x_D + \alpha_0(x_{Q_i} - x_{Q_o}) - x_{D'} \,.
  \end{aligned}\right.
$$
  with order tuple $\B{n} := (n_K,n_{Q_i},n_P,n_{Q_o},n_D) = (0,0,1,1,1)$ and the exogenous processes are given by $E_K(t,\omega) := k_0$, $E_{Q_i}(t,\omega) := q_0$. The \SDEfull s of this \CDM, denoted by $\C{R}$, read
$$
  \left\{\begin{aligned}
    X_K     &= E_K \\
    X_{Q_i} &= E_{Q_i} \\
    X_P     &= \alpha_4 X_D - \alpha_2^{-1}X_P' \\
    X_{Q_o} &= \alpha_1 X_K X_P - \alpha_3^{-1}X_{Q_o}' \\
    X_D     &= X_D + \alpha_0(X_{Q_i} - X_{Q_o}) - X_D' \,,
  \end{aligned}\right. 
$$
  and have the same solutions as the system of equations~\eqref{eq:Bathtub} (see also Footnote~\ref{fn:RDE2CDM}). The corresponding \CDM\ graph is depicted in Figure~\ref{fig:Bathtub} (top left). This \CDM\ of the bathtub model satisfies Assumption~\asref{ass:StructExplicitlySolvable}{(\{P,Q_o,D\}\subseteq\C{I})} with $\C{I}=\{K, Q_i, P, Q_o, D\}$, and hence, after any sufficiently smooth stochastic perfect intervention $\intervene(L,\B{K}_L)$ with $L \subseteq \C{I}$, the intervened bathtub model $\C{R}_{\intervene(L,\B{K}_L)}$ satisfies Assumption~\asref{ass:StructExplicitlySolvable}{(\{P,Q_o,D\}\setminus L\subseteq\C{I})}. Since the induced RDE of the intervened model $\C{R}_{\intervene(L,\B{K}_L)}$ on the endogenous processes $\{P,Q_o,D\}\setminus L$ is linear in these endogenous processes, it follows from Lemma~\ref{lem:ExistenceAndUniquenessSolutions} that (for sufficiently smooth exogenous process $\B{K}_L$) $\C{R}_{\intervene(L,\B{K}_L)}$ has an a.s.\ unique solution for any partial initial condition $(t_0,\widebar{\B{X}}^{(\B{n}_{\{P,Q_o,D\}\setminus L})}_{\{P,Q_o,D\}\setminus L,[0]})$.
\end{example}
\begin{figure}
  \centering
  \begin{tikzpicture}[auto]
    \draw[|->,shorten >=0.1cm,shorten <=0.1cm,line width=0.2mm] (3.6,-0.7) to node {\small $t \rightarrow \infty$} (5.0,-0.7);
    \draw[|->,shorten >=0.1cm,shorten <=0.1cm,line width=0.2mm] (3.6,-6.5) to node {\small $t \rightarrow \infty$} (5.0,-6.5);
    \draw[|->,shorten >=0.1cm,shorten <=0.1cm,line width=0.2mm] (0,-3.4) to node {\small $\intervene(D, K_D)$} (0,-4.4);
    \draw[|->,shorten >=0.1cm,shorten <=0.1cm,line width=0.2mm] (8.0,-3.4) to node {\small $\intervene(D, K_D^*)$} (8.0,-4.4);
    \begin{scope}[scale=0.85]
      \node[var] (Qi) at (-3.0,-1.4) {$X_{Q_i}$};
      \node[var] (D) at (-1.5,0) {$X_D$};
      \node[var] (dD) at (-1.5,-1.4) {$X_D'$};
      \node[var] (P) at (0.0,-1.4) {$X_P$};
      \node[var] (dP) at (0.0,-2.8) {$X_P'$};
      \node[var] (Qo) at (1.5,0) {$X_{Q_o}$};
      \node[var] (dQo) at (1.5,-1.4) {$X_{Q_o}'$};
      \node[var] (K) at (3.0,-1.4) {$X_K$};
      \draw[draw=black!60] (-3.5,-0.9) rectangle (-2.5,-1.9);
      \draw[draw=black!60] (-2.0,0.5) rectangle (-1.0,-1.9);
      \draw[draw=black!60] (-0.5,-0.9) rectangle (0.5,-3.3);
      \draw[draw=black!60] (1.0,0.5) rectangle (2.0,-1.9);
      \draw[draw=black!60] (2.5,-0.9) rectangle (3.5,-1.9);
      \node[text=black!60] at (-3,-2.3) {$\widebar{X}^{(n_{Q_i})}_{Q_i}$};
      \node[text=black!60] at (-1.5,-2.3) {$\widebar{X}^{(n_D)}_D$};
      \node[text=black!60] at (0,-0.5) {$\widebar{X}^{(n_P)}_P$};
      \node[text=black!60] at (1.5,-2.3) {$\widebar{X}^{(n_{Q_o})}_{Q_o}$};
      \node[text=black!60] at (3.0,-2.3) {$\widebar{X}^{(n_K)}_K$};
      \draw[arr,dashed, bend left=15] (P) to (dP);
      \draw[arr,dashed, bend left=15] (Qo) to (dQo);
      \draw[arr,dashed, bend left=15] (D) to (dD);
      \draw[arr, dashed, bend left=15] (dP) to (P);
      \draw[arr] (D) to (P);
      \draw[arr, dashed, bend left=15] (dQo) to (Qo);
      \draw[arr] (K) to (Qo);
      \draw[arr] (P) to (Qo);
      \draw[arr, dashed, bend left=15] (dD) to (D);
      \draw[arr, dashed] (D) to [out=60,in=120,looseness=5] (D);
      \draw[arr] (Qi) to (D);
      \draw[arr] (Qo) to (D);
      \node at (0,0.8) {\scriptsize $\C{G}(\C{R})$:};
    \end{scope}
    \begin{scope}[xshift=8.0cm,yshift=-0.5cm]
      \node[var] (Qin) at (-2,-1) {$X_{Q_i}$};
      \node[var] (D) at (-1,0) {$X_D$};
      \node[var] (P) at (0,-1) {$X_P$};
      \node[var] (Qout) at (1,0) {$X_{Q_o}$};
      \node[var] (K) at (2,-1) {$X_K$};
      \draw[arr] (Qin) edge (D);
      \draw[arr] (Qout) edge (D);
      \draw[arr] (D) edge (P);
      \draw[arr] (P) edge (Qout);
      \draw[arr] (K) edge (Qout);
      \draw[arr, dashed] (D) to [out=60,in=120,looseness=5] (D);
      \node at (0,1.165) {\scriptsize $\C{G}(\C{M}_{\C{R}})$:};
    \end{scope}
    \begin{scope}[scale=0.85,yshift=-6.8cm]
      \node[var] (Qi) at (-3.0,-1.4) {$X_{Q_i}$};
      \node[var] (D) at (-1.5,0) {$X_D$};
      \node[var] (dD) at (-1.5,-1.4) {$X_D'$};
      \node[var] (P) at (0.0,-1.4) {$X_P$};
      \node[var] (dP) at (0.0,-2.8) {$X_P'$};
      \node[var] (Qo) at (1.5,0) {$X_{Q_o}$};
      \node[var] (dQo) at (1.5,-1.4) {$X_{Q_o}'$};
      \node[var] (K) at (3.0,-1.4) {$X_K$};
      \draw[draw=black!60] (-3.5,-0.9) rectangle (-2.5,-1.9);
      \draw[draw=black!60] (-2.0,0.5) rectangle (-1.0,-1.9);
      \draw[draw=black!60] (-0.5,-0.9) rectangle (0.5,-3.3);
      \draw[draw=black!60] (1.0,0.5) rectangle (2.0,-1.9);
      \draw[draw=black!60] (2.5,-0.9) rectangle (3.5,-1.9);
      \node[text=black!60] at (-3,-2.3) {$\widebar{X}^{(n_{Q_i})}_{Q_i}$};
      \node[text=black!60] at (-1.5,-2.3) {$\widebar{X}^{(n_D)}_D$};
      \node[text=black!60] at (0,-0.5) {$\widebar{X}^{(n_P)}_P$};
      \node[text=black!60] at (1.5,-2.3) {$\widebar{X}^{(n_{Q_o})}_{Q_o}$};
      \node[text=black!60] at (3.0,-2.3) {$\widebar{X}^{(n_K)}_K$};
      \draw[arr,dashed, bend left=15] (P) to (dP);
      \draw[arr,dashed, bend left=15] (Qo) to (dQo);
      \draw[arr,dashed] (D) to (dD);
      \draw[arr, dashed, bend left=15] (dP) to (P);
      \draw[arr] (D) to (P);
      \draw[arr, dashed, bend left=15] (dQo) to (Qo);
      \draw[arr] (K) to (Qo);
      \draw[arr] (P) to (Qo);
      \node at (0,0.9) {\scriptsize $\C{G}(\C{R}_{\intervene(D,\,K_D)})$:};
    \end{scope}
    \begin{scope}[xshift=8.0cm,yshift=-6.2cm]
      \node[var] (Qin) at (-2,-1) {$X_{Q_i}$};
      \node[var] (D) at (-1,0) {$X_D$};
      \node[var] (P) at (0,-1) {$X_P$};
      \node[var] (Qout) at (1,0) {$X_{Q_o}$};
      \node[var] (K) at (2,-1) {$X_K$};
      \draw[arr] (D) edge (P);
      \draw[arr] (P) edge (Qout);
      \draw[arr] (K) edge (Qout);
      \node at (0,1.045) {\scriptsize $\C{G}(\C{M}_{\C{R}_{\intervene(D,\,K_D)}}) = \C{G}((\C{M}_{\C{R}})_{\intervene(D,K_D^*)}):$};
    \end{scope}
\end{tikzpicture}
\caption{Graphs of the bathtub model: original model $\C{R}$ of Example~\ref{ex:Bathtub} (top left), the equilibrated model $\C{M}_{\C{R}}$ (top right), the intervened model $\C{R}_{\intervene(D,K_D)}$ (bottom left), and the intervened and equilibrated model $\C{M}_{\C{R}_{\intervene(D,K_D)}}$ (bottom right) of Example~\ref{ex:Bathtub_do_D}.}
\label{fig:Bathtub}
\end{figure}

\subsection{Markov property for \CDM s with initial conditions}
\label{sec:MarkovPropertyCDM}

Theoretical results of key importance concerning SCMs are their so-called Markov properties, which allow to read off conditional independencies in the solutions of an SCM from the graph of the SCM \citep{FM17,BPSM19}.
The two most well-known Markov properties for SCMs are the $d$-separation criterion (which applies to acyclic SCMs, amongst others), and the $\sigma$-separation criterion (which applies for example to the more general class of simple SCMs that can contain causal cycles).
Here we derive a Markov property for \CDM s with initial conditions that is analogous to the $\sigma$-separation criterion for SCMs.

Key to proving Markov properties is the existence and uniqueness of solutions for each subsystem consisting of one strongly connected component of the collapsed graph of the \CDM, augmented with initial conditions.
By reinterpreting continuous stochastic processes as random variables taking values in a space of continuous functions, we can make use of the existing $\sigma$-separation Markov property for SCMs to derive Markov properties for \CDM s.

To avoid complicating matters further with smoothness assumptions, we will assume that the order tuple is as small as possible.
\begin{definition}[Tight order tuple]
  Let $\C{R} = \langle \C{I}, \C{J}, \BC{X}, \BC{E}, \B{n}, \B{f}, \B{E} \rangle$ be an \CDM. Its order tuple $\B{n}$ is called \emph{tight} if for each $i \in \C{I}$, either $n_i = 0$, or $n_i > 0$ and the edge $i^{(n_i)} \gto i^{(0)}$ appears in $\C{G}^a(\C{R})$.
\end{definition}
Note that the order tuple is tight if and only if each cluster $\widebar{i}^{(n_i)}$ in the augmented graph $\C{G}^a(\C{R})$ forms a cycle in the cluster, that is, if there is a directed path in the cluster from each node in the cluster to any other node in the cluster.
\begin{definition}[Augmented collapsed graph for \CDM s]
Let $\C{R} = \langle \C{I}, \C{J}, \BC{X}, \BC{E}, \B{n}, \B{f}, \B{E} \rangle$ be an \CDM\ with tight order tuple.
  We define the \emph{augmented collapsed graph} $\C{G}_{[0]}^+(\C{R})$ of $\C{R}$ as the directed graph with nodes $\C{I} \cup \C{J} \cup \C{I}_{[0]}$, where $\C{I}_{[0]} := \{i_{[0]} : i \in \C{I} : n_i \ge 1\}$, directed edges $k \gto i$ (but dashed $i \gtod i$ if $k$ = $i$) if either $k^{(m_k)}\in\widebar{\C{I}}^{(\B{n})}$ is functional parent of $i \in \C{I}$ for some $m_k$ or $k\in\C{J}$ is functional parent of $i \in \C{I}$, and additional directed edges $i_{[0]} \gto i$ for those $i \in \C{I}$ with $i_{[0]} \in \C{I}_{[0]}$.
\end{definition}
The nodes $i_{[0]}$ represent partial initial conditions $(t_0,\widebar{X}_{[0],i}^{(n_i-1)})$, while the nodes in $\C{I} \cup \C{J}$ represent endogenous stochastic processes $X_i$ for $i \in \C{I}$, and exogenous stochastic processes $E_j$ for $j \in \C{J}$.
The augmented collapsed graph of an \CDM\ (with tight order tuple) is similar to its augmented graph, except that clusters are collapsed and nodes representing initial conditions have been added.
Figure~\ref{fig:Bathtub_collapsed} (top right) shows the augmented collapsed graph for the bathtub model of Example~\ref{ex:Bathtub}, and for comparison, the augmented graph is also shown (top left).

We can now prove that under conditions that guarantee the existence and uniqueness of a solution locally for each strongly
connected component of the augmented collapsed graph, there exists a global solution that is unique and satisfies the 
$\sigma$-separation criterion with respect to the augmented collapsed graph.

\begin{theorem}[Markov property for \CDM s with initial conditions]\label{thm:Markov_property_SDCM}
  Let $\C{R} = \langle \C{I}, \C{J}, \BC{X}, \BC{E}, \B{n}, \B{f}, \B{E} \rangle$ be an \CDM\ with tight order tuple.
  Suppose that for each strongly connected component $S\subseteq\C{I}$ of $\C{G}_{[0]}^+(\C{R})$,
  $\C{R}$ satisfies Assumption~\asref{ass:ExplicitlySolvable}{(I_S\subseteq S)} for some subset $I_S \subseteq S$ and $\langle \C{R}, I_S, S \rangle$ is uniquely solvable. Then for any partial initial condition $\big(t_0,(\widebar{X}_{[0],i}^{(n_i-1)})_{i\in\C{I}_{[0]}}\big)$, the \CDM\ $\C{R}$ has an a.s.\ unique solution with that partial initial condition.
  If $(\widebar{X}_{[0],i}^{(n_i-1)})_{i\in\C{I}_{[0]}}$ is independent, and independent of $\B{E}$, the solution $\B{X}$ satisfies the following Markov property:
$$A \SEP_{\C{G}_{[0]}^+(\C{R})}^\sigma B \given C \implies \B{Z}_A \CI \B{Z}_B \given \B{Z}_C$$
for all subsets of nodes $A,B,C$ of $\C{G}^+_{[0]}(\C{R})$, where
$\B{Z}_A := (\B{X}_{A \cap \C{I}},\widebar{\B{X}}_{[0],A \cap \C{I}_{[0]}}^{(\B{n}_{A \cap \C{I}_{[0]}})},\B{E}_{A \cap \C{J}})$
for $A \subseteq \C{I} \cup \C{I}_{[0]} \cup \C{J}$.
\end{theorem}
The conditional independence in this Markov property requires to interpret the endogenous process $\B{X}$ as a random element of $\C{C}^{\B{n}}(T,\BC{X})$ and the exogenous process $\B{E}$ as a random element of $\C{C}^{0}(T,\BC{E})$.
In other words, we may conclude the independence of entire processes (and initial conditions), conditional on entire processes (and initial conditions).

We can extend this result to obtain a Markov property for the solutions evaluated at times $t_0$ and $t_1$.
For this, we extend the graph with nodes that correspond to evaluating the endogenous processes at time $t_1$.
\begin{definition}[Evaluated augmented collapsed graph for \CDM s]
\label{def:EvaluatedAugmentedCollapsedGraph}
Let $\C{R} = \langle \C{I}, \C{J}, \BC{X}, \BC{E}, \B{n}, \B{f}, \B{E} \rangle$ be an \CDM\ with tight order tuple.
We define
the \emph{evaluated augmented collapsed graph} $\C{G}_{[0]\dots[1]}^+(\C{R})$ of $\C{R}$ as the augmented collapsed graph $\C{G}_{[0]}^+$, extended with additional nodes $\C{I}_{[1]} := \{ i_{[1]} : i \in \C{I} \}$ and directed edges $i \gto i_{[1]}$ for $i \in \C{I}$.
\end{definition}
The additional nodes in the evaluated augmented collapsed graph $\C{G}^+_{[0]\dots[1]}(\C{R})$ correspond with the evaluation of a process at time $t_1$, that is, $i_{[1]}$ corresponds with $\widebar{X}_i^{(n_i)}(t_1)$. 
Figure~\ref{fig:Bathtub_collapsed} (bottom left) shows the evaluated augmented collapsed graph for the bathtub model of Example~\ref{ex:Bathtub}.
We get the following corollary almost for free.
\begin{corollary}
\label{cor:evaluated_Markov_property_SDCM}
Under the assumptions of Theorem~\ref{thm:Markov_property_SDCM}, the following Markov property also holds:
$$
  A \SEP_{\C{G}^+_{[0]\dots[1]}(\C{R})}^\sigma B \given C \implies \B{Z}_A \CI \B{Z}_B \given \B{Z}_C
$$
  for any subsets of nodes $A,B,C$ of the evaluated augmented collapsed graph $\C{G}^+_{[0]\dots[1]}(\C{R})$, where
  for $A \subseteq \C{I} \cup \C{I}_{[0]} \cup \C{I}_{[1]} \cup \C{J}$ we write
  $\B{Z}_A := (\B{X}_{A \cap \C{I}},\widebar{\B{X}}_{[0],A \cap \C{I}_{[0]}}^{(\B{n}_{A \cap \C{I}_{[0]}})},\widebar{\B{X}}_{A \cap \C{I}_{[1]}}^{(\B{n}_{A\cap\C{I}_{[1]}})}(t_1),\B{E}_{A \cap \C{J}})$
  with $\B{X}$ being an a.s.\ unique solution of $\C{R}$ with initial condition 
$\big(t_0,(\widebar{X}_{[0],i}^{(n_i-1)})_{i\in\C{I}_{[0]}}\big)$.
\end{corollary}

We can also marginalize out the ``process nodes'' and retain only the ``random variable'' nodes, in effect only considering observations of the processes at times $t_0$ and $t_1$.
\begin{definition}[Transition graph for \CDM s]
\label{def:TransitionGraph}
Let $\C{R} = \langle \C{I}, \C{J}, \BC{X}, \BC{E}, \B{n}, \B{f}, \B{E} \rangle$ be an \CDM\ with tight order tuple.
We define
the \emph{transition graph} $\C{G}_{[0]\dots[1]}(\C{R})$ of $\C{R}$ as the directed graph with nodes $\C{I}_{[0]}\cup\C{I}_{[1]}\cup\C{J}$, where $\C{I}_{[0]} := \{i_{[0]} : i \in \C{I} : n_i \ge 1\}$ and $\C{I}_{[1]} := \{ i_{[1]} : i \in \C{I} \}$, and directed edges $i\gto j$ if there exists a directed path $i\gto \ldots \gto j$ in the evaluated augmented collapsed graph $\C{G}_{[0]\dots[1]}^+(\C{R})$.
\end{definition}
The transition graph $\C{G}_{[0]\dots[1]}(\C{R})$ is obtained from the evaluated augmented collapsed graph $\C{G}^+_{[0]\dots[1]}(\C{R})$ by graphically marginalizing\footnote{The result of a graphical marginalization is also known as the ``latent projection'', see for example \citep{BPSM19}.} out the nodes $\C{I}$ representing the full endogenous processes, and keeping only the nodes $\C{I}_{[0]} \cup \C{I}_{[1]}$ corresponding with the evaluations of the processes at time $t_0$ and time $t_1$, in addition to the nodes $\C{J}$ corresponding with the exogenous processes.
Figure~\ref{fig:Bathtub_collapsed} (bottom right) shows the transition graph for the bathtub model of Example~\ref{ex:Bathtub}.

\begin{corollary}
\label{cor:transition_Markov_property_SDCM}
Under the assumptions of Theorem~\ref{thm:Markov_property_SDCM}, the following Markov property also holds:
$$
  A \SEP_{\C{G}_{[0]\dots[1]}(\C{R})}^\sigma B \given C \implies \B{Z}_A \CI \B{Z}_B \given \B{Z}_C
$$
  for any subsets of nodes $A,B,C$ of the transition graph $\C{G}_{[0]\dots[1]}(\C{R})$, where
  for $A \subseteq \C{I}_{[0]} \cup \C{I}_{[1]} \cup\C{J}$ we write
      $\B{Z}_A := (\widebar{\B{X}}_{[0],A \cap \C{I}_{[0]}}^{(\B{n}_{A \cap \C{I}_{[0]}})},\widebar{\B{X}}_{A \cap \C{I}_{[1]}}^{(\B{n}_{A\cap\C{I}_{[1]}})}(t_1),\B{E}_{A\cap\C{J}})$
  with $\B{X}$ being an a.s.\ unique solution of $\C{R}$ with initial condition 
$\big(t_0,(\widebar{X}_{[0],i}^{(n_i-1)})_{i\in\C{I}_{[0]}}\big)$.
\end{corollary}

\begin{example}[Markov properties for the bathtub model]
\label{ex:MarkovPropertyBathtub}
  The bathtub model of Example~\ref{ex:Bathtub} satisfies the assumptions of Theorem~\ref{thm:Markov_property_SDCM} and its Corollaries~\ref{cor:evaluated_Markov_property_SDCM} and \ref{cor:transition_Markov_property_SDCM}.
  The corresponding graphs are illustrated in Figure~\ref{fig:Bathtub_collapsed}.
  We can, for example, read off from the augmented collapsed graph $\C{G}^+_{[0]}(\C{R})$ that $X_{Q_i} \CI X_K$.
  From the evaluated augmented collapsed graph $\C{G}^+_{[0]\dots[1]}(\C{R})$ and the transition graph $\C{G}_{[0]\dots[1]}(\C{R})$ we can read off that $X_{K}(t_1) \CI X_{Q_i}(t_1)$, that is,
  the inflow through the faucet is independent of the drain diameter at time $t_1$, provided they are at $t_0$.
  The latter is hardly surprising, but serves to illustrate how one can use the Markov properties to arrive at conditional independence statements about the solution without actually solving the \CDM, by carefully tracing the functional relations encoded in the \SDEfull s of the model.
\end{example}

\begin{figure}
  \centering
  \begin{tikzpicture}[auto]
    \begin{scope}
      \node at (0,2.7) {\scriptsize augmented graph $\C{G}^a(\C{R})$:};
      \node[var] (Qi) at (-2,-1) {$X_{Q_i}$};
      \node[exvar] (EQin) at (-2,1) {$E_{Q_i}$};
      \node[var] (D) at (-1,0) {$X_D$};
      \node[var] (dD) at (-1,-1.5) {$X_D'$};
      \node[var] (P) at (0,-1) {$X_P$};
      \node[var] (dP) at (0,-2.5) {$X_P'$};
      \node[var] (Qo) at (1,0) {$X_{Q_o}$};
      \node[var] (dQo) at (1,-1.5) {$X_{Q_o}'$};
      \node[var] (K) at (2,-1) {$X_K$};
      \node[exvar] (EK) at (2,1) {$E_K$};
      \draw[draw=black!60] (-2.4,-1.4) rectangle (-1.6,-0.6);
      \draw[draw=black!60] (-1.4,0.45) rectangle (-0.6,-1.9);
      \draw[draw=black!60] (-0.4,-0.6) rectangle (0.4,-2.9);
      \draw[draw=black!60] (0.55,0.45) rectangle (1.45,-1.95);
      \draw[draw=black!60] (1.6,-1.4) rectangle (2.4,-0.6);
      \draw[arr] (EQin) edge (Qi);
      \draw[arr] (EK) edge (K);
      \draw[arr,dashed, bend left=15] (P) to (dP);
      \draw[arr,dashed, bend left=15] (Qo) to (dQo);
      \draw[arr,dashed, bend left=15] (D) to (dD);
      \draw[arr, dashed, bend left=15] (dP) to (P);
      \draw[arr] (D) to (P);
      \draw[arr, dashed, bend left=15] (dQo) to (Qo);
      \draw[arr] (K) to (Qo);
      \draw[arr] (P) to (Qo);
      \draw[arr, dashed, bend left=15] (dD) to (D);
      \draw[arr, dashed] (D) to [out=60,in=120,looseness=5] (D);
      \draw[arr] (Qi) to (D);
      \draw[arr] (Qo) to (D);
    \end{scope}
    \begin{scope}[xshift=7.0cm]
      \node[var] (Qin) at (-2,-1) {$X_{Q_i}$};
      \node[exvar] (EQin) at (-2,1) {$E_{Q_i}$};
      \node[var] (D) at (-1,0) {$X_D$};
      \node[var] (P) at (0,-1) {$X_P$};
      \node[var] (Qout) at (1,0) {$X_{Q_o}$};
      \node (D0) at (-1,2) {$X_D(t_0)$};
      \node (P0) at (0,1) {$X_P(t_0)$};
      \node (Qout0) at (1,2) {$X_{Q_o}(t_0)$};
      \node[var] (K) at (2,-1) {$X_K$};
      \node[exvar] (EK) at (2,1) {$E_K$};
      \draw[arr] (EQin) edge (Qin);
      \draw[arr] (Qin) edge (D);
      \draw[arr] (Qout) edge (D);
      \draw[arr] (D) edge (P);
      \draw[arr] (P) edge (Qout);
      \draw[arr] (EK) edge (K);
      \draw[arr] (K) edge (Qout);
      \draw[arr] (D0) edge (D);
      \draw[arr] (P0) edge (P);
      \draw[arr] (Qout0) edge (Qout);
      \draw[arr, dashed] (D) to [out=-60,in=-120,looseness=5] (D);
      \node at (0,2.7) {\scriptsize augmented collapsed graph $\C{G}^+_{[0]}(\C{R})$:};
    \end{scope}
    \begin{scope}[yshift=-6.5cm]
      \node[var] (Qin) at (-2,-1) {$X_{Q_i}$};
      \node[exvar] (EQin) at (-2,1) {$E_{Q_i}$};
      \node[var] (D) at (-1,0) {$X_D$};
      \node[var] (P) at (0,-1) {$X_P$};
      \node[var] (Qout) at (1,0) {$X_{Q_o}$};
      \node[var] (K) at (2,-1) {$X_K$};
      \node (D0) at (-1,2) {$X_D(t_0)$};
      \node (P0) at (0,1) {$X_P(t_0)$};
      \node (Qout0) at (1,2) {$X_{Q_o}(t_0)$};
      \node (Qin1) at (-2,-3) {$X_{Q_i}(t_1)$};
      \node (D1) at (-1,-2) {$X_D(t_1)$};
      \node (P1) at (0,-3) {$X_P(t_1)$};
      \node (Qout1) at (1,-2) {$X_{Q_o}(t_1)$};
      \node (K1) at (2,-3) {$X_K(t_1)$};
      \node[exvar] (EK) at (2,1) {$E_K$};
      \draw[arr] (EQin) edge (Qin);
      \draw[arr] (Qin) edge (D);
      \draw[arr] (Qout) edge (D);
      \draw[arr] (D) edge (P);
      \draw[arr] (P) edge (Qout);
      \draw[arr] (EK) edge (K);
      \draw[arr] (K) edge (Qout);
      \draw[arr] (D0) edge (D);
      \draw[arr] (P0) edge (P);
      \draw[arr] (Qout0) edge (Qout);
      \draw[arr] (Qin) edge (Qin1);
      \draw[arr] (D) edge (D1);
      \draw[arr] (P) edge (P1);
      \draw[arr] (Qout) edge (Qout1);
      \draw[arr] (K) edge (K1);
      \draw[arr, dashed] (D) to [out=-60,in=-120,looseness=5] (D);
      \node at (0,2.7) {\scriptsize evaluated augmented collapsed graph $\C{G}^+_{[0]\dots[1]}(\C{R})$:};
    \end{scope}
    \begin{scope}[xshift=7.0cm,yshift=-6.5cm]
      \node[exvar] (EQin) at (-2,1) {$E_{Q_i}$};
      \node (D0) at (-1,2) {$X_D(t_0)$};
      \node (P0) at (0,1) {$X_P(t_0)$};
      \node (Qout0) at (1,2) {$X_{Q_o}(t_0)$};
      \node (Qin1) at (-2,-3) {$X_{Q_i}(t_1)$};
      \node (D1) at (-1,-2) {$X_D(t_1)$};
      \node (P1) at (0,-3) {$X_P(t_1)$};
      \node (Qout1) at (1,-2) {$X_{Q_o}(t_1)$};
      \node (K1) at (2,-3) {$X_K(t_1)$};
      \node[exvar] (EK) at (2,1) {$E_K$};
      \draw[arr] (EQin) edge (D1);
      \draw[arr] (EQin) edge[out=-55,in=110] (P1);
      \draw[arr] (EQin) edge (Qout1);
      \draw[arr] (Qout0) edge[out=-105,in=60] (D1);
      \draw[arr] (Qout0) edge (P1);
      \draw[arr] (Qout0) edge (Qout1);
      \draw[arr] (EK) edge (D1);
      \draw[arr] (EK) edge[out=-125,in=70] (P1);
      \draw[arr] (EK) edge (Qout1);
      \draw[arr] (D0) edge (D1);
      \draw[arr] (D0) edge (P1);
      \draw[arr] (D0) edge[out=-75,in=120] (Qout1);
      \draw[arr] (P0) edge (D1);
      \draw[arr] (P0) edge (P1);
      \draw[arr] (P0) edge (Qout1);
      \draw[arr] (EQin) edge (Qin1);
      \draw[arr] (EK) edge (K1);
      \node at (0,2.7) {\scriptsize transition graph $\C{G}_{[0]\dots[1]}(\C{R})$:};
    \end{scope}
  \end{tikzpicture}
\caption{Different graphs of the bathtub model of Examples~\ref{ex:Bathtub} and \ref{ex:MarkovPropertyBathtub}.}
\label{fig:Bathtub_collapsed}
\end{figure}

\section{Equilibration of \CDM s}
\label{sec:EquilibrationOfCDMs}

In this section, we will take $T = [t_0,\infty)$ and study the equilibrium states of \CDM s and, in particular, of steady \CDM s, which are \CDM s for which the \SDEfull s and exogenous processes become explicitly time-independent asymptotically as $t \to \infty$. We introduce an equilibration operation on a steady \CDM, which equilibrates the model to an SCM such that all the equilibrium states of the \CDM\ are described by the solutions of the SCM. Intuitively, this equilibration operation separately equilibrates each dynamic causal mechanism, which corresponds mathematically to transforming each \SDEfull\ into a structural equation of the SCM. We show that this equilibration operation commutes with perfect stochastic interventions, without requiring the strong global stability assumption of \citet{MJS13}, which assumes that all the solutions equilibrate to the same static equilibrium state. This allows to study the causal semantics of the equilibrium states of steady \CDM s within the framework of SCMs. 

We start in Section~\ref{sec:EquilibratingSolutions} with the definition of equilibrating solutions and their corresponding equilibrium states. In Section~\ref{sec:SteadyCDM}, we define the class of steady \CDM s which have several convenient convergence properties. In Section~\ref{sec:Equilibration}, we show how one can equilibrate a steady \CDM\ to an SCM. In Section~\ref{sec:GraphOfTheEquilibration}, we show how the equilibration acts on the graph of an \CDM. In Section~\ref{sec:CommutationEquilibrationIntervention}, we show that the equilibration operation commutes with intervention. 
We discuss in Section~\ref{sec:NonRecoverabilityCDMfromSCM} the inverse problem of finding steady \CDM s for which all the solutions equilibrate to solutions of the SCM independently of the initial condition. We provide sufficient conditions under which one can construct a first-order steady \CDM\ such that its equilibration coincides with a given linear SCM. This establishes a class of linear SCMs that model the causal equilibrium semantics of certain linear dynamical systems.
In Section~\ref{sec:CausalInterpretationGraphEquilibratedCDM}, we discuss some subtleties in the causal interpretation of the graph of the equilibrated \CDM. 

\subsection{Equilibrating solutions and equilibrium states}
\label{sec:EquilibratingSolutions}

In this subsection, we define the equilibrating solutions of an \CDM\ as those solutions for which all the higher-order derivatives that are considered in the model converge to zero a.s.. For a stochastic process $\B{X}$ we say that it \emph{converges almost surely} to a random variable $\B{X}^*$, if the limit $\lim_{t\rightarrow \infty}\B{X}_t$ exists almost surely\footnote{In that case, it defines a random variable, because $\lim_{t\rightarrow\infty}\B{X}_t = \lim_{\substack{t\rightarrow\infty \\ t\in\NN}}\B{X}_t$ a.s., and the latter is a random variable.} and is a.s.\ equal to $\B{X}^*$. In this case, we call $\B{X}$ \emph{almost surely convergent}. 

\begin{definition}[Equilibrating solution, equilibrium state]
\label{def:EquilibratingSolution}
Let $\B{X}$ be a solution of an \CDM\ $\C{R}$. We call $\B{X}$ an \emph{equilibrating solution}, if $\widebar{\B{X}}^{(n)}$ is a.s.\ convergent. In particular, an equilibrating solution $\B{X}$ converges almost surely to a random variable $\B{X}^*$, and we say that $\B{X}$ \emph{equilibrates} to $\B{X}^*$ and call $\B{X}^*$ an \emph{equilibrium state} of $\C{R}$. 
\end{definition}
An example of an \CDM\ with equilibrium states is the price, supply and demand model of Example~\ref{ex:SupplyDemandModel}, where the equilibrium states correspond to ``market equilibrium'', as illustrated in the following example.
\begin{example}[Market equilibrium]
\label{ex:MarketEquilibrium}
Consider the price, supply and demand model of Example~\ref{ex:SupplyDemandModel} with $E_S$ and $E_D$ constant exogenous processes. Market equilibrium for this model is reached if 
$$
X_D^* - X_S^* = 0 \,,
$$
that is, if the demanded and supplied quantities become equal asymptotically.
The solutions that satisfy this condition are equilibrating solutions for which 
$$
\begin{aligned}
  &X_P'^* = 0\,,\quad & X_P^* = \frac{E_D-E_S}{\beta_S - \beta_D}\,,\quad & X_S^* = X_D^* = \frac{\beta_S E_D - \beta_D E_S}{\beta_S - \beta_D} \,.
\end{aligned}
$$
\end{example}
In fact, for every solution $\B{X}$ that equilibrates, the higher-order derivatives of $\B{X}$ must converge to zero almost surely.
\begin{proposition}
\label{prop:EquilibriumSolution}
Let $\B{X}$ be a solution of an \CDM\ $\C{R}$. 
If $\B{X}$ equilibrates, then 
  $\lim_{t\to\infty} \widebar{X}_i^{(n_i)} = (X_i^*,0,\dots,0)$ a.s.\ for all $i\in\C{I}$, where $X_i^*$ is the $i^{\text{th}}$ component of the corresponding equilibrium state $\B{X}^*$.
\end{proposition}

In particular, for linear \CDM s we can show that all the solutions of the \CDM\ equilibrate under certain conditions.
\begin{proposition}
\label{prop:StabilityResult}
  Let $\C{R}$ be a linear \CDM\ that satisfies Assumption~\asref{ass:ExplicitlySolvable}{(I\subseteq\C{I})} for a subset $I\subseteq\C{I}$ with an order tuple $\B{n}_I = 1$ and an exogenous process $\B{E}$ that is constant in time.\footnote{In general, we can let $\B{E}$ be a continuous exogenous process that depends on time as long as both $\B{E}_t$ and $\exp(At)\int_{t_0}^t \exp(-As)C\B{E}(s) ds$ converge almost surely for $t\to\infty$, where $A:=B_{II'}^{-1}(B_{IJ}B_{JJ}^{-1}B_{JI}-B_{II}+\mathbb{I}_I)$ and $C:=B_{II'}^{-1}(B_{IJ}B_{JJ}^{-1}\Gamma_{J\C{J}} - \Gamma_{I\C{J}})$. In that case, the order tuple may matter, and it must be checked whether the solutions are sufficiently smooth.} By Proposition~\ref{prop:LinearCDMExplicitlySolvable}, the dynamical causal mechanism $\B{f}$ is of the form
$$
  \left\{\begin{aligned}
    \B{f}_I(\widebar{\B{x}}^{(\B{n})},\B{e}) &:= B_{I I'} \B{x}'_I + B_{II}\B{x}_I + 
    B_{IJ}\B{x}_J + \Gamma_{I \C{J}}\B{e} \\
    \B{f}_J(\widebar{\B{x}}^{(\B{n})},\B{e}) &:= B_{JI}\B{x}_I + B_{JJ} \B{x}_J + \B{x}_J + \Gamma_{J \C{J}}\B{e} \,,
  \end{aligned}\right.
$$
  where $J:=\C{I}\setminus I$ and $B_{II'}$ and $B_{JJ}$ are invertible matrices. If the matrix $B_{II'}^{-1}(B_{IJ}B_{JJ}^{-1}B_{JI}-B_{II}+\mathbb{I}_I)$, where $\mathbb{I}_I$ denotes the identity matrix, is Hurwitz (that is, every eigenvalue has a strictly negative real part), then every solution $\B{X}$ of $\C{R}$ equilibrates to the same equilibrium state, irrespective of the initial condition.
\end{proposition}

This proposition allows us to derive a condition for which the price, supply and demand model always reaches market equilibrium.
\begin{example}[Market equilibrium, continued]
Applying Proposition~\ref{prop:StabilityResult} to the price, supply and demand model of
Example~\ref{ex:SupplyDemandModel} shows that $B_{II'}^{-1}(B_{IJ}B_{JJ}^{-1}B_{JI}-B_{II}+\mathbb{I}_I) = \lambda (\beta_D - \beta_S)$. This matrix is Hurwitz if and only if $\lambda(\beta_D - \beta_S) < 0$. Thus, since $\lambda > 0$,
the price $X_P$, supply $X_S$ and demand $X_D$ equilibrate for constant exogenous processes $E_D$ and $E_S$ if $\beta_S > \beta_D$.
\end{example}

\subsection{Steady \CDM s}
\label{sec:SteadyCDM}

In this subsection, we define the class of steady \CDM s which have the convenient property that their dynamics become explicitly time-independent asymptotically for $t \to \infty$. 
\begin{definition}[Steady \CDM]
\label{def:SteadyCDM}
We call an \CDM\ $\C{R}$ \emph{steady}, if it has a dynamic causal mechanism $\B{f}$ that is continuous and an exogenous process $\B{E}$ that is a.s.\ convergent.
\end{definition}

The continuity of the dynamic causal mechanism and the convergence assumption on the exogenous process assure us that the equilibrium states satisfy asymptotic \SDEfull s.
\begin{lemma}
\label{lemm:EquilibratingSolutionsCDM}
Let $\C{R}$ be a steady \CDM\ and let $\B{E}^*$ be the random variable to which the exogenous process $\B{E}$ converges a.s.. If $\B{X}$ is an equilibrating solution of a steady \CDM\ $\C{R}$, then the random variable $\widebar{\B{X}}^{(\B{n})*}$ to which the complete $\B{n}^{\text{th}}$-order derivative $\widebar{\B{X}}^{(\B{n})}$ converges satisfies 
$$
\B{X}^* = \B{f}(\widebar{\B{X}}^{(\B{n})*},\B{E}^*) \quad\text{a.s..}
$$
\end{lemma}
In general, not all solutions of a steady \CDM\ have to be equilibrating solutions, as one sees for example in Example~\ref{ex:HigherOrderRandomInitialConditions}.

The class of steady \CDM s is not closed under stochastic perfect interventions, since performing a stochastic perfect intervention that is not a.s.\ convergent yields an \CDM\ that is not steady. However, the class of steady \CDM s is closed under the following class of interventions.
\begin{definition}[Steady stochastic perfect intervention]
We call a stochastic perfect intervention $\intervene(I,\B{K}_I)$ a \emph{steady stochastic perfect intervention} if the process $\B{K}_I$ converges a.s.\ to a random variable $\B{K}_I^*$. We call it a \emph{steady perfect intervention} if in addition $\B{K}_I^*\in\BC{X}_I$ (that is, it does not depend on $\omega$).
\end{definition}

\subsection{Equilibration of a steady \CDM}
\label{sec:Equilibration}

In this subsection, we show how we can equilibrate a steady \CDM\ to an SCM, such that the equilibrium states of the \CDM\ are described by the SCM.
In the previous subsections, we saw that for an equilibrating solution of a steady \CDM, all the higher-order derivatives converge to zero, and the corresponding equilibrium state satisfies the asymptotic \SDEfull s. Hence, we can construct an SCM from a steady \CDM\ such that every equilibrium state of the steady \CDM\ is a solution of this SCM.
\begin{definition}[Equilibration of an \CDM]
\label{def:EquilibrationCDM}
Let $\C{R} = \langle \C{I}, \C{J}, \BC{X}, \BC{E}, \B{n}, \B{f}, \B{E} \rangle$ be a steady \CDM\ and let $\B{E}^*$ be a random variable such that $\B{E}$ converges a.s.\ to it. We call the SCM $\C{M}_{\C{R}} := \langle \C{I}, \C{J}, \BC{X}, \BC{E}, \B{f}^*, \B{E}^* \rangle$ with the \emph{equilibrated dynamic causal mechanism} $\B{f}^* : \BC{X}\times\BC{E}\to\BC{X}$ given by
$$
\B{f}^*(\B{x},\B{e}) := \B{f}(\widebar{\B{\iota}}(\B{x}),\B{e}) \,,
$$
an \emph{equilibration} of $\C{R}$,
where the mapping $\widebar{\B{\iota}} : \BC{X} \to \BC{X}^{\B{n}+1}$ defined by 
$$
\widebar{\iota}_i^{(k_i)}(\B{x}) = 
\begin{cases}
  x_i & \text{if } k_i=0 \\
  0 &\text{otherwise,}
\end{cases}
$$
is the embedding that sets all the higher-order derivatives of the endogenous processes to $0$.
\end{definition}
In other words, the equilibration of an \CDM\ sets all the higher-order derivative entries in its dynamic causal mechanism to zero and replaces its exogenous process by its limiting random variable. In particular, linearity is preserved under equilibration, that is, a steady linear \CDM\ equilibrates to a linear SCM.

The equilibration of an \CDM\ is well defined due to the following result, which shows that the independence property for the family of exogenous processes $(\B{E}_j)_{j\in\C{J}}$ is preserved in the limit when time tends to infinity.
\begin{proposition}
\label{prop:ConvergenceOfIndependentProcesses}
Let $(\B{E}_j)_{j\in\C{J}}$ be a family of stochastic processes, where $\C{J}$ is some finite index set, such that $\B{E}_j$ converges almost surely to the random variable $\B{E}_j^*$, for every $j\in\C{J}$. Then, if $(\B{E}_j)_{j\in\C{J}}$ is independent, so is the family of random variables $(\B{E}_j^*)_{j\in\C{J}}$.
\end{proposition}

This equilibration of an \CDM\ to an SCM leads to the main insight that SCMs are capable of modeling all the equilibrium states of steady \CDM s.
\begin{theorem}
\label{thm:EquilibratingSolutionsSCM}
If $\B{X}$ is an equilibrating solution of a steady \CDM\ $\C{R}$, then its limit $\B{X}^*$ is a solution of the corresponding equilibration $\C{M}_{\C{R}}$.
\end{theorem}
Intuitively, the equilibration of a steady \CDM\ to an SCM can be seen as the approximation of the \SDEfull s by the structural equations of the SCM, which becomes exact at equilibrium. This is illustrated in the following example.
\begin{example}[Equilibrated damped coupled harmonic oscillator]
\label{ex:HarmonicOscillatorEquilibration}
Consider the intervened damped coupled harmonic oscillator of
Example~\ref{ex:HarmonicOscillatorIntervened} for which the \SDEfull s are specified by
$$
  \left\{\begin{aligned}
   X_1 &= 0 \\
   X_i &= \frac{\kappa_i}{\kappa_i + \kappa_{i-1}} (X_{i+1} - L_i) + \frac{\kappa_{i-1}}{\kappa_i + \kappa_{i-1}} (X_{i-1} + L_{i-1}) \\
              & \phantom{:=} -\frac{b_i}{\kappa_i + \kappa_{i-1}} X_i' - \frac{m_i}{\kappa_i + \kappa_{i-1}} X_i'' \quad\quad(i=2,\dots,d-1)\\
   X_d  &= L \,,
  \end{aligned}\right.
$$
and where the exogenous processes $\B{L}$ are random variables. In the limit, as time tends to infinity, the equilibrating solutions of the \CDM\ converge to the equilibrium states of the equilibrated \CDM, which can be obtained by setting the higher-order derivatives to zero. This yields the equations
$$
  \left\{\begin{aligned}
   X_1^* &= 0 \\
   X_i^* &= \frac{\kappa_i(X_{i+1}^* - L_i) + \kappa_{i-1}(X_{i-1}^* + L_{i-1})}{\kappa_i + \kappa_{i-1}} \quad\quad(i=2,\dots,d-1)\\
   X_d^*  &= L \,,
  \end{aligned}\right.
$$
which describe the equilibrium states for the positions of the masses. Not all solutions necessarily equilibrate to an equilibrium, which happens for example in the case when there is no friction, that is, $b_i= 0$ for all $i \in \{2,\dots,d-1\}$. In this case, if any mass $m_i$ starts at an off-equilibrium position (that is, if $X_{i}'(t_0)\neq 0$ or $X_{i}(t_0) \neq X_i^*$ for some $i \in \{2,\dots,d-1\}$), the solution will not equilibrate, but will keep on oscillating forever.
\end{example}
In case there is friction and the exogenous processes $\B{L}$ are fixed to constant values, the equilibrated damped coupled harmonic oscillator exactly coincides with the deterministic SCM derived in~\citep{MJS13}. In Section~\ref{sec:CommutationEquilibrationIntervention} we will show that the equilibration operation, as defined in Definition~\ref{def:EquilibrationCDM}, also preserves the causal semantics. The next example illustrates that our equilibration operation can also be applied to models that cannot be treated with the theory of \citep{MJS13}. 
\begin{example}[Equilibrated price, supply and demand model]
\label{ex:SupplyDemandModelEquilibrated}
Setting the higher-order derivatives of the price, supply and demand model $\C{R}$ of Example~\ref{ex:SupplyDemandModel} to zero yields the structural equations:
$$
  \left\{\begin{aligned}
    X_P^*  &= X_P^* + \lambda (X_D^* - X_S^*) \\
    X_S^*  &= \beta_S X_P^* + E_S^* \\
    X_D^*  &= \beta_D X_P^* + E_D^* \,.
  \end{aligned}\right.
$$
The equations describe the market equilibrium states. In Figure~\ref{fig:SupplyDemandModelSimulation}, we simulate the solutions of the \CDM\ $\C{R}$ for random constant exogenous influences $E_S$ and $E_D$ and random consistent initial conditions. The dispersion of $X_P$, $X_S$ and $X_D$ at large $t$ illustrates that the equilibrium state is not unique and depends on the initial condition. Hence, this example cannot be treated with the theory of \citep{MJS13}. 
\begin{figure*}
\begin{center}
\includegraphics[width=0.32\linewidth]{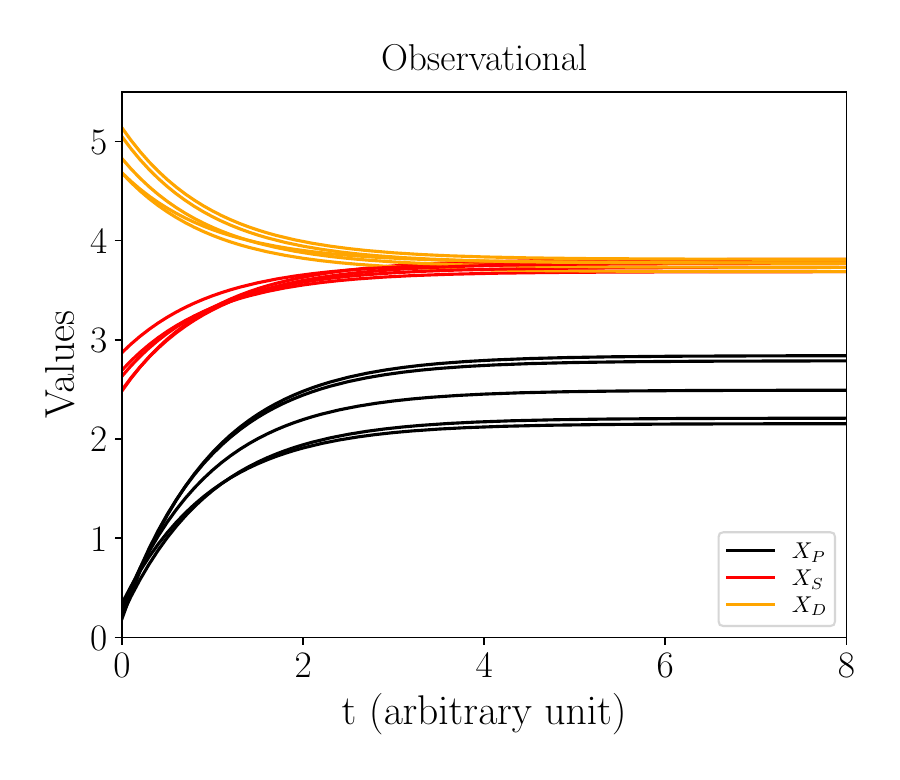}
\includegraphics[width=0.32\linewidth]{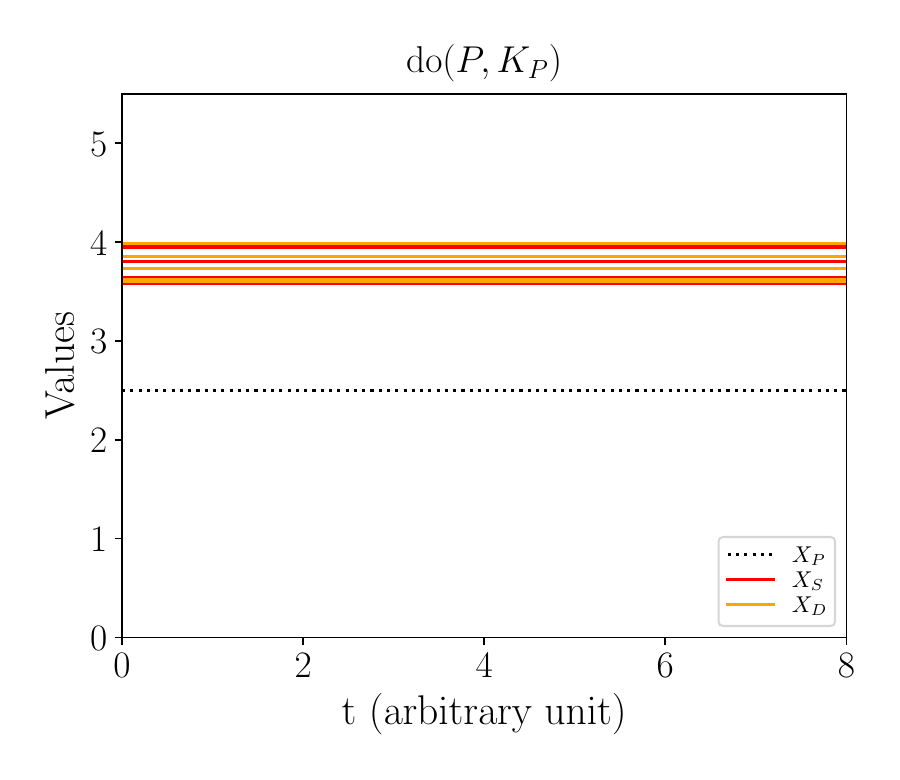}
\includegraphics[width=0.32\linewidth]{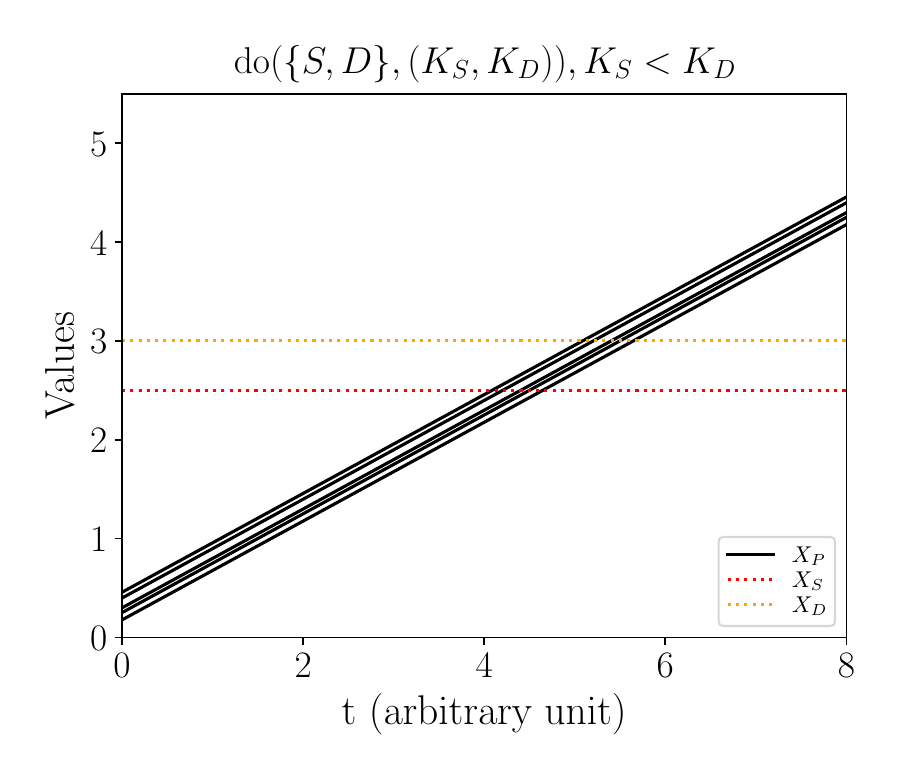}
\end{center}
\vspace{-1.5\baselineskip}
\caption{Simulation of solutions of the \CDM\ of the price, supply and demand model of Example~\ref{ex:SupplyDemandModelEquilibrated} under different steady perfect interventions.}
\label{fig:SupplyDemandModelSimulation}
\end{figure*}
\end{example}
\citet{RichardsonRobins2014} argue that the price, supply and demand model cannot be modeled at equilibrium as an SCM without self-cycles. We conclude that it can be modeled by an SCM that contains self-cycles, with the corresponding graph depicted in Figure~\ref{fig:SupplyDemandExample} (right). 

A consequence of Theorem~\ref{thm:EquilibratingSolutionsSCM} is that if the SCM $\C{M}_{\C{R}}$ has no solutions, then the \CDM\ $\C{R}$ has no equilibrating solutions. However, the converse does not hold in general, as the following example illustrates.
\begin{example}
Let $\C{R} = \langle \{1,2\}, \{3\}, \BC{X}, \C{E}, \B{n}, \B{f}, E \rangle$ be the steady \CDM\ with $\BC{X}=\RN^2$, $\C{E}=\RN$, $\B{n}=(0,1)$, the dynamic causal mechanism $\B{f}$ given by $f_1(\widebar{\B{x}}^{(\B{n})},e) = x_{2'}$ and $f_2(\widebar{\B{x}}^{(\B{n})},e) = e$, and the exogenous process $E$ given by $E_t=\sin(t^3)/t$. The \SDEfull s associated to $\C{R}$ are given by
$$
X_1 = X_2'\,,\quad X_2 = E \,.
$$
This model can be equilibrated to the model $\C{M}_{\C{R}}$ with structural equations
$$
X^*_1 = 0\,,\quad X^*_2 = E^* \,,
$$
and exogenous variable $E^*=0$. Although the SCM $\C{M}_{\C{R}}$ clearly has a solution, the \CDM\ $\C{R}$ has no equilibrium states, since $X_1=X_2'=E'$ is not a.s.\ convergent.
\end{example}

The following result shows that if the exogenous process is constant in time, this cannot happen.
\begin{proposition}
\label{prop:NoEquilibratedInitialConditionNoEquilibriumSolution}
Let $\C{R}$ be a steady \CDM\ such that the exogenous process $\B{E}$ is a random variable (i.e., $\B{E}$ is constant in time). If the \CDM\ $\C{R}$ has no equilibrating solution, then its equilibration $\C{M}_{\C{R}}$ has no solutions.
\end{proposition}

\subsection{Graphs of the equilibrated \CDM}
\label{sec:GraphOfTheEquilibration}

In this subsection, we show how the equilibration operation acts on the (augmented) graph of the \CDM. 
%
%
%
%
%
%


\begin{proposition}[Graph of the equilibrated \CDM\ is a subgraph of the original mixed graph]
\label{prop:EquilibrationSubgraph}
  Let $\C{R}$ be a steady \CDM. The graph $\C{G}(\C{M}_{\C{R}})$  of the equilibrated \CDM\ $\C{M}_{\C{R}}$ is the mixed graph obtained from the graph $\C{G}(\C{R})$ of $\C{R}$ by removing the partition into clusters and removing the nodes $i^{(k_i)}$ for $i \in I$ and $k_i >0$ together with their adjacent edges.
An analogous statement holds for the augmented graph $\C{G}^a(\C{M}_{\C{R}})$.
\end{proposition}

The following example illustrates this for the equilibrated price, supply and demand model.
\begin{example}[Price, supply and demand, continued]
\label{ex:SupplyDemandModelGraph}
Consider the price, supply and demand model $\C{R}$ of Example~\ref{ex:SupplyDemandModel} for a very large $\lambda$,
that is, for which the price adjusts very quickly to changes in supply and demand. This system can
be approximated by the equilibrated price, supply and demand model $\C{M}_{\C{R}}$. The graph of this equilibrated model $\C{M}_{\C{R}}$ is a subgraph of the graph of the original model $\C{R}$, as can be seen in Figure~\ref{fig:SupplyDemandExample}.
\end{example}

\subsection{Equilibration commutes with intervention}
\label{sec:CommutationEquilibrationIntervention}

Theorem~\ref{thm:EquilibratingSolutionsSCM} states that the equilibrium states of a steady \CDM\ are solutions of the SCM to which the \CDM\ equilibrates. In the previous subsection, we showed, moreover, that the functional relationships between the endogenous processes that are encoded in the \SDEfull s are preserved under equilibration. This leads to another important result: the equilibration operation preserves the causal semantics of the equilibrium states, as is illustrated in Figure~\ref{fig:CommutingDiagramCDMSCM} in Section~\ref{sec:Intro}.

\begin{theorem}
\label{thm:InterventionCommutesWithEquilibrationCDMSCM}
Let $\C{R}$ be a steady \CDM\ and let $\intervene(I,\B{K}_I)$ be a steady stochastic
perfect intervention for some subset $I\subseteq\C{I}$ and stochastic process $\B{K}_I$ that converges a.s.\ to a random variable $\B{K}_I^*$. Then the steady stochastic perfect intervention commutes with equilibration, that is
$$
(\C{M}_{\C{R}})_{\intervene(I,\B{K}_I^*)} = 
  \C{M}_{(\C{R}_{\intervene(I,\B{K}_I)})} \,.
$$
\end{theorem}

This result allows us to perform causal reasoning on the equilibrium states of the \CDM\ by considering only the equilibrated model, as is illustrated in the following example.

\begin{example}[Bathtub model, continued]
\label{ex:Bathtub_do_D}
In Example~\ref{ex:Bathtub} we defined the \CDM\ for the bathtub model. The equilibrium states of this model can be described by the structural equations of the equilibrated model, as depicted in the top row of the following diagram.
\begin{center}
  \begin{tikzpicture}[auto]
    \node[text width=6cm,rectangle,draw=black,rounded corners,scale=0.9] (A) at (0,0) {steady \CDM\\[0.5\baselineskip]
    $\begin{cases}
        X_K     &= E_K \\
        X_{Q_i} &= E_{Q_i} \\
        X_P     &= \alpha_4 X_D - \alpha_2^{-1}X_P' \\
        X_{Q_o} &= \alpha_1 X_K X_P - \alpha_3^{-1}X_{Q_o}' \\
        X_D     &= X_D + \alpha_0(X_{Q_i} - X_{Q_o}) - X_D'
    \end{cases}$};
      \node[text width=6cm,right of=A,node distance=7.5cm,rectangle,draw=black,rounded corners,scale=0.9] (B) {equilibrated \CDM\\[0.5\baselineskip]
      $\begin{cases}
        X_K^* &= E_K^* \\
        X_{Q_i}^* &= E_{Q_i}^* \\
        X_P^* &= \alpha_4 X_D^* \\
        X_{Q_o}^* &= \alpha_1 X_K^* X_P^* \\
        X_D^* &= X_D^* + \alpha_0(X_{Q_i}^* - X_{Q_o}^*)
      \end{cases}$};
    \node[text width=6cm,below of=A,node distance=5cm,rectangle,draw=black,rounded corners,scale=0.9] (A2) {intervened steady \CDM\\[0.5\baselineskip]
      $\begin{cases}
        X_K     &= E_K \\
        X_{Q_i} &= E_{Q_i} \\
        X_P     &= \alpha_4 X_D - \alpha_2^{-1}X_P' \\
        X_{Q_o} &= \alpha_1 X_K X_P - \alpha_3^{-1}X_{Q_o}' \\
        X_D     &= K_D
      \end{cases}$};
    \node[text width=6cm,right of=A2,node distance=7.5cm,rectangle,draw=black,rounded corners,scale=0.9] (B2) {intervened  and equilibrated \CDM\\[0.5\baselineskip]
    $\begin{cases}
      X_K^* &= E_K^* \\
      X_{Q_i}^* &= E_{Q_i}^* \\
      X_P^* &= \alpha_4 X_D^* \\
      X_{Q_o}^* &= \alpha_1 X_K^* X_P^* \\
      X_D^* &= K_D^*
    \end{cases}$};
   \draw[|->,shorten >=0.1cm,shorten <=0.1cm,line width=0.2mm] (A) to node {\small $t \rightarrow \infty$} (B);
    \draw[|->,shorten >=0.1cm,shorten <=0.1cm,line width=0.2mm] (A2) to node {\small $t \rightarrow \infty$} (B2);
    \draw[|->,shorten >=0.1cm,shorten <=0.1cm,line width=0.2mm] (A) to node {\small $\intervene(D,K_D)$} (A2);
    \draw[|->,shorten >=0.1cm,shorten <=0.1cm,line width=0.2mm] (B) to node {\small $\intervene(D,K_D^*)$} (B2);
  \end{tikzpicture}
\end{center}
After equilibration, one can perform causal reasoning on the level of the equilibrated \CDM, without needing to resort to the original \CDM\ description. Indeed, we see in the above diagram that it doesn't matter whether we first perform the steady stochastic perfect intervention $\intervene(D,K_D)$, and then let the system equilibrate, or the other way around. The graphs of the \CDM, the equilibrated \CDM\ and their corresponding intervened models are depicted in Figure~\ref{fig:Bathtub}. Choosing different a.s.\ convergent processes for $K_D$ yields different solution processes
$X_P$ and $X_{Q_o}$ of the intervened \CDM, but the solution processes for $X_{Q_i}$ and
$X_K$ stay unchanged. Similarly, the perfect intervention $\intervene(D,K_D^*)$ on the equilibrated \CDM\ yields different solutions $X_P^*$ and $X_{Q_o}^*$ of the intervened SCM depending on the value of $K_D^*$, but does not change the solutions $X_{Q_i}^*$ and $X_K^*$. This behavior is also reflected in the graphs depicted in the bottom row of Figure~\ref{fig:Bathtub}.

Intuitively, one would indeed expect the chosen intervention value for the depth to have an effect
on pressure and outflow (but not on inflow or drain size) at equilibrium. For example, one could (approximately) implement such a perfect intervention by adding a water level control device that constantly monitors the level and that can pump water in and out of the bathtub via a hose, regulating the depth at $K_D$ at all times by using an optimal control feedback loop, independently of the exogenous processes $E_K$ and $E_{Q_i}$. Indeed, the depth directly determines the pressure $X_P$
exerted by the water in the bathtub at the drain, and the outflow rate $X_{Q_o}$ is a direct 
consequence of that. Once the other processes in the system have equilibrated, the
processes $X_P$ and $X_{Q_o}$ will also equilibrate to random variables that depend on
$K_D^*$. The inflow $X_{Q_i}^*$ of water through the faucet no longer 
needs to be equal to the outflow $X_{Q_o}^*$ through the drain at equilibrium because 
water is also constantly added or removed via the hose by the water level control device
in order to maintain the (eventually) constant depth $K_D^*$.\footnote{At equilibrium, the total inflow of water through the faucet and the hose has to be equal to the total outflow through the drain and the hose.}
\end{example}
This sheds some new light on the violation of the equilibration-manipulation commutability property (the ``EMC-property'') of \citet{Das05}, who shows the---at first sight contradictory---result that equilibration does not always commute with intervention. The paradox is resolved by noting that \citet{Das05} defines a different notion of ``equilibration'', inspired by \citet{IS94}, for which commutativity with perfect intervention indeed does not always hold. One can readily verify that the ``equilibration'' operation
of \citet{Das05} does not preserve the functional relationships between the endogenous processes that are encoded in the equations under the equilibration. Recently, \citet{BM21} showed that the ``equilibration'' operation of \citet{Das05} maps an \CDM\ $\C{R}$ to a Markov ordering graph that encodes the conditional independencies in solutions of $\C{M}_{\C{R}}$ instead of the functional relationships. In contrast, our equilibration operation, defined in Definition~\ref{def:EquilibrationCDM}, preserves the functional relationships between the endogenous processes, since each \SDEfull\ equilibrates to a structural equation associated to the same endogenous process/variable. This is also reflected in Proposition~\ref{prop:EquilibrationSubgraph} where we showed that the graph of the equilibrated \CDM\ is a subgraph of the mixed graph of the \CDM.

Theorem~\ref{thm:EquilibratingSolutionsSCM} and \ref{thm:InterventionCommutesWithEquilibrationCDMSCM} together imply that our equilibration operation preserves the equilibrium states of a steady \CDM\, while also preserving the causal semantics. In particular, we do not require that all solutions of the steady \CDM\ have to equilibrate. As a consequence, the equilibrium states of the model may depend on the (consistent) initial conditions. This is in contrast to the work of \citet{MJS13}, who assume that the equilibrium state of the dynamical system is unique and independent of the initial condition. This is a strong assumption that limits the applicability of the theory, since this does not allow for any stochasticity at equilibrium. Indeed, many random dynamical systems have multiple equilibrium states that depend on the chosen initial condition, as is illustrated in the following example.
\begin{example}[Bathtub model, continued]
\label{ex:BathtubSimulation}
Consider again the bathtub model $\C{R}$ of Example~\ref{ex:Bathtub}. Figure~\ref{fig:BathtubSimulation} (top left) illustrates some numerical solutions of the \SDE s, with $\B{\alpha}=(1,1,1,1,4/5)$, $E_K=1/2,E_{Q_i}=1$ and for randomly drawn consistent initial conditions $(0,\B{X}_{[0]})$ of order $\B{0}$. We see that the solutions equilibrate to the a.s.\ unique equilibrium state $(X_K^*,X_{Q_i}^*,X_P^*,X_{Q_o}^*,X_D^*) = (1/2,1,2,5/2,1)$ corresponding to the solution of the equilibrated \CDM\ $\C{M}_{\C{R}}$. If we now perform the perfect intervention $\intervene(Q_o,K_{Q_o})$ on the system $\C{R}$, where we force the water outflow $X_{Q_o}$ to be equal to the water inflow $X_{Q_i}$ at all time, that is, $K_{Q_o}=E_{Q_i}$, then this does not give an a.s.\ unique equilibrium state, but the equilibrium state that is obtained depends on the initial condition, as can be seen in Figure~\ref{fig:BathtubSimulation} (top center). Indeed, the depth $X_D^*$ at equilibrium will equal the initial depth $X_{[0],D}$ at $t_0=0$, if the inflow $X_{Q_i}$ equals the outflow $X_{Q_o}$. This example cannot be treated with the theory of \citep{MJS13}, which assumes that the equilibrium state is unique and does not depend on the initial condition. However, if instead we perform the perfect intervention $\intervene(Q_o,K_{Q_o})$ on $\C{R}$ where $K_{Q_o} < E_{Q_i}$, then the depth $X_D$ will not reach equilibrium, but will increase indefinitely, since the rate of water flowing into the bathtub is larger than the outflow rate. This is illustrated in Figure~\ref{fig:BathtubSimulation} (top right). This is also reflected in the equilibrated \CDM\ $\C{M}_{\C{R}}$, which does not have any solution after the corresponding perfect intervention $\intervene(Q_o,K_{Q_o})$.
\begin{figure*}
\begin{center}
\includegraphics[width=0.32\linewidth]{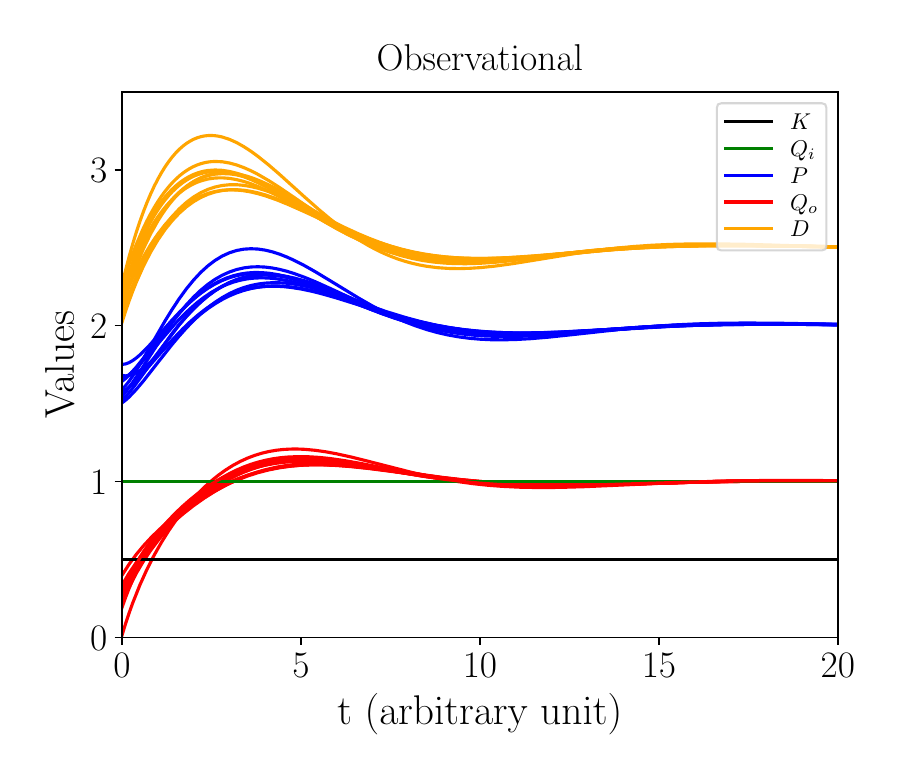}
\includegraphics[width=0.32\linewidth]{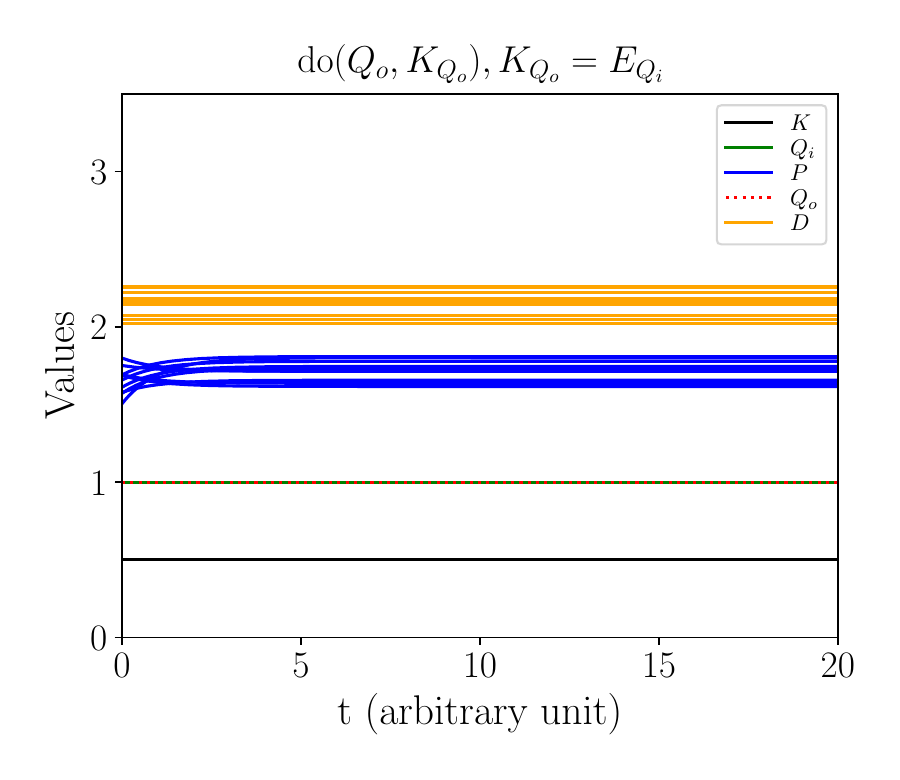}
\includegraphics[width=0.32\linewidth]{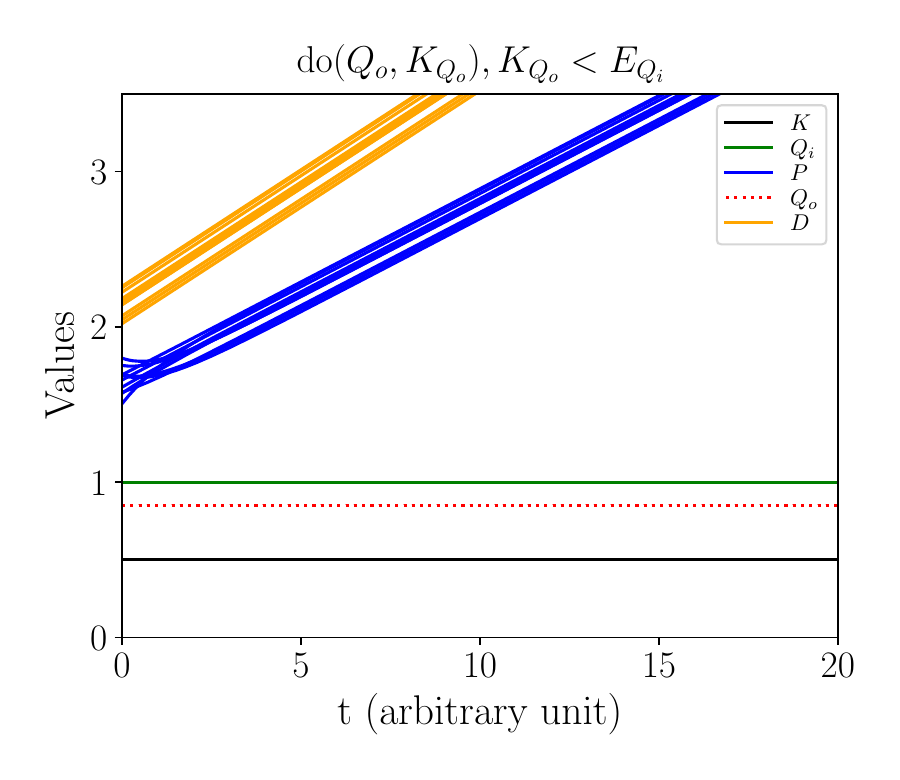}
\end{center}

\vspace{-1.0cm}
\begin{center}
\includegraphics[width=0.32\linewidth]{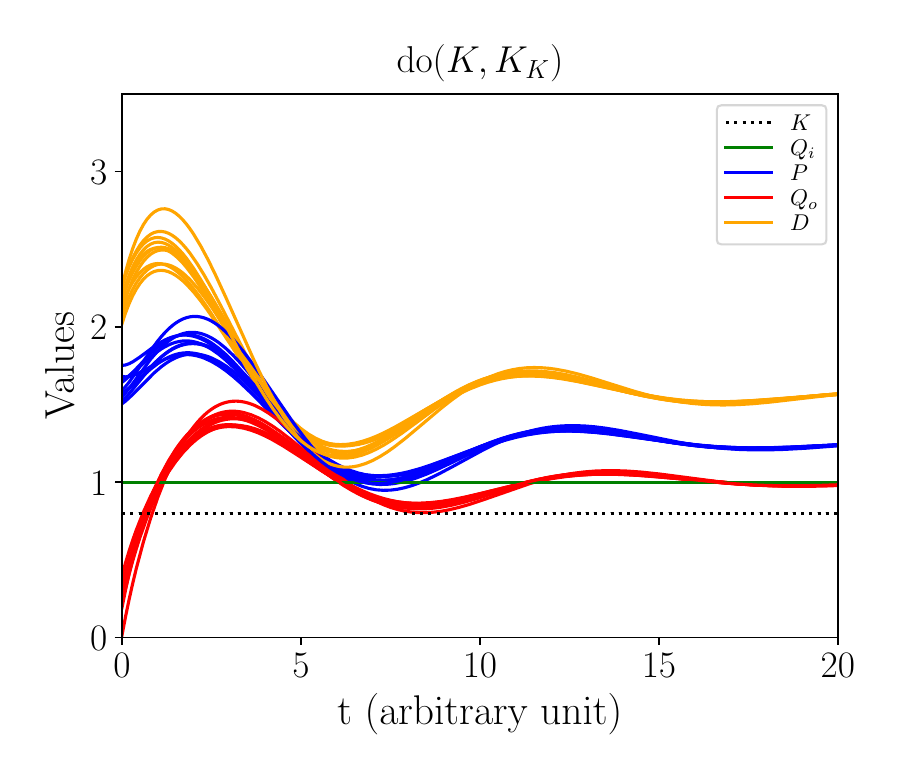}
\includegraphics[width=0.32\linewidth]{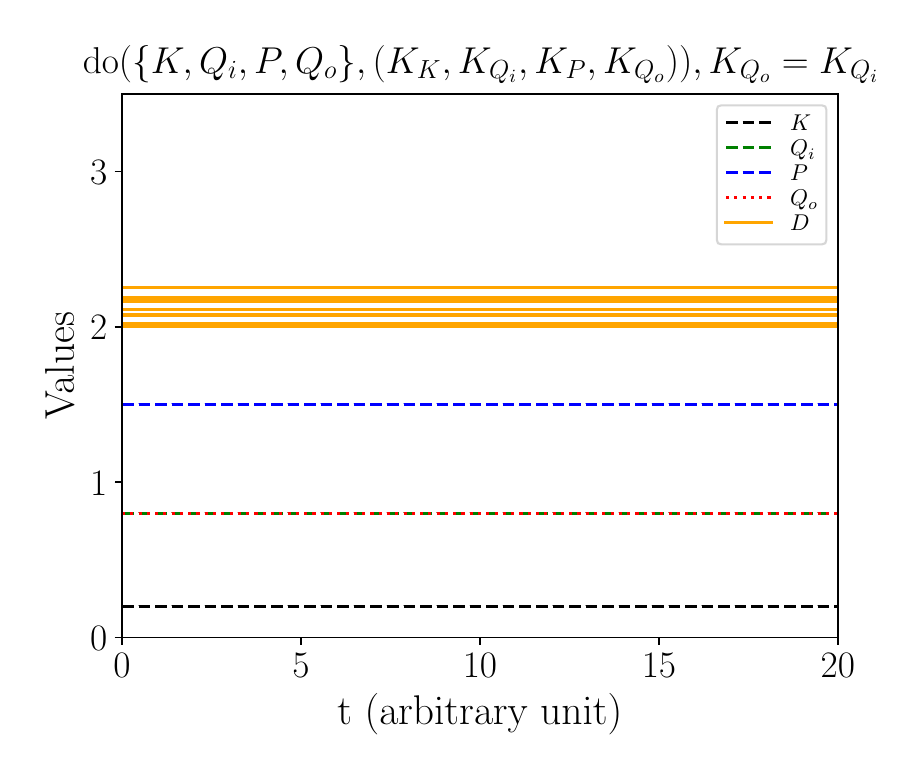}
\includegraphics[width=0.32\linewidth]{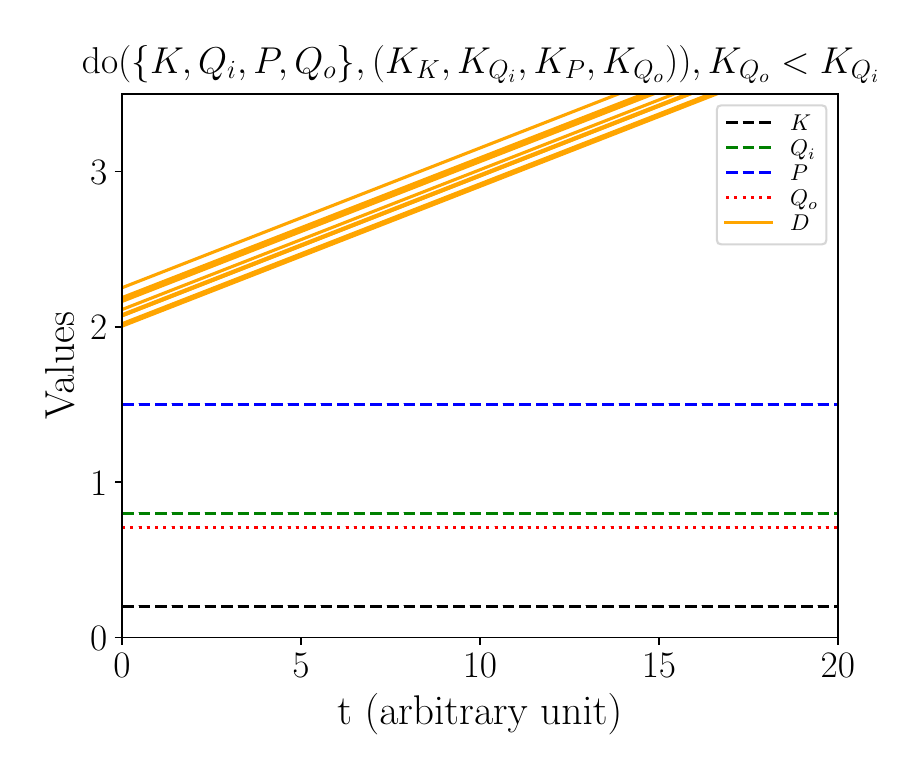}
\end{center}
\vspace{-1.5\baselineskip}
\caption{Simulation of solutions of the \CDM\ of the bathtub model of Example~\ref{ex:BathtubSimulation} and \ref{ex:BathtubCausalRelations} under different steady perfect interventions.}
\label{fig:BathtubSimulation}
\end{figure*}
\end{example}
Similar behavior is observed for the equilibrium states of the price, supply and demand model $\C{R}$ of Example~\ref{ex:SupplyDemandModelEquilibrated}. For example, the model $\C{R}$ will reach market equilibrium if one holds the price fixed at all times by the perfect intervention $\intervene(P,K_P)$, but will not reach equilibrium if the supply and demand are fixed at all times by the perfect intervention $\intervene(\{S,D\},(K_S,K_D))$ for which $K_S < K_D$ (see Figure~\ref{fig:SupplyDemandModelSimulation} center and right respectively). In all the cases depicted in Figure~\ref{fig:SupplyDemandModelSimulation} we see a dependence of the equilibrium states on the initial condition.

In summary, the equilibration of a steady \CDM\ to an SCM generalizes the work of \citep{MJS13} in three directions: (i) the deterministic setting is replaced with a more general stochastic setting, (ii) the \SDEfull s can be of arbitrary order (including zeroth-order), rather than only first-order, which prevents complications with the causal interpretation (see, for example, Example~\ref{ex:HigherOrderDerivativeAsEndogenousProcess}), and (iii) the equilibrium state is allowed to depend on initial conditions.
Together, this substantially extends the applicability of the theory.

\subsection{Realizing a given SCM as a stable \CDM}
\label{sec:NonRecoverabilityCDMfromSCM}

Although each steady \CDM\ equilibrates to an SCM, not all solutions of the \CDM\ need to equilibrate to solutions of the corresponding SCM (see, for example, Example~\ref{ex:HarmonicOscillatorEquilibration}). In this subsection, we address the inverse problem of finding steady \CDM s with non-trivial dynamics for which all solutions equilibrate to solutions of a specified SCM. This can be thought of as realizing the given SCM as a ``stable'' \CDM.
In Proposition~\ref{prop:StabilityResult} we provided certain conditions under which all the solutions of a linear \CDM\ equilibrate. Based on this result and some results in the linear systems theory literature, we show that for a certain class of SCMs one can construct a first-order \CDM\ such that \emph{all} its solutions equilibrate to the solutions of the SCM. Moreover, we show that under certain stronger conditions, the \CDM\ can be chosen such that its solutions still equilibrate to the solutions of the intervened SCM after any constant stochastic perfect intervention. Hence, the constructed \CDM\ realizes the causal semantics of the SCM at equilibrium.

First, we observe that one cannot uniquely recover an \CDM\ from its equilibration in general.
\begin{example}
\label{ex:stabilize_linear_CDM}
Consider the linear \CDM\ $\C{R}$ with \SDE\ given by
$$
\B{X} = B\B{X} - \B{X}' + \Gamma \B{E} \,, 
$$
where the matrix $A:=\mathbb{I}-B$ is invertible and the exogenous process $\B{E}$ is a random variable. Consider another \CDM\ $\tilde{\C{R}}$ which differs only in its dynamic causal mechanism, and has the \SDE
\begin{equation}
\label{eq:simple_linear_CDM} 
\B{X} = B\B{X} - \Lambda\B{X}' + \Gamma \B{E} \,,
\end{equation}
where $\Lambda$ is some invertible diagonal matrix. The equilibrated \CDM s $\C{M}_{\C{R}}$
and $\C{M}_{\tilde{\C{R}}}$ coincide, and have structural equations of the form
\begin{equation}
\label{eq:SCM_of_linear_CDM} 
  \B{X}^* = B\B{X}^* + \Gamma \B{E}^* \,.
\end{equation}
Hence, the equilibrium states $\B{X}^*$ of the \CDM s $\C{R}$ and $\tilde{\C{R}}$ are indistinguishable, since both have to satisfy $\B{X}^*=A^{-1}\Gamma\B{E}^*$ a.s..
  Furthermore, if the matrix $\mathbb{I}_I-B_{II}$ is invertible for some subset $I\subseteq\C{I}$, then also for $J:=\C{I}\setminus I$ the intervened equilibrium states of $\C{R}_{\intervene(J,\B{K}_J)}$ and $\tilde{\C{R}}_{\intervene(J,\B{K}_J)}$ are indistinguishable for any sufficiently smooth steady stochastic intervention $\intervene(J,\B{K}_J)$.
\end{example}
Although the equilibrated \CDM\ in Example~\ref{ex:stabilize_linear_CDM} describes the possible equilibrium state of both \CDM s, it is not necessarily guaranteed that the solutions of both \CDM s equilibrate. 
One might hope that for any given linear SCM of the form (\ref{eq:SCM_of_linear_CDM}), one can always find an invertible diagonal matrix $\Lambda$ such that one can construct a steady \CDM\ of the form \eqref{eq:simple_linear_CDM} for which all solutions of the \CDM\ equilibrate to the (a.s.\ unique) solution of the SCM (see Proposition~\ref{prop:StabilityResult}).
Such a ``stabilization matrix'' $\Lambda$ does not always exist. A sufficient condition for its existence was given in \citep{Fis58,Fis72}, leading to the following result.
\begin{corollary}
\label{corr:FisherSufficientCondition}
Let $\C{M}$ be a linear SCM with structural equations
$$
\B{X} = B\B{X} + \Gamma \B{E} \,,
$$
where $\C{I}=\{1,\dots,d\}$, $\C{J}=\{1,\dots, e\}$, $B \in \RN^{d \times d}$, $\Gamma \in \RN^{d\times e}$,
and with $\B{E}$ a random variable. Write $A := \mathbb{I} - B$.
For a diagonal matrix $\Lambda \in \RN^{d\times d}$, consider the linear \CDM\ $\C{R}_{\C{M},\Lambda}$ with \SDE\ of the form
$$
\B{X} = B \B{X} - \Lambda\B{X}' + \Gamma \B{E}.
$$
If there exists a sequence of matrices $M_d, M_{d-1}, \dots, M_1$ with $M_d = A$ such that for $k=2,\dots,d$ each $M_{k-1}$ is a principal $(k-1)\times(k-1)$ submatrix of $M_k$, with $\det M_k \ne 0$
for all $k=1,\dots,d$, then there exists 
 a diagonal stabilization matrix $\Lambda \in \RN^{d\times d}$ such that the linear \CDM\ $\C{R}_{\C{M},\Lambda}$
has the properties that (i) its equilibrated \CDM\ is $\C{M}_{\C{R}_{\C{M},\Lambda}} = \C{M}$, and
  (ii) all its solutions equilibrate to an a.s.\ unique equilibrium state that satisfies the structural equations of the SCM $\C{M}$, independent of the initial condition.
\end{corollary}
While this sufficient condition guarantees the existence of a stabilization matrix $\Lambda$ such that the \emph{observational} equilibrium distribution of the SCM is recovered as the distribution of the equilibrium state of the \CDM,
it does not guarantee that after a stochastic perfect intervention on the \CDM, all solutions will equilibrate to an (a.s.\ unique) equilibrium solution of the corresponding intervened SCM.
Indeed, a certain $\Lambda$ that stabilizes the dynamics in the absence of the intervention may no longer stabilize the dynamics after the intervention has been carried out.
Can we, under some conditions, find a single $\Lambda$ that will stabilize the dynamics after \emph{any} stochastic perfect intervention?
The answer is affirmative, as was shown by \citet{LS12} who provide a necessary and sufficient condition
for the existence of an invertible diagonal stabilization matrix $\Lambda$ that simultaneously stabilizes all subsystems.\footnote{\citet{LS12} consider an extension of the stabilization problem studied by \citet{Fis58}. Whereas \citet{Fis58} consider the problem of finding a diagonal matrix $\Lambda\in\RN^{d\times d}$ for a matrix $A\in\RN^{d\times d}$ such that the matrix $\Lambda A$ is Hurwitz (for which they provide a sufficient condition), \citet{LS12} consider the case where \emph{all} the principal submatrices of $\Lambda A$ should be Hurwitz, and provide a condition that is both sufficient and necessary, as well as a construction of such a stabilization matrix $\Lambda$.} 
This leads to the following result on how one can ``realize'' a given linear SCM as a stable linear first-order \CDM.
\begin{corollary}
\label{corr:RealizingLinearSCMAsCDM}
Let $\C{M}$ be a linear SCM with structural equations
$$
\B{X} = B\B{X} + \Gamma \B{E} \,,
$$
where $\C{I}=\{1,\dots,d\}$, $\C{J}=\{1,\dots, e\}$, $B \in \RN^{d \times d}$, $\Gamma \in \RN^{d\times e}$,
and with $\B{E}$ a random variable. Write $A := \mathbb{I} - B$.
For a diagonal matrix $\Lambda \in \RN^{d\times d}$, consider the linear \CDM\ $\C{R}_{\C{M},\Lambda}$ with \SDE\ of the form
$$
\B{X} = B \B{X} - \Lambda\B{X}' + \Gamma \B{E}.
$$
If 
  \begin{equation}\label{eq:LocatelliSchiavoni}
    \det(A_{II})\det(\mathrm{diag}(A_{II}))>0 \qquad \forall I \subseteq \C{I} \,,
  \end{equation}
then there exists 
 an invertible diagonal stabilization matrix $\Lambda \in \RN^{d\times d}$ such that the linear \CDM\ $\C{R}_{\C{M},\Lambda}$
has the properties that (i) its equilibrated \CDM\ is $\C{M}_{\C{R}_{\C{M},\Lambda}} = \C{M}$, and
(ii) under every stochastic perfect intervention $\intervene(J,\B{K}_J)$ with the exogenous process $\B{K}_J$ constant in time, all solutions of $(\C{R}_{\C{M},\Lambda})_{\intervene(J,\B{K}_J)}$ equilibrate to an a.s.\ unique equilibrium state that is the unique solution of the SCM $\C{M}_{\intervene(J,\B{K}_J^*)}$, independent of the initial condition.
\end{corollary}
Condition (\ref{eq:LocatelliSchiavoni}) implies that the matrices $\mathbb{I}_I - B_{II}$ are invertible for every subset $I\subseteq\C{I}$.
Such linear SCMs are special cases of the class of \emph{simple} SCMs (see \citep{BPSM19}). 
Simple SCMs have the convenient property that their solutions are a.s.\ unique after any stochastic perfect intervention. 
We conclude that for the subclass of simple linear SCMs that satisfy condition (\ref{eq:LocatelliSchiavoni}), we can construct a linear first-order \CDM\ whose causal semantics at equilibrium ``realizes'' that described by the SCM.
We speculate that this result can be extended to higher-order and nonlinear systems, but we will not pursue these questions here.


\begin{example}
  We show that the equilibrated \CDM\ of Example~\ref{ex:HarmonicOscillatorEquilibration} (see also Example~\ref{ex:HarmonicOscillator}), modeling the equilibrium states of a damped coupled harmonic oscillator, satisfies condition (\ref{eq:LocatelliSchiavoni}).
  Indeed, taking $\C{I} = \{1, \dots, d\}$, 
  the matrix $B$ of this linear SCM is tridiagonal, given as
  $$B = \left[\begin{array}{ccccc}
    0                     & \frac{\kappa_1}{\kappa_0 + \kappa_1} &                       &                               & \\
    \frac{\kappa_1}{\kappa_1 + \kappa_2} & 0                     & \frac{\kappa_2}{\kappa_1 + \kappa_2} &                               & \\
                          & \frac{\kappa_2}{\kappa_2 + \kappa_3} & 0                     & \ddots                        & \\
                          &                       & \ddots                & \ddots                        & \frac{\kappa_{d-1}}{\kappa_{d-2} + \kappa_{d-1}} \\
                          &                       &                       & \frac{\kappa_{d-1}}{\kappa_{d-1} + \kappa_d} & 0 \\
  \end{array}\right]\,,$$
where $\kappa_0 = \kappa_d =0$. Hence $A = \mathbb{I} - B = D C$ with diagonal
  $$D = \left[\begin{array}{ccccc}
    \frac{1}{\kappa_0 + \kappa_1} &                     &                     &        & \\
                        & \frac{1}{\kappa_1 + \kappa_2} &                     &        & \\
                        &                     & \frac{1}{\kappa_2 + \kappa_3} &        & \\
                        &                     &                     & \ddots & \\
                        &                     &                     &        & \frac{1}{\kappa_{d-1} + \kappa_d} \\
  \end{array}\right]$$
  and tridiagonal
  $$C = \left[\begin{array}{ccccc}
    \kappa_0 + \kappa_1 & -\kappa_1      &           &          & \\
    -\kappa_1      & \kappa_1 + \kappa_2 & -\kappa_2      &          & \\
              & -\kappa_2      & \kappa_2 + \kappa_3 & \ddots   & \\
              &           & \ddots    & \ddots   & -\kappa_{d-1} \\
              &           &           & -\kappa_{d-1} & \kappa_{d-1} + \kappa_d \\
  \end{array}\right]\,.$$
  The determinants of $D$ and $C$ can be expressed in closed form as
  $$\det D = \prod_{i=1}^d \frac{1}{\kappa_{i-1} + \kappa_i}, \qquad \det C = \sum_{i=0}^d \prod_{j=0 \atop j\ne i}^d \kappa_j.$$
  Hence, since $\kappa_i > 0$ for $i = 1, \dots, d-1$, $\det A = (\det C) (\det D) > 0$.
  Also, we clearly have $\det \mathrm{diag}(A) > 0$.
  Hence, condition (\ref{eq:LocatelliSchiavoni}) holds for $I = \C{I}$.
  A similar calculation (and exploiting the block structure of the principal submatrices) shows that 
  condition (\ref{eq:LocatelliSchiavoni}) holds for all $I \subseteq \C{I}$.
\end{example}
Remarkably, we can thus apply Corollary~\ref{corr:RealizingLinearSCMAsCDM} to the damped harmonic oscillator SCM to obtain a realization
of this causal equilibrium model as a \emph{first-order} linear \CDM\ (remember that the original \CDM\ is a second-order linear \CDM).

\subsection{Causal interpretation of the graph of the equilibrated \CDM}
\label{sec:CausalInterpretationGraphEquilibratedCDM}

While the graph of an acyclic SCM  has a straightforward causal interpretation, this need not be the case for general SCMs with cycles \citep{BPSM19}.\footnote{The straightforward causal interpretation of acyclic SCMs actually extends to a much more general class of possibly cyclic SCMs, referred to by \citet{BPSM19} as \emph{simple} SCMs.}
While an acyclic SCM induces a unique ``observational'' distribution, cyclic SCMs may induce none, one or several different observational distributions \citep{Hal98,BPSM19}.
Similarly, after performing a perfect intervention on some of the variables, a cyclic SCM may induce none, one or several different corresponding interventional distributions.
In general, one has to be careful in how to causally interpret the graph of an SCM if cycles are present; in particular, this caveat holds for SCMs that are obtained as the equilibration of an \CDM.
First, not all directed edges and directed paths in the graph can easily be identified from differences in interventional distributions in case cycles are present~\citep{BPSM19}.
Second, if cycles are present, ``nonancestral'' effects may exist~\citep{Nea00,BPSM19}, that is, an intervention on a variable may change the distribution of some of its nondescendants in the graph. In this subsection, we show how these subtleties and counterintuitive nonancestral effects in cyclic equilibrated \CDM s can be explained in terms of properties of the underlying \CDM .

In general, the presence or absence of a directed edge or path in the graph of an SCM $\C{M}$ cannot always be identified from the observational and interventional distributions. In the cyclic setting, the following sufficient condition can be used to identify such directed edges or paths between nodes $i$ and $j$ (see Proposition~7.1 in \citep{BPSM19} for the exact formulation).
\begin{itemize}
  \item A \emph{direct causal effect} of $i$ on $j$ can be identified, that is, there exists a $i\gto j \in \C{G}(\C{M})$, if (i) the structural equation of $j$ can be solved a.s.\ uniquely for $X_j$ in terms of the other variables that appear in the equation, and (ii) there exist values $\B{K}_I\in\BC{X}_I$ and $K_i \ne \tilde{K}_i \in \C{X}_i$, where $I=\C{I}\setminus\{i,j\}$, and a measurable set $\C{B}_j\subseteq\C{X}_j$ such that the following probabilities are uniquely defined and do not coincide:
$$
\Prb_{(\C{M}_{\intervene(I,\B{K}_I)})_{\intervene(i,K_i)}}(X_j\in\C{B}_j)
  \ne 
\Prb_{(\C{M}_{\intervene(I,\B{K}_I)})_{\intervene(i,\tilde{K}_i)}}(X_j\in\C{B}_j) \,;
$$
\item An \emph{indirect causal effect} of $i$ on $j$ can be identified, that is, there exists a directed path $i \gto\cdots\gto j$ in $\C{G}(\C{M})$, if (i) the structural equations of the ancestors of $j$ in $\C{G}(\C{M})_{\setminus i}$ (that is, the graph $\C{G}(\C{M})$ where we removed the node $i$ and its adjacent edges) can be solved a.s.\ uniquely for their associated variables in terms of the other variables that appear in these equations, and (ii) there exist values $K_i \ne \tilde{K}_i \in \C{X}_i$ and a measurable set $\C{B}_j\subseteq\C{X}_j$ such that the following probabilities are uniquely defined and do not coincide:
$$
\Prb_{\C{M}_{\intervene(i,K_i)}}(X_j\in\C{B}_j)
  \ne 
\Prb_{\C{M}_{\intervene(i,\tilde{K}_i)}}(X_j\in\C{B}_j) \,.
$$
\end{itemize}
In the following example, we illustrate how we can interpret the directed edges and paths of the equilibrated bathtub model that cannot be identified by this sufficient condition from an \CDM\ perspective.

  \begin{table}[bt]
    \hspace{0.3em}
    $\arraycolsep=1.0em
\begin{array}{cccc}
  \begin{tabular}[t]{c c c|c|c|c|c}
    & & \multicolumn{5}{c}{directed path to} \\
    & \multicolumn{1}{c V{2.5}}{} & $K$ & $Q_i$ & $P$ & $Q_o$ & $D$ \\ \Cline{2-7}{1pt}
    \multirow{5}{*}{\rotatebox[origin=c]{90}{from}} 
    & \multicolumn{1}{c V{2.5}}{$K$} & {\color{black} - } & {\color{black} $\times$} & $\checkmark$ & $\checkmark$ & $\checkmark$ \\ \cline{2-7}
    & \multicolumn{1}{c V{2.5}}{$Q_i$} & {\color{black} $\times$} & {\color{black} - } & $\checkmark$ & $\checkmark$ & $\checkmark$ \\ \cline{2-7}
    & \multicolumn{1}{c V{2.5}}{$P$} & {\color{black} $\times$} & {\color{black} $\times$} & {\color{black} - } & $\checkmark$ & $\checkmark$ \\ \cline{2-7}
    & \multicolumn{1}{c V{2.5}}{$Q_o$} & {\color{black} $\times$} & {\color{black} $\times$} & $\checkmark$ & {\color{black} - } & $\checkmark$ \\ \cline{2-7}
    & \multicolumn{1}{c V{2.5}}{$D$} & {\color{black} $\times$} & {\color{black} $\times$} & $\checkmark$ & $\checkmark$ & $\checkmark$ \\
  \end{tabular}
 & &
  \begin{tabular}[t]{c c c|c|c|c|c c}
    & & \multicolumn{5}{c}{directed edge to} & \multirow{7}{*}{} \rdelim\}{7}{-35mm}[\parbox{3cm-\tabcolsep-\widthof{$\Bigg]$}}{\centering
    according \\
    to $\C{G}(\C{M}_{\C{R}})$ \\
  (see Figure~\ref{fig:Bathtub} \\ top right)}] \\
    & \multicolumn{1}{c V{2.5}}{} & $K$ & $Q_i$ & $P$ & $Q_o$ & $D$ \\ \Cline{2-7}{1pt}
    \multirow{5}{*}{\rotatebox[origin=c]{90}{from}} 
    & \multicolumn{1}{c V{2.5}}{$K$} & {\color{black} - } & {\color{black} $\times$} & {\color{black} $\times$} & $\checkmark$ & {\color{black} $\times$ } \\ \cline{2-7}
    & \multicolumn{1}{c V{2.5}}{$Q_i$} & {\color{black} $\times$} & {\color{black} - } & {\color{black} $\times$} & {\color{black} $\times$} & $\checkmark$ \\ \cline{2-7}
    & \multicolumn{1}{c V{2.5}}{$P$} & {\color{black} $\times$} & {\color{black} $\times$} & {\color{black} -} & $\checkmark$ & {\color{black} $\times$ } \\ \cline{2-7}
    & \multicolumn{1}{c V{2.5}}{$Q_o$} & {\color{black} $\times$} & {\color{black} $\times$} & {\color{black} $\times$} & {\color{black} - } & $\checkmark$ \\ \cline{2-7}
    & \multicolumn{1}{c V{2.5}}{$D$} & {\color{black} $\times$} & {\color{black} $\times$} & $\checkmark$ & {\color{black} $\times$} & $\checkmark$ \\
  \end{tabular} \\
  \begin{tabular}[t]{c c c|c|c|c|c}
    & & \multicolumn{5}{c}{indirect effect} \\
    & \multicolumn{1}{c V{2.5}}{} & $K$ & $Q_i$ & $P$ & $Q_o$ & $D$ \\ \Cline{2-7}{1pt}
    \multirow{5}{*}{\rotatebox[origin=c]{90}{cause}} 
    & \multicolumn{1}{c V{2.5}}{$K$} & {\color{black} - } & {\color{gray} ? } & $\checkmark$ & {\color{black} ?} & $\checkmark$ \\ \cline{2-7}
    & \multicolumn{1}{c V{2.5}}{$Q_i$} & {\color{gray} ? } & {\color{black} - } & $\checkmark$ & $\checkmark$ & $\checkmark$ \\ \cline{2-7}
    & \multicolumn{1}{c V{2.5}}{$P$} & {\color{gray} ?? } & {\color{gray} ?? } & {\color{black} - } & {\color{black} ?? } & {\color{black} ?? } \\ \cline{2-7}
    & \multicolumn{1}{c V{2.5}}{$Q_o$} & {\color{gray} ?? } & {\color{gray} ?? } & {\color{black} ?? } & {\color{black} - } & {\color{black} ?? } \\ \cline{2-7}
    & \multicolumn{1}{c V{2.5}}{$D$} & {\color{gray} ? } & {\color{gray} ? } & $\checkmark$ & $\checkmark$ & {\color{black} - } \\
  \end{tabular}
 & &
  \begin{tabular}[t]{c c c|c|c|c|c c}
    & & \multicolumn{5}{c}{direct effect} & \multirow{7}{*}{} \rdelim\}{7}{-35mm}[\parbox{3cm-\tabcolsep-\widthof{$\Bigg]$}}{\centering
    Identifiable from \\
    the observational \\
    and interventional \\
    distributions with \\
    Prop.~7.1 in \citep{BPSM19}}] \\
    & \multicolumn{1}{c V{2.5}}{} & $K$ & $Q_i$ & $P$ & $Q_o$ & $D$ & \\ \Cline{2-7}{1pt}
    \multirow{5}{*}{\rotatebox[origin=c]{90}{cause}} 
    & \multicolumn{1}{c V{2.5}}{$K$} & {\color{black} - } & {\color{gray} ? } & {\color{gray} ? } & $\checkmark$ & {\color{gray} ?? } \\ \cline{2-7}
    & \multicolumn{1}{c V{2.5}}{$Q_i$} & {\color{gray} ? } & {\color{black} - } & {\color{gray} ? } & {\color{gray} ? } & {\color{black} ?? } \\ \cline{2-7}
    & \multicolumn{1}{c V{2.5}}{$P$} & {\color{gray} ? } & {\color{gray} ? } & {\color{black} -} & $\checkmark$ & {\color{gray} ?? } \\ \cline{2-7}
    & \multicolumn{1}{c V{2.5}}{$Q_o$} & {\color{gray} ? } & {\color{gray} ? } & {\color{gray} ? } & {\color{black} - } & {\color{black} ?? } \\ \cline{2-7}
    & \multicolumn{1}{c V{2.5}}{$D$} & {\color{gray} ? } & {\color{gray} ? } & $\checkmark$ & {\color{gray} ? } & {\color{black} - } \\
  \end{tabular}
\end{array}$
\vspace{1.0em}
\caption{The directed paths/edges (top tables) of the equilibrated bathtub model $\C{M}_{\C{R}}$ and the (in)direct causal effects that can be identified by Proposition~7.1 in \citep{BPSM19} (bottom tables) are denoted by a ``$\checkmark$''. Those that cannot be identified are denoted by the question marks ``?'' and ``??''. A single question mark ``?'' denotes that 
  condition (i) is satisfied, but not condition (ii), while a double question mark ``??'' denotes that condition (i) is not satisfied.}
\label{tab:BathtubInterventions}
\end{table}
\begin{example}[Bathtub model, continued]
\label{ex:BathtubCausalRelations}
Consider again the bathtub model $\C{R}$ of Example~\ref{ex:Bathtub}. We simulated some numerical solutions, with parameters as given in Example~\ref{ex:BathtubSimulation}, shown in Figure~\ref{fig:BathtubSimulation} (top left). 
  In Table~\ref{tab:BathtubInterventions} (bottom left) one can read off all the indirect causal effects that can be identified by comparing different interventional distributions from the equilibrated model $\C{M}_{\C{R}}$ with the help of Proposition~7.1 in \citep{BPSM19}. The indirect causal effects of $P$ and $Q_o$ cannot be identified by comparing interventional distributions, since the intervened equilibrated models $(\C{M}_{\C{R}})_{\intervene(P,K_P)}$ and $(\C{M}_{\C{R}})_{\intervene(Q_o,K_{Q_o})}$ do not have a solution (except for one special choice of $K_P$ respectively $K_{Q_o}$), and hence condition (i) is not satisfied. 
This was already illustrated for the perfect intervention $\intervene(Q_o,K_{Q_o})$ in Figure~\ref{fig:BathtubSimulation} (top center/right) of Example~\ref{ex:BathtubSimulation}. 

The direct causal effects that can be identified from $\C{M}_{\C{R}}$ are given in Table~\ref{tab:BathtubInterventions} (bottom right). The direct causes of $D$ cannot be identified due to the self-cycle at $D$, which means that condition (i) is not satisfied, that is, the structural equation of $D$ cannot be a.s.\ uniquely solved for $D$ in terms of the other variables.
Indeed, the depth $D$ will not equilibrate, but will increase indefinitely, if the rate of water into the bathtub is larger than the outflow rate, that is, $K_{Q_o}<K_{Q_i}$ (see Figure~\ref{fig:BathtubSimulation} bottom right). On the other hand, it will reach an equilibrium state only if the rate of water into and out of the bathtub are equal, that is, $K_{Q_o}=K_{Q_i}$. In this case, the depth $D$ will remain constant over all times, as illustrated in Figure~\ref{fig:BathtubSimulation} (bottom center). 

The directed path from $K$ to $Q_o$ in the graph of the equilibrated model $\C{M}_{\C{R}}$ cannot be straightforwardly identified as an indirect causal effect at equilibrium, because the equilibrium distribution of $Q_o$ does not change due to perfect interventions on $K$ (this corresponds to the single question mark in the Table~\ref{tab:BathtubInterventions}, bottom left), as explicit calculations reveal. 
However, at some finite time point one does observe changes in the distribution of $Q_o$ when performing perfect interventions on $K$ (Figure~\ref{fig:BathtubSimulation} (left)).
Together, this implies that this system is capable of perfect adaptation \citep{BM21}. 
Interestingly, the direct edge $K \gto Q_o$ in the graph of the equilibrated model $\C{M}_{\C{R}}$ can be identified by changes in the equilibrium distribution of $Q_o$ under perfect interventions on $K,Q_i,P,D$ (which then also implies that there is a directed path from $K$ to $Q_o$ in the graph of the equilibrated model).
\end{example}

In particular, this example illustrates that one can run into several problems when one attempts to identify directed edges and paths of the graph of the SCM from the differences in equilibrium distributions under interventions on the \CDM: 
\begin{itemize}
  \item if the intervened SCM has no solutions, then the descendants of the intervention targets cannot be easily identified;
  \item if the graph of the SCM has a self-cycle at some variable, then the parents of that variable cannot be easily identified;
  \item if the equilibrium distribution of some descendants of the intervention target variable remain insensitive to the intervention (for example, when the dynamical system exhibits perfect adaptation~\citep{BM21}), these descendants cannot be easily identified.
\end{itemize}

In Example~\ref{ex:BathtubCausalRelations}, the identified indirect causal relationships are a subset of the ancestral relationships. This can be seen from observing that each ``$\checkmark$'' in Table~\ref{tab:BathtubInterventions} (bottom left) has a corresponding ``$\checkmark$'' in Table~\ref{tab:BathtubInterventions} (top left). 
In other words, performing a perfect intervention on a variable can only change the distribution of its descendants in the graph.
In general, however, it can happen that an intervention on a nonancestor of a variable can change the distribution of that variable \citep{Nea00,BPSM19}. This counterintuitive behavior of ``nonancestral'' effects in an equilibrated \CDM\ can be explained by the dependence of the equilibrium states on the initial conditions in combination with the fact that not each initial condition corresponds to an equilibrating solution. The following example illustrates this.
\begin{example}[Selection bias leading to nonancestral effects in an equilibrated \CDM]
\label{ex:NealSimulation}
Consider the \CDM\ $\C{R}$ with \SDEfull s given by 
$$
  \left\{\begin{aligned}
    X_1 &= X_1 - X_1' + 2X_2 - X_3 \\
    X_2 &= X_2 - X_2' \\
    X_3 &= E \,,
  \end{aligned}\right. 
$$
  with order tuple $\B{n}=(1,1,0)$ and $E$ some constant in $\RN$. Denote $I = \{1,2\}$ and note that $\C{R}$ satisfies Assumption~\asref{ass:ExplicitlySolvable}{(I\subseteq\C{I})}. The equilibrated model $\C{M}_{\C{R}}$ is given by 
$$
  \left\{\begin{aligned}
    X_1^* &= X_1^* + 2X_2^* - X_3^* \\
    X_2^* &= X_2^* \\
    X_3^* &= E \,.
  \end{aligned}\right. 
$$
The graphs of $\C{R}$ and $\C{M}_{\C{R}}$ are depicted in Figure~\ref{fig:NealSimulationGraphs}.
\begin{figure}
\begin{center}
\adjustbox{scale=0.85}{%
\begin{tikzpicture}
  \begin{scope}
  \node at (-2.0,-0.7) {$\C{G}(\C{R})$:};
  \node[var] (X1) at (0,0) {$X_1$};
  \node[var] (X1p) at (0,-1.4) {$X'_1$};
  \node[var] (X2) at (2.8,0) {$X_2$};
  \node[var] (X2p) at (2.8,-1.4) {$X'_2$};
  \node[var] (X3) at (1.4,-1.4) {$X_3$};
  \draw[draw=black!60] (-0.45,0.45) rectangle (0.45,-1.85);
  \draw[draw=black!60] (2.35,0.45) rectangle (3.25,-1.85);
  \draw[draw=black!60] (0.95,-1.85) rectangle (1.85,-0.95);
  \draw[arr] (X2) -- (X1);
  \draw[arr] (X3) -- (X1);
  \draw[arr, dashed] (X1) to [out=60,in=120,looseness=5] (X1);
  \draw[arr, dashed] (X2) to [out=60,in=120,looseness=5] (X2);
  \draw[arr,dashed, bend left=15] (X1) to (X1p);
  \draw[arr,dashed, bend left=15] (X1p) to (X1);
  \draw[arr,dashed, bend left=15] (X2) to (X2p);
  \draw[arr,dashed, bend left=15] (X2p) to (X2);
  \node[text=black!60] at (0.0,-2.15) {$\widebar{X}^{(n_1)}_1$};
  \node[text=black!60] at (1.4,-2.15) {$\widebar{X}^{(n_3)}_3$};
  \node[text=black!60] at (2.8,-2.15) {$\widebar{X}^{(n_2)}_2$};
  \end{scope}
  \begin{scope}[xshift=7.5cm]
  \node at (-2.0,-0.7) {$\C{G}(\C{M}_{\C{R}})$:};
  \node[var] (X1) at (0,0) {$X_1$};
  \node[var] (X2) at (2.8,0) {$X_2$};
  \node[var] (X3) at (1.4,-1.4) {$X_3$};
  \draw[arr] (X2) -- (X1);
  \draw[arr] (X3) -- (X1);
  \draw[arr, dashed] (X1) to [out=60,in=120,looseness=5] (X1);
  \draw[arr, dashed] (X2) to [out=60,in=120,looseness=5] (X2);
\end{scope}
\end{tikzpicture}}
\end{center}
\caption{Graphs of the \CDM\ $\C{R}$ (left) and the corresponding equilibrated model $\C{M}_{\C{R}}$ (right) of Example~\ref{ex:NealSimulation}.}
\label{fig:NealSimulationGraphs}
\end{figure}
First observe that the induced equilibrium distribution of $X_2^*$ differs for two constant perfect interventions $\intervene(3,K_3)$ and $\intervene(3,\tilde{K}_3)$ with $K_3 \ne \tilde{K}_3$, since the equilibrium state has to satisfy $X_2^*=X_3^*/2$ a.s.. However, there is no directed path from the variable $X_3$ to the variable $X_2$ in the graph of the SCM $\C{M}_{\C{R}}$. This counterintuitive behavior 
can be explained by taking the initial conditions of the solutions of the \CDM\ into account, as we shall now explain. 

In Figure~\ref{fig:NealSimulation}, we plot the solutions of the \CDM\ $\C{R}$ for different partial initial conditions $(t_0,\B{X}^i_{I,[0]})$ at $t_0=0$ (for $i=a,b,\dots, g$) under two steady perfect interventions, namely $\intervene(3,K_3=1.0)$ and $\intervene(3,\tilde{K}_3=0.6)$. For illustration purposes, we consider here only non-random initial conditions, because we can then identify the initial conditions with ``individual'' solutions, as depicted in Figure~\ref{fig:NealSimulation} (note that Corollary~\ref{cor:ExistenceAndUniquenessLinearCase} applies). Observe that the set of partial initial conditions that correspond to equilibrating solutions differs for the two interventions. For the intervened model $\C{R}_{\intervene(3,K_3)}$, the only solution that equilibrates is the one with initial condition $(t_0, \B{X}_{I,[0]}^a)$ (denoted by the dark solid lines Figure~\ref{fig:NealSimulation} (top left)), whereas for the intervened model $\C{R}_{\intervene(3,\tilde{K}_3)}$ the only solution that equilibrates is the one with initial condition $(t_0, \B{X}_{I,[0]}^b)$ (denoted by the dark dotted lines in Figure~\ref{fig:NealSimulation} (top right)). 
This explains the counterintuitive behavior of nonancestral effects in the equilibrium SCM: 
The chosen value for $X_3$ affects which solutions will equilibrate, and thereby affects the equilibrium distribution of $X_2$.

Note that at any finite point in time, these ``nonancestral'' effects do not occur; indeed, Figure~\ref{fig:NealSimulation} shows that the distribution of $X_1$ differs for the two interventions at finite time, while that of $X_2$ remains unaffected.

\Stephan{
First observe that the induced equilibrium distribution of $X_2^*$ differs for perfect interventions $\intervene(3,K_3)$ and $\intervene(3,\tilde{K}_3)$ (if $K_3 \ne \tilde{K}_3$), since the equilibrium state has to satisfy $X_2^*=X_3^*/2$ a.s.. However, there is no directed path from the variable $X_3$ to the variable $X_2$ in the graph. The counterintuitive behavior that a perfect intervention on a nonancestor of a variable of this SCM can change the distribution of that variable can be explained by taking the solutions of the \CDM\ into account. 

Upon closer inspection, the different solutions of $X_2^*$ under different interventions on $X_3^*$ actually stem from different initial conditions at $t_0$. To see this, consider the intervened model $\C{R}_{\intervene(3,K_3)}$, where $K_3=1$. The only initial conditions of $(t_0,\B{X}_0)$ at $t_0=0$ that equilibrate are those with $X_{20}=1/2$. This is illustrated in Figure~\ref{fig:NealSimulation} (left) for some numerical solutions of the intervened model $\C{R}_{\intervene(3,K_3)}$ for random consistent initial conditions $(t_0,\B{X}_0)$, where the red dashed line corresponds to the intervened process $X_3=K_3$ and the solid lines correspond to the processes $X_1$ and $X_2$ where $X_{10}=1.5$ and $X_{20}=0.5$. Performing instead the steady perfect intervention $\intervene(3,\tilde{K}_3)$ with $\tilde{K}_3=0.6$ and if we consider the same initial conditions as those of the solid lines for the processes $X_1$ and $X_2$ (i.e., with $X_{10}=1.5$ and $X_{20}=0.5$), then we see a difference (in distribution) at some finite point $t$ in time for the process $X_1$, but not for the process $X_2$, which is illustrated by the dotted lines in Figure~\ref{fig:NealSimulation} (right). This illustrates that performing a steady perfect intervention on $X_3$ does not affect the nonancestor process $X_2$ with the same initial conditions for the nonintervened processes as the one before the intervention. However, after the intervention $\intervene(3,\tilde{K}_3)$, the solution with this initial condition (represented by the solid lines) is not equilibrating. The only initial conditions $(t_0,\B{X}_0)$ that are equilibrating are the ones with $X_{20}=0.3$ and are depicted in Figure~\ref{fig:NealSimulation} (right) by the dotted lines. Therefore, equilibrating the \CDM\ under different steady perfect interventions may yield different intervened equilibrated \CDM s that describe equilibrium states that stem from rather different initial conditions. \Joris{This is a lot to swallow, it is a complicated message that is not yet presented in a very clear way\dots}}
\begin{figure*}
\begin{center}
\includegraphics[width=1.05\textwidth,keepaspectratio]{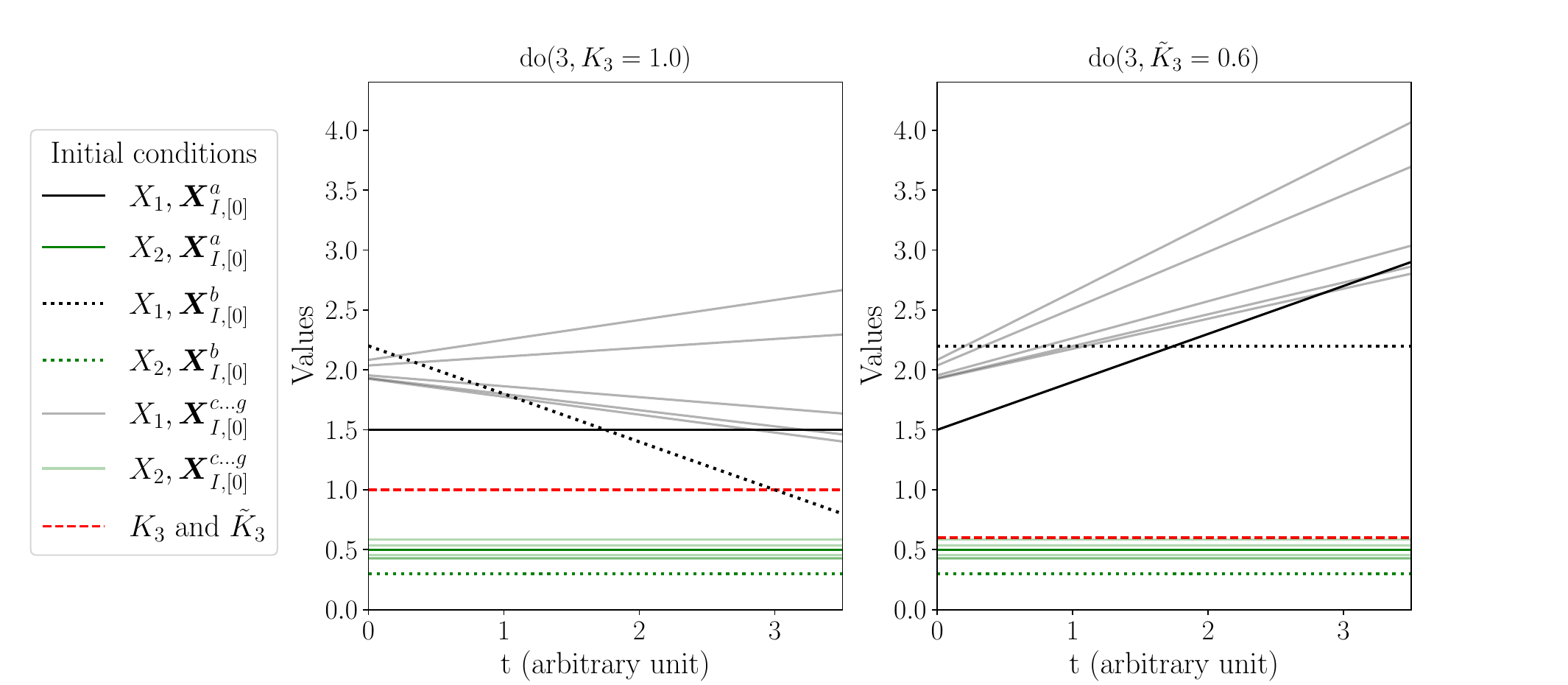}
$$
\begin{array}{c|cc}
  & (X^i_{I,[0]})_1 & (X^i_{I,[0]})_2 \\ \hline
  i = a & 1.5 & 0.5 \\ 
  i = b & 2.2 & 0.3 \\
  i = c\dots g & \text{random} & \text{random} 
\end{array}
$$
\end{center}
\vspace{-1.0\baselineskip}
\caption{Simulation of solutions of the \CDM\ of Example~\ref{ex:NealSimulation} under different steady perfect interventions on $X_3$ (top left and right). The simulations in the top left and right plots are performed under the same set of initial conditions, summarized in the bottom table, but under different interventions.}
\label{fig:NealSimulation}
\end{figure*}
\end{example}
This example shows that the nonancestral effects in an equilibrated SDCM can be explained by the 
dependence of the equilibrium states on the initial conditions, in combination with the fact that 
not each initial condition corresponds to an equilibrating solution.
Another way to think about this is as selection bias due to the assumption that the system has reached equilibrium. An intervention targeting a certain variable may change the set of equilibrating initial conditions of the system, and it can even change initial conditions for non-ancestors of the intervention target. By only considering these equilibrating initial conditions, this may appear as a causal effect of a variable on some of its non-ancestors at equilibrium. When seen from this perspective, these ``causal effects'' can be considered to be spurious as they do not appear on an ``individual level'', that is, for individual trajectories (at finite time $t$), but only appear on a ``population level'' when selecting on some later event (namely, the system being at equilibrium). 
One can indeed think of this as selection bias due to equilibration.  


\Stephan{Rephrase:\\

and indicates that 

f the indirect effect variable at equilibrium, do 

not coincides.

This example illustrates that all the directed paths and edges between a cause and an effect can be determined by observing a difference in the distribution of the effect at a certain time or at equilibrium after certain interventions on the cause.

This also sheds more light on

In \citep{BBM19} they consider very stationary systems, where under each intervention the system together with the initail conditions equilibrates to all the solutions of the SCM.

In contrast, to the work by \citep{BBM19} we do not consider the initial conditions to be part of the model. Given an initial condition of some order $\B{m}$, each equilibrating solution with this initial condition can be mapped to a solution of the SCM to which this solution of the \CDM\ equilibrates. In general, this mapping is neither injective nor surjective. An example of an \CDM\ where this mapping is not injective, is the \CDM\ $\C{R}$ with the SDE $X=X$ and the equilibrating initial condition $(0,X_0)$ with $X_0=0$. This \CDM\ has, for example, many solutions $X$ with this initial condition that equilibrate to $X^*=0$ (for example, $X_t=\pm \sin(t) e^{-t}$). 
An example of an \CDM\ where this mapping is not surjective is the intervened bathtub model

basic enzyme reaction model where all the equilibrating solutions $X_E$ equilibrate to the same random variable $X_E^*$ (see Table~1 in~\citet{BBM19}), while $X_E^*$ can be any random variable in the completely equilibrated SCM. In the recent work by \citep{BBM19} they call an SCM representation of the equilibrium states of a well behaved class of first-order dynamical systems \emph{complete}, if this mapping is a bijective mapping. A generalization of SCMs, known as Causal Constraints Models, has been proposed as an alternative \citep{BBM19}, and this class is rich enough to completely model the causal semantics (under perfect interventions) of the equilibrium states of such a dynamical systems given some initial condition. Here, we take a more agnostic approach, where the consistently equilibrating initial conditions are not part of the definition of the model. This allows us to causally model all the equilibrium states of any convergent \CDM\ in an SCM, which has many appealing features, such as a graphical representation that offers Markov properties and causal insight, which is still lacking for CCMs, as of today.
}


\section{Discussion}
\label{sec:Discussion}

Dynamical models consisting of (ordinary or random) differential equations are widely applied in science and engineering to model the dynamics of systems that are composed of several components. These differential equations by themselves do not have a clearcut causal interpretation. Although they may implicitly explain a particular phenomenon in terms of its causes, the causal semantics of the constituent components are generally not explicitly defined without additional assumptions. 

In this work, we introduced \CDMfull s that formally encode causal semantics of stochastic processes by means of a structured set of random differential equations. 
\CDM s can be seen as stochastic-process versions of structural causal models, where the random variables are replaced by stochastic processes and their derivatives. By viewing the (higher-order) derivatives $X_i^{(k_i)}$ to be aspects of the process $X_i$ we arrive at a natural causal interpretation, where it is not necessary to even consider questions like ``does position cause velocity, or does velocity cause position, or both?''.

For steady \CDM s (for which the explicit time-dependence of the dynamics vanishes as $t\to\infty$) we introduced an equilibration operation that equilibrates the dynamic causal mechanism of each component separately. This led to the important result that intervention and equilibration commute, thus connecting the causal semantics at equilibrium with the causal semantics of the dynamics. 
It generalizes the analogous result of \citet{MJS13} in three directions: (i) we replaced the deterministic setting with a more general stochastic setting, which allows us to address both cycles and confounders, (ii) we allowed the order of the \SDEfull s to be arbitrary, including zeroth-order, rather than only allowing for first-order differential equations, and (iii) we have dropped the strong assumption that the dynamical model needs to have a single globally attractive equilibrium state. This allows us to study the causal semantics of the equilibrium states of a plethora of dynamical systems subject to time-varying random disturbances encountered in science and engineering within the framework of structural causal models.

Our commutation result may appear to be at odds with the possible ``violation of the equilibration-manipulation commutability property'' pointed out by \citet{Das05}. Under our notion of equilibration---contrary to that of Dash---each \SDEfull\ of the \CDM\ becomes a structural equation of the SCM. This one-to-one correspondence between the equations leads to the preservation of the causal semantics under equilibration. We can reinterpret the phenomenon that Dash observed as the fact that the equilibrium distributions of certain dynamical systems (for example ones that exhibit perfect adaptation) are not faithful to the graph of the equilibrated SCM \citep{BM21}, in the sense that they can contain conditional independencies not explained by this graph. For dynamical systems exhibiting perfect adaptation, these faithfulness violations are due to the structure of the dynamics, rather than ``accidental'' parameter cancellations. This has serious repercussions for attempts at inferring the causal structure from (conditional independencies in) equilibrium data \citep{BM21}. Thus, in a different way we arrive at the same conclusion as Dash obtained.

In comparison with the causal constraints models of \citet{BBM19}, our modeling framework is more ``agnostic'' as we decided not to incorporate the initial conditions into the model.\footnote{This is analogous to the difference between an ODE and an initial-value problem.} This allowed us to causally model \emph{all} the equilibrium states of a steady \CDM\ with a single SCM. However, that single SCM may not provide a \emph{complete} description of the causal semantics at equilibrium \citep{BBM19}. This is indeed a modeling tradeoff: the simpler structure of SCMs compared to that of causal constraints models can come at the cost of a less complete description of the equilibrium behavior of certain dynamical systems. On the other hand, the connection between the structure of the SCM and that of the underlying \CDM\ is straightforward, whereas it is not well understood at present how one can easily derive a concise yet complete representation of an equilibrated \CDM\ (and a corresponding initial condition) as a causal constraints model.

However, allowing for multiple (or no) solutions also comes at a cost: the causal interpretation of the SCM is more subtle than that of acyclic (or more generally, simple) SCMs, and in particular, does not straightforwardly relate to properties of its graph.
We illustrated for the bathtub model how one can causally interpret the directed edges and paths of the graph of the SCM that models the equilibrium states of the underlying \CDM .
We saw that one may run into several problems when attempting to identify aspects of the SCM graph from comparing differences in equilibrium distributions after intervening on some of the variables:
\begin{itemize}
  \item if the intervened SCM has no solution (which may happen if the intervened \CDM\ does not converge to a finite equilibrium state, but instead diverges to infinity, or reaches a periodic limit cycle, for example), descendants of the intervention targets cannot be easily identified;
  \item if the SCM graph has a self-cycle at some variable (which may happen if the causal mechanism for that variable does not equilibrate for certain values of its parents), then the parents of that variable cannot be easily identified;
  \item if the equilibrium distributions of some descendants of the intervention target variable remain insensitive to the intervention (which may happen in dynamical systems exhibiting perfect adaptation), these descendants cannot be easily identified.
\end{itemize}

Even worse, the equilibrium SCM may entail distribution changes under interventions that appear to be of a causal nature, while no corresponding causal relations are present in the dynamics (and therefore, no corresponding ancestral relations are present in the SCM graph), as we pointed out in Example~\ref{ex:NealSimulation}. 
These counterintuitive ``nonancestral causal effects'' can be understood as arising from the implicit selection bias due to conditioning on the system having reached an equilibrium state. Indeed, the solutions of the equilibrium SCM correspond to those solutions of the \CDM\ that have equilibrated, while the non-equilibrating solutions of the \CDM\ are ignored. In other words, the SCM provides the ``population-level'' causal semantics of the population of equilibrating \CDM\ solutions (at $t = \infty$), which can deviate from the ``individual-level'' causal semantics of (possibly non-equilibrating) \CDM\ solutions (at finite $t$).
The phenomenon that population-level causality may differ from individual-level causality due to post-intervention selection bias is well-known in other contexts. For example, a car mechanic who only observes cars that don't start may conclude that replacing the battery causes start engines to fail. While this appears as a genuine causal effect on the population level, it would be foolish to conclude that this causal effect also pertains to individual cars. Intuitively, one might prefer to interpret such phenomena as not representing ``truely causal'' relations. On the other hand, if one is only interested in the effects of interventions on a population level, there seems to be no harm in considering these distribution changes as causal. Thus, as long as one is explicit whether one refers to population-level or individual-level causality, both notions of causality can meaningfully co-exist. The important take-away, from our point of view, is that focussing on equilibrated systems may lead to selection bias.

As a side note, Example~\ref{ex:NealSimulation} also shows that SCMs may not fully capture such population-level causal relations graphically. We note that the recently proposed framework of \citet{BlomVanDiepenMooij_JMLR_21} is better suited in general to read off such population-level causal effects graphically from the structure of the equilibrium equations, under certain ``local'' solvability assumptions on these equations (rather than having to study global solutions of intervened equilibrium equations, as we did here).

Apart from these subtleties regarding their causal semantics, SCMs with cycles bring about several other challenges in general. 
For example, they generally do not have a Markov property, and the class of cyclic SCMs is not closed under marginalization. 
The subclass consisting of simple SCMs \citep{BPSM19} allows for cycles, but simple SCMs share many of the convenient properties of acyclic SCMs. 
Hence, these convenient properties are directly applicable to the equilibrium states of those steady \CDM s that equilibrate to a simple SCM. 
This enables one to study the equilibrium states of those \CDM s by statistical tools and discovery methods available for simple SCMs. 
For example, one can apply adjustment criteria and Pearl's do-calculus~\citep{FM19}. 
Several causal discovery algorithms, originally designed for acyclic SCMs, like Local Causal Discovery (LCD) \citep{Coo97}, Y-structures \citep{Man06}, and the Fast Causal Inference (FCI) algorithm~\citep{SMR99,Zha08,MC20}, are directly applicable to simple SCMs as well \citep{MMC20}. 
Furthermore, the Joint Causal Inference (JCI) framework can be applied to combine data from different contexts (for example, observational and interventional) for causal discovery and inference purposes \citep{MMC20}.

Given that steady \CDM s for which all solutions equilibrate give rise to SCMs at equilibrium, the inverse problem becomes interesting as well: given an SCM, can we find an \CDM\ (with non-trivial dynamics) that equilibrates to this SCM and for which all solutions equilibrate? This question was answered affirmatively for a certain class of linear simple SCMs with additional constraints on the parameters by leveraging existing results from linear systems theory. We speculate that this result can be further generalized to allow for non-linearity. Perhaps surprisingly, this result allows to start from a second-order \CDM\ modeling a system of damped coupled harmonic oscillators, equilibrate it to obtain an SCM, and from that then construct a \emph{first-order} \CDM\ with the same equilibrium SCM that describes all equilibrium states under any constant stochastic perfect intervention. This shows that the order of the \SDEfull s is not necessarily constrained by the equilibrium SCM. Thus, the properties of the system at equilibrium may contain not enough information to identify the order of the dynamical equations.

We hope that the framework of \CDM s provides a natural starting point for modeling the causal mechanisms that underlie the dynamics of various systems, which could, in principle, be inferred from observations and experiments~\citep[see, for example,][]{BGMB17,PBP19,LLWS+20}. We believe that most of this work can also easily be adapted to discrete time by replacing the differential equations by difference equations. Future work might consist of (i) investigating the notion of local independence in \CDM s, (ii) studying how \CDM\ graphs can be interpreted causally, in particular if self-cycles or zeroth order equations are present, (iii) developing structure and parameter learning algorithms for \CDM s, and (iv) investigating possible extensions to stochastic dynamics by means of stochastic differential equations.

\section*{Acknowledgements}

We are grateful to Fabian Wirth for pointing us to the work of \citet{Fis58},
and to Patrick Forr\'e for stimulating discussions.
Stephan Bongers was supported by NWO, the Netherlands
Organization for Scientific Research (VIDI grant 639.072.410). Tineke Blom and
Joris Mooij were
supported by the European Research Council (ERC) under the European Union’s
Horizon 2020 research and innovation programme (grant agreement 639466).

\newpage
\appendix

\part*{Appendix} 

\section{Proofs}
\label{app:Proofs}


\begin{proof}[Proof of Lemma~\ref{lem:ExistenceAndUniquenessSolutions}]
Let $\B{X}_P:T \times \Omega \to \BC{X}_P$ be a $C^{\B{n}_P}$-stochastic process.
For every $i\in I$ we can write the random differential equations 
$$
X^{(n_i)}_i =
  g_i(\widebar{\B{X}}^{(\B{n}_I-1)}_I,\B{X}_J,\widebar{\B{X}}_P^{(\B{n}_P)},\B{E}_P)
$$
as a system of first-order random differential equations
$$
\frac{d}{dt} \widebar{X}_i^{(n_i-1)} = \widetilde{\B{g}}_i(\widebar{X}_i^{(n_i-1)}, \widebar{\B{X}}^{(\B{n}_{I \setminus
  i}-1)}_{I \setminus i},\B{X}_J,\widebar{\B{X}}_P^{(\B{n}_P)},\B{E}_P) \,,
$$
  where $\widetilde{\B{g}}_i:\C{X}^{n_i}_i\times\BC{X}^{\B{n}_{I \setminus i}}_{I \setminus i}\times\BC{X}_J\times\BC{X}_P^{\B{n}_P+1}\times\BC{E}_P \to\C{X}_i^{n_i}$ is the mapping defined by
$$
  \widetilde{\B{g}}_i(\widebar{x}_i^{(n_i -1)},\widebar{\B{x}}^{(\B{n}_{I \setminus i}-1)}_{I \setminus i},\B{x}_J,\widebar{\B{x}}_P^{(\B{n}_P)},\B{e}_P):=(x_i^{(1)},\dots,x_i^{(n_i-1)},g_i(\widebar{\B{x}}^{(\B{n}_I-1)}_I,
\B{x}_J,\widebar{\B{x}}_P^{(\B{n}_P)},\B{e}_P)).
$$
Note that $\widebar{X}_i^{(n_i-1)}=(X_i,X_i^{(1)},\dots, X_i^{(n_i-1)})$ and $\frac{d}{dt} \widebar{X}_i^{(n_i-1)}=(X_i^{(1)},\dots, X_i^{(n_i-1)}, X_i^{(n_i)})$.

Substituting the functions $\B{g}_J$ yields the following first-order RDE:
$$
\frac{d}{dt} \widebar{X}_i^{(n_i-1)} = \widetilde{\B{g}}_i\big (\widebar{X}_i^{(n_i-1)}, \widebar{\B{X}}^{(\B{n}_{I \setminus
i}-1)}_{I \setminus
i},\B{g}_J(\widebar{X}_i^{(n_i-1)}, \widebar{\B{X}}^{(\B{n}_{I \setminus
  i}-1)}_{I\setminus i},\widebar{\B{X}}_P^{(\B{n}_P)},\B{E}_P),\widebar{\B{X}}_P^{(\B{n}_P)}, \B{E}_P\big) \,. 
$$
  Let $\widetilde{\B{h}}_i(\widebar{\B{x}}^{(\B{n}_I-1)}_I,\widebar{\B{x}}_P^{(\B{n}_P)},\B{e}_P):=\widetilde{\B{g}}_i(\widebar{\B{x}}^{(\B{n}_I-1)}_I,\B{g}_J(\widebar{\B{x}}^{(\B{n}_I-1)}_I,\widebar{\B{x}}_P^{(\B{n}_P)},\B{e}_P),\widebar{\B{x}}_P^{(\B{n}_P)},\B{e}_P)$. 
We can then write the \SDE s as:
\begin{equation}\label{eq:SDE_as_RDE}
  \frac{d}{dt} \widebar{\B{X}}_I^{(\B{n}_I-1)} = \widetilde{\B{h}}_I\big (\widebar{\B{X}}_I^{(\B{n}_I-1)}, \widebar{\B{X}}_P^{(\B{n}_P)}, \B{E}_P\big) \,. 
\end{equation}
  The assumed continuity of $g_i$ and $\B{g}_J$, the continuity of the exogenous process $\B{E}_P$ and the assumption that $\B{X}_P$ is 
a $C^{\B{n}_P}$-stochastic process together imply that for almost all $\omega\in\Omega$ the function
  $(t,\widebar{\B{x}}^{(\B{n}_I-1)}_I) \mapsto \widetilde{\B{h}}_i(\widebar{\B{x}}^{(\B{n}_I-1)}_I,\widebar{\B{X}}_P^{(\B{n}_P)}(t,\omega),\B{E}_P(t,\omega))$ is continuous on $T\times\BC{X}^{\B{n}_I}_I$. Moreover, for each
$\widebar{\B{x}}^{(\B{n}_I-1)}_I\in\BC{X}^{\B{n}_I}_I$ the function 
  $(\widebar{\B{x}}_P^{(\B{n}_P)},\B{e}_P)\mapsto \widetilde{\B{h}}_i(\widebar{\B{x}}^{(\B{n}_I-1)}_I,\widebar{\B{x}}_P^{(\B{n}_P)},\B{e}_P)$ is continuous 
and in particular measurable.
Hence, for all
$(t,\widebar{\B{x}}^{(\B{n}_I-1)}_I)\in T\times\BC{X}^{\B{n}_I}_I$
the function 
$\omega \mapsto 
  \widetilde{\B{h}}_i(\widebar{\B{x}}^{(\B{n}_I-1)}_I,\widebar{\B{X}}_P^{(\B{n}_P)}(t,\omega),\B{E}_P(t,\omega))$ is $\C{F}$-measurable. 

Under the assumed condition, the following inequality holds for all 
$\widebar{\B{x}}^{(\B{n}_I-1)}_I, \widebar{\B{y}}^{(\B{n}_I-1)}_I \in \BC{X}^{\B{n}_I}_I$, for all $\widebar{\B{x}}_{P}^{(\B{n}_P)} \in \BC{X}_P^{\B{n}_P+1}$ and for all $\B{e}_P \in \BC{E}_P$:
\begin{gather*}
  \sum_{i\in I} \big\|
  \widetilde{\B{h}}_i(\widebar{\B{x}}^{(\B{n}_I-1)}_I,\widebar{\B{x}}_P^{(\B{n}_P)},\B{e}_P)
  -
  \widetilde{\B{h}}_i(\widebar{\B{y}}^{(\B{n}_I-1)}_I,\widebar{\B{x}}_P^{(\B{n}_P)},\B{e}_P)
  \big\|^2 \\ 
  = \sum_{i\in I} \big\|
  \widetilde{\B{g}}_i\big(\widebar{\B{x}}^{(\B{n}_I-1)}_I,\B{g}_J(\widebar{\B{x}}^{(\B{n}_I-1)}_I,\widebar{\B{x}}_P^{(\B{n}_P)},\B{e}_P),\widebar{\B{x}}_P^{(\B{n}_P)},\B{e}_P\big)
  -
  \widetilde{\B{g}}_i\big(\widebar{\B{y}}^{(\B{n}_I-1)}_I,\B{g}_J(\widebar{\B{y}}^{(\B{n}_I-1)}_I,\widebar{\B{x}}_P^{(\B{n}_P)},\B{e}_P),\widebar{\B{x}}_P^{(\B{n}_P)},\B{e}_P\big)
  \big\|^2 \\ 
  = \sum_{i\in I} \Big[ \| x_i^{(1)} - y_i^{(1)} \|^2 + \dots + \| x_i^{(n_i-1)} -
  y_i^{(n_i-1)} \|^2 +  \\
\big\|
  g_i\big(\widebar{\B{x}}^{(\B{n}_I-1)}_I,\B{g}_J(\widebar{\B{x}}^{(\B{n}_I-1)}_I,\widebar{\B{x}}_P^{(\B{n}_P)},\B{e}_P),\widebar{\B{x}}_P^{(\B{n}_P)},\B{e}_P\big)
  -
  g_i\big(\widebar{\B{y}}^{(\B{n}_I-1)}_I,\B{g}_J(\widebar{\B{y}}^{(\B{n}_I-1)}_I,\widebar{\B{x}}_P^{(\B{n}_P)},\B{e}_P),\widebar{\B{x}}_P^{(\B{n}_P)},\B{e}_P\big)
\big\|^2 \Big] \\
\leq \sum_{i\in I} \big[ \| x_i^{(1)} - y_i^{(1)} \|^2 + \dots + \| x_i^{(n_i-1)} -
y_i^{(n_i-1)} \|^2 + \kappa^2 \| x_i - y_i \|^2 \big]  \\
\leq (1+\kappa^2) \| \widebar{\B{x}}^{(\B{n}_I-1)}_I -
  \widebar{\B{y}}^{(\B{n}_I-1)}_I \|^2 \,.
\end{gather*}
Hence the conditions of Theorem~1.2 in \citet{Bun72} (or Theorem~3.2 in \citet{NR13}) are satisfied, which proves that there exists an a.s.\ unique solution $\B{X}_I$ of the system (\ref{eq:SDE_as_RDE}) of first-order RDEs for any partial initial condition $(t_0,\widebar{\B{X}}_{I,[0]}^{(\B{n}_I-1)})$. Note that the solution $\B{X}_I$ is a $C^{\B{n}_I}$-stochastic process. Extend this to a solution $\B{X}_{\C{O}}$ on $\C{O}$ by setting $\B{X}_J = \B{g}_J(\widebar{\B{X}}^{(\B{n}_{I}-1)}_{I},\widebar{\B{X}}_P^{(\B{n}_{P})},\B{E}_P)$. The result satisfies the smoothness requirement; indeed, from the assumptions it follows for each $j \in J$ that $X_j$ is a $C^{n_j}$-stochastic process.
\end{proof}

\begin{proof}[Proof of Proposition~\ref{prop:StructuralExplicitSolvabilityIntervention}]
  Let $g_i:\C{X}_i^{n_i}\times\BC{X}_{\C{O}\setminus i}\times\BC{X}_P^{\B{n}_P+1}\times\BC{E}_P\to\C{X}_i$ and $g_j:\BC{X}_I\times\BC{X}_P^{\B{n}_P+1}\times\BC{E}_P\to\C{X}_j$ for $i\in I$ and $j\in J$ be continuous mappings that make $\C{R}$ satisfy Assumption~\asref{ass:StructExplicitlySolvable}{(I\subseteq\C{O})}. Consider the stochastic perfect intervention $\intervene(L,\B{K}_L)$ with $L\subseteq\C{O}$. Then, the mappings $h_i:\C{X}_i^{n_i}\times\BC{X}_{\C{O}\setminus i}\times\BC{X}_P^{\B{n}_P+1}\times(\BC{X}_L\times\BC{E}_P)\to\C{X}_i$ for $i \in I \setminus L$ defined by 
$$
  h_i(\widebar{x}_i^{(n_i-1)},\B{x}_{\C{O}\setminus i},\widebar{\B{x}}_P^{(\B{n}_P)},(\tilde{\B{e}}_L,\B{e}_P)) := g_i(\widebar{x}_i^{(n_i-1)},\B{x}_{\C{O}\setminus i},\widebar{\B{x}}_P^{(\B{n}_P)},\B{e}_P)
$$
  and the mappings $h_j:\BC{X}_{I\setminus L}\times\BC{X}_P^{\B{n}_P+1}\times(\BC{X}_L\times\BC{E}_P)\to\C{X}_j$ for $j \in \C{O} \setminus (I \setminus L)$ defined by
$$
  h_j(\B{x}_{I\setminus L},\widebar{\B{x}}_P^{(\B{n}_P)},(\tilde{\B{e}}_L,\B{e}_P)) := \begin{cases} g_j((\B{x}_{I\setminus L}, \tilde{\B{e}}_L),\widebar{\B{x}}_P^{(\B{n}_P)},\B{e}_P) &\text{if }j\notin L \\
    \tilde{e}_j &\text{if $j \in L$}
\end{cases}
$$
  make $\C{R}_{\intervene(L,\B{K}_L)}$ satisfy Assumption~\asref{ass:StructExplicitlySolvable}{(I\setminus L\subseteq\C{O})}.
\end{proof}

\begin{proof}[Proof of Proposition~\ref{prop:LinearCDMExplicitlySolvable}]
  If the causal mechanism $\B{f}_{\C{O}}$ is defined as in the proposition, then the mappings $\B{g}_I:\BC{X}_I^{\B{n}_I}\times\BC{X}_J\times\BC{X}_P^{\B{n}_P+1}\times\BC{E}_P\to\BC{X}_I$ and $\B{g}_J:\BC{X}_I^{\B{n}_I}\times\BC{X}_P^{\B{n}_P+1}\times\BC{E}_P\to\BC{X}_J$ are given by 
\begin{align*}
  \B{g}_I(\widebar{\B{x}}^{(\B{n}_I-1)}_I,\B{x}_J,\widebar{\B{x}}_P^{(\B{n}_P)},\B{e}_P) & = -B_{I I^{(\B{n}_I)}}^{-1} (B_{I \widebar{I}^{(\B{n}_I-1)}}\widebar{\B{x}}^{(\B{n}_I-1)}_I - \B{x}_I + 
  B_{I J}\B{x}_J + B_{I \widebar{P}^{(\B{n}_P)}}\widebar{\B{x}}_P^{(\B{n}_P)} + \Gamma_{I P}\B{e}_P) \\
  \B{g}_J(\widebar{\B{x}}^{(\B{n}_I-1)}_I,\widebar{\B{x}}_P^{(\B{n}_P)},\B{e}_P) &= -B_{J J}^{-1} (B_{J \widebar{I}^{(\B{n}_I-1)}}\widebar{\B{x}}^{(\B{n}_I-1)}_I + B_{J \widebar{P}^{(\B{n}_P)}}\widebar{\B{x}}_P^{(\B{n}_P)} + \Gamma_{J P}\B{e}_P) \,.
\end{align*}
The converse is shown by taking for $B_{I I^{(\B{n}_I)}}$ and $B_{J J}$ the identity matrices.
\end{proof}


\begin{proof}[Proof of Corollary~\ref{cor:ExistenceAndUniquenessLinearCase}]
  For a linear \CDM\ $\C{R}$ that satisfies Assumption~\asref{ass:ExplicitlySolvable}{(I\subseteq\C{O})} there always exists a $\kappa\in\RN$ such that the uniformly-Lipschitz
condition of Lemma~\ref{lem:ExistenceAndUniquenessSolutions}
holds.
\end{proof}


\begin{proof}[Proof of Corollary~\ref{cor:ExistenceAndUniquenessLinearCaseStructural}]
This follows directly from Corollary~\ref{cor:ExistenceAndUniquenessLinearCase} and Proposition~\ref{prop:StructuralExplicitSolvabilityIntervention}. 
\end{proof}

\begin{proof}[Proof of Theorem~\ref{thm:Markov_property_SDCM}]
Let $\C{G}_{[0]}^+ := \C{G}_{[0]}^+(\C{R})$ denote the augmented collapsed graph of $\C{R}$.
We can construct the a.s.\ unique global solution $\B{X}$ of $\C{R}$ by recursively substituting the solutions into each other along the topological ordering of the directed acyclic graph formed by the strongly connected components $S\subseteq\C{I}$ of $\C{G}^+_{[0]}$.

We construct an SCM that has $\C{G}^+_{[0]}$ as its graph.
Consider the SCM with endogenous variables $X_i$ taking values in $\C{C}^{n_i}(T,\C{X}_i)$ for $i \in \C{I}$, 
exogenous variables $\widebar{X}_{[0],i}^{(n_i-1)}$ taking values in $\C{X}_i^{n_i}$ for $i \in \C{I}_{[0]}$,
as well as exogenous variables $E_j$ taking values in $\C{C}^0(T,\C{E}_j)$ for $j \in \C{J}$.
The structural equations of this SCM are taken to be of the following form.
Let $S\subseteq\C{I}$ be a strongly connected component of
  $\C{G}^+_{[0]}$ and write $P := \pa_{\C{G}^+_{[0]}}(S)\setminus S$. Observe that from Assumption~\asref{ass:ExplicitlySolvable}{(I_S\subseteq S)} and $\C{R}$ having a tight order tuple it follows that $\B{n}_{J_S} = 0$ for $J_S = S\setminus I_S$. 
The structural equations for $j \in J_S$ are taken to be of the form:
  \begin{equation}\label{eq:SCM_from_SDCM_AE}
    X_{j} = g_{j}(\widebar{\B{X}}_{I_S}^{(\B{n}_{I_S}-1)},\widebar{\B{X}}_{P}^{(\B{n}_P)},\B{E}_P).
  \end{equation}
For $i \in I_S$, we integrate the equation
  \begin{equation*}
    X_i^{(n_i)} = g_i(\widebar{\B{X}}_{I_S}^{(\B{n}_{I_S}-1)},\B{X}_{J_S},\widebar{\B{X}}_{P}^{(\B{n}_P)},\B{E}_P),
  \end{equation*}
$n_i$ times to turn it into
\begin{equation*}\begin{split}
  X_i &= \iota(X_{[0],i}^{(0)},X_i^{(1)}) \\
            &= \iota(X_{[0],i}^{(0)},\iota(X_{[0],i}^{(1)},X_i^{(2)})) \\
            &= \iota(X_{[0],i}^{(0)},\iota(X_{[0],i}^{(1)},\iota(X_{[0],i}^{(2)},X_i^{(3)}))) \\
            &= \dots \\
            &= \iota(X_{[0],i}^{(0)},\iota(X_{[0],i}^{(1)},\iota(X_{[0],i}^{(2)}, \quad\cdots\quad \iota(X_{[0],i}^{(n_i-1)},X_i^{(n_i)})))) \\
            &= \iota(X_{[0],i}^{(0)},\iota(X_{[0],i}^{(1)},\iota(X_{[0],i}^{(2)}, \quad\cdots\quad \iota(X_{[0],i}^{(n_i-1)}, g_i(\widebar{\B{X}}_{I_S}^{(\B{n}_{I_S}-1)},\B{X}_{J_S},\widebar{\B{X}}_{P}^{(\B{n}_P)},\B{E}_P))))) \\
            &=: F_i(\widebar{X}_{[0],i}^{(n_i-1)}, \widebar{\B{X}}_{I_S}^{(\B{n}_{I_S}-1)},\B{X}_{J_S},\widebar{\B{X}}_{P}^{(\B{n}_P)},\B{E}_P)
\end{split}\end{equation*}
where we explicitly incorporate the initial conditions.
The mapping 
  $F_i : \C{X}_i^{n_i} \times \C{C}^{\B{n}_{I_S\cup J_S\cup P}}(T,\BC{X}_{I_S\cup J_S\cup P}) \times \C{C}^{0}(T,\BC{E}_P) \to \C{C}^{n_i}(T,\C{X}_i)$
defined in this way is continuous (being a composition of continuous mappings), and hence, measurable.
The structural equations for $i \in I_S$ are then taken to be of the form
\begin{equation}\label{eq:SCM_from_SDCM_RDE}
  X_i = F_i(\widebar{X}_{[0],i}^{(n_i-1)}, \widebar{\B{X}}_{I_S}^{(\B{n}_{I_S}-1)},\B{X}_{J_S},\widebar{\B{X}}_{P}^{(\B{n}_P)},\B{E}_P).
\end{equation}
  The structural equations in (\ref{eq:SCM_from_SDCM_AE}) and (\ref{eq:SCM_from_SDCM_RDE}) for all strongly connected components $S\subseteq\C{I}$ of $\C{G}^+_{[0]}$ together specify a well-defined SCM. Its exogenous variables 
  $(\B{E}_j)_{j \in \C{J}}$ and $(\widebar{X}_{[0],i}^{(n_i-1)})_{i\in\C{I}_{[0]}}$ are assumed independent.
  The graph of the SCM is $\C{G}^+_{[0]}$.
  As $\langle \C{R}, I_S, S \rangle$ is assumed to be uniquely solvable, it follows that this SCM is uniquely solvable w.r.t.\ $I_S$.
  As this holds for every strongly connected component $S\subseteq\C{I}$ of $\C{G}^+_{[0]}$,
  the $\sigma$-separation Markov property (see Theorem 6.3.(2) in \citep{BPSM19}) applies, proving the statement.
\end{proof}

\begin{proof}[Proof of Corollary~\ref{cor:evaluated_Markov_property_SDCM}]
  Extend the SCM constructed in the proof of Theorem~\ref{thm:Markov_property_SDCM} with endogenous variables
  $\widebar{X}_{[1],i}^{(n_i)}$ taking values in $\C{X}_i^{n_i+1}$ for $i \in \C{I}_{[1]}$, and with the corresponding structural equations
  $$
    \widebar{X}_{[1],i}^{(n_i)} = \pi(X_i,\partial(X_i), \dots, \partial^{n_i}(X_i)) \,.
  $$
  These functions are continuous, hence measurable, and therefore the extended SCM is well-defined with the evaluated augmented collapsed graph $\C{G}^+_{[0]\dots[1]}(\C{R})$ as its graph. The additional nodes are sink nodes that form their own strongly connected components, and the SCM is obviously also uniquely solvable w.r.t.\ each of these additional nodes. Hence, the $\sigma$-separation Markov property (see Theorem 6.3.(2) in \citep{BPSM19}) holds for this extended SCM with graph $\C{G}^+_{[0]\dots[1]}(\C{R})$.
\end{proof}

\begin{proof}[Proof of Corollary~\ref{cor:transition_Markov_property_SDCM}]
  Observe that the transition graph $\C{G}_{[0]\dots[1]}(\C{R})$ is obtained from the evaluated augmented collapsed graph $\C{G}^+_{[0]\dots[1]}(\C{R})$ by graphically marginalizing out the nodes $\C{I}$. The statement then follows from Lemma~3.3.2 in \citep{FM17}, which states that $\sigma$-separations are preserved under graphical marginalization.
\end{proof}

\begin{proof}[Proof of Proposition~\ref{prop:EquilibriumSolution}]
We show that if $\B{X}$ is an equilibrating solution and $i \in \C{I}$, then $\widebar{X}_i^{(n_i)*}=(X^*_i,0,\dots,0)$ almost surely. For all $0\leq k_i \leq n_i$ we have for almost all $\omega\in\Omega$
$$
\begin{aligned}
  \lim_{t\rightarrow\infty} X_i^{(k_i)}(t,\omega) &= X_i^{(k_i)*}(\omega) \,.
\end{aligned}
$$
Let $0\leq m_i < n_i$. Let $\omega \in \Omega$ such that $\widebar{X}_i^{(n_i)*}(t,\omega)$ converges.
If $X_i^{(m_i+1)*}(\omega) > 0$, then there exists a $\bar{t}\in T$ such
that $X_i^{(m_i+1)}(t,\omega) > \tfrac{1}{2} X_i^{(m_i+1)*}(\omega)$ for $t >
\bar{t}$. From the mean value theorem, it follows that there exists a $c \in (\bar{t},t)$
such that 
$$
    X_i^{(m_i)}(t,\omega) -  X_i^{(m_i)}(\bar{t},\omega) =  
    X_i^{(m_i+1)}(c,\omega) (t-\bar{t})
    > \tfrac{1}{2} X_i^{(m_i+1)*}(\omega) (t-\bar{t})
$$
and hence $X_i^{(m_i)}(t,\omega)$ cannot converge to $X_i^{(m_i)*}(\omega)$.
We get a similar contradiction under the assumption
$X_i^{(m_i+1)*}(\omega) < 0$, and hence $X_i^{(m_i+1)*}(\omega) =
0$. We conclude that $\widebar{X}_i^{(n_i)*}=(X^*_i,0,\dots,0)$ almost surely.
\end{proof}

\begin{proof}[Proof of Proposition~\ref{prop:StabilityResult}]
  We can rewrite the \SDEfull s of $\C{R}$ as
  $$
  \left\{\begin{aligned}
    \B{X}_I' &= -B_{I I'}^{-1}(B_{II}-\mathbb{I}_I)\B{X}_I - B_{I I'}^{-1}B_{IJ}\B{X}_J - B_{I I'}^{-1}\Gamma_{I \C{J}}\B{E}  \\
    \B{X}_J &= -B_{JJ}^{-1}B_{JI}\B{X}_I - B_{JJ}^{-1}\Gamma_{J \C{J}}\B{E} \,.
  \end{aligned} \right.
  $$
  Eliminating $\B{X}_J$ from the right-hand side by substitution yields the RDE
  $$
  \B{X}_I' = A\B{X}_I + C\B{E} \,,
  $$
  where $A:=B_{II'}^{-1}(B_{IJ}B_{JJ}^{-1}B_{JI}-B_{II}+\mathbb{I}_I)$ and $C:=B_{II'}^{-1}(B_{IJ}B_{JJ}^{-1}\Gamma_{J\C{J}} - \Gamma_{I\C{J}})$. The matrix $A$ is a Hurwitz matrix by assumption and thus invertible (note $\det(A)\neq 0$). The solutions of the ODE 
  $\B{x}' = A\B{x} + C \B{e}$, where the vector $\B{e}$ does not depend on time, are of the form
  $\B{x}= \exp(At)\B{x}_0 - A^{-1}C\B{e}$, where $\B{x}_0$ is some vector. For any matrix $A$ there
  exists a nonsingular matrix $P$ (possibly complex) that transforms $A$ into its
  Jordan normal form, that is, $P^{-1}AP = \Lambda$ is a block diagonal matrix where
  each block $\Lambda_i$ is a Jordan block associated with the eigenvalue $\lambda_i$
  of $A$, and is a square matrix of order $m_i$ of the form 
  $$
  \Lambda_i  = \begin{bmatrix} \lambda_i & 1 & 0 & \cdots & \cdots & 0 \\ 
        0 & \lambda_i & 1 & 0 & \cdots & 0 \\ 
        \vdots & & \ddots & \ddots & & \vdots \\
        \vdots & & &\ddots & \ddots & 0 \\
        \vdots & & & & \ddots & 1 \\
        0 & \cdots & \cdots & \cdots & 0 & \lambda_i 
      \end{bmatrix} \,.
  $$
  Therefore,
  $$
    \begin{aligned}
      (\B{X}_I)_t &= \exp(At)\B{X}_{I,[0]} - A^{-1}C\B{E}_t \\
                  &= \exp(P\Lambda P^{-1}t) \B{X}_{I,[0]} - A^{-1}C\B{E}_t \\
            &= \sum_{i=1}^n \sum_{j=1}^{m_i} t^{j-1}\exp(\lambda_i t)
            R_{ij}\B{X}_{I,[0]} - A^{-1}C\B{E}_t
    \end{aligned}
  $$
  with $\B{X}_{I,[0]}$ some random variable, $n$ the total number of block diagonal
  matrices, and the $R_{ij}$'s certain block matrices that depend on $P$ and $\Lambda$ \citep{Kha96}. Since $A$ is a Hurwitz
  matrix by assumption and $\B{E}$ is constant in time, we conclude that for all solutions $\B{X}$ of $\C{R}$,
  $$
  \lim_{t\rightarrow\infty} (\B{X}_I)_t 
  = -A^{-1}C \B{E}
  $$
  and
  $$
  \lim_{t\rightarrow\infty} (\B{X}_J)_t 
  = (B_{JJ}^{-1}B_{JI}A^{-1}C-B_{JJ}^{-1}\Gamma_{J\C{J}}) \B{E}
  $$
  almost surely.

  At last, we consider replacing the condition that the exogenous process $\B{E}$ is constant in time by the assumption that $\B{E}$ may depend on time but is continuous, and that both $\B{E}_t$ and $\exp(At)\int_{t_0}^t \exp(-As)C\B{E}_s ds$ converge almost surely. Observe that the general solutions of $\B{x}' = A\B{x} + C \B{e}$, where we allow $\B{e}$ to be a time-dependent vector, are of the form $\B{x} = \exp(At)\B{x}_0 + \exp(At)\int_{t_0}^t \exp(-As)C\B{E}_s ds$. Then, replacing the term $-A^{-1}C\B{E}_t$ in the equation above for $(\B{X}_I)_t$ by $\exp(At)\int_{t_0}^t \exp(-As)C\B{E}_s ds$ implies also that $(\B{X}_I)_t$ converges a.s., from which the result follows.
\end{proof}

\begin{proof}[Proof of Lemma~\ref{lemm:EquilibratingSolutionsCDM}]
Let $\B{X}$ be an equilibrating solution and let $\B{E}$ converge a.s.\ to the random variable $\B{E}^*$. Then
$$
    \B{X}^* = \lim_{t\rightarrow\infty} \B{X}_t = \lim_{t\rightarrow\infty} \B{f}\big(\widebar{\B{X}}^{(\B{n})}_t,\B{E}_t\big) = \B{f}\Big(\lim_{t\rightarrow\infty}\widebar{\B{X}}^{(\B{n})}_t, \lim_{t\rightarrow\infty} \B{E}_t\Big) 
            = \B{f}(\widebar{\B{X}}^{(\B{n})*},\B{E}^*)  
$$
almost surely, where in the third equality we used the continuity of $\B{f}$.
\end{proof}

\begin{proof}[Proof of Proposition~\ref{prop:ConvergenceOfIndependentProcesses}]
Consider the finite index set $\C{J}=\{1,\dots,e\}$ for some $e\in\NN$. The independence of $(\B{E}_j)_{j\in\C{J}}$ implies that, in particular, for every $t\in T$ the family of random variables $\tilde{\B{E}} := \big((\B{E}_j)_t\big)_{j\in\C{J}}$ is independent, that is, we have $\Prb^{\tilde{\B{E}}_t}=\prod_{j\in\C{J}}\Prb^{(\B{E}_j)_t}$, where $\tilde{\B{E}}_t:=\big( (\B{E}_1)_t,\dots ,(\B{E}_e)_t \big)$. 

Because $\lim_{t\rightarrow\infty}\tilde{\B{E}}_t = \lim_{\substack{n\rightarrow\infty \\ n\in\NN}}\tilde{\B{E}}_n$ a.s., we have $\lim_{\substack{n\rightarrow\infty \\ n\in\NN}}\tilde{\B{E}}_n=\tilde{\B{E}}^*$ a.s., where $\tilde{\B{E}}^*:=(\B{E}^*_1,\dots\B{E}^*_e)$. This implies that $\tilde{\B{E}}_n$ converges in distribution to $\tilde{\B{E}}^*$ (see Remark~6.4 and Corollary~13.19 in \citet{Kle14}), that is, the distribution of $\tilde{\B{E}}_n$ converges weakly to the distribution of $\tilde{\B{E}}^*$, that is, $\text{w-lim}_{n\rightarrow\infty} \Prb^{\tilde{\B{E}}_n} = \Prb^{\tilde{\B{E}}^*}$.\footnote{Let $\Prb, \Prb_1, \Prb_2, \dots$ be probability distributions over $\RN^d$, then \emph{$\Prb_n$ converges weakly to $\Prb$}, denoted by $\text{w-lim}_{n\rightarrow\infty} \Prb_n = \Prb$, if $\lim_{n\rightarrow\infty} \Prb_n(U) = \Prb(U)$ for all measurable sets $U$ in $\RN^d$ with $\Prb(\partial U)=0$, where $\partial U$ is the boundary of $U$, that is, the closure of $U$ minus the interior of $U$.} Similarly, we have $\text{w-lim}_{n\rightarrow\infty} \Prb^{(\B{E}_j)_n} = \Prb^{\B{E}^*_j}$ for every $j\in\C{J}$. Applying Theorem~2.8 in \citet{Bil99} gives that
$$
\Prb^{\tilde{\B{E}}^*} =  \text{w-lim}_{n\rightarrow\infty} \Prb^{\tilde{\B{E}}_n} = \text{w-lim}_{n\rightarrow\infty} \prod_{j\in\C{J}}\Prb^{(\B{E}_j)_n} = \prod_{j\in\C{J}} \Prb^{\B{E}^*_j} \,.
$$
We conclude that the family of random variables $(\B{E}^*_j)_{j\in\C{J}}$ is independent.
\end{proof}

\begin{proof}[Proof of Theorem~\ref{thm:EquilibratingSolutionsSCM}]
  Let $\B{X}$ be an equilibrating solution and let $\B{E}$ converge a.s.\ to the random variable $\B{E}^*$. From Lemma~\ref{lemm:EquilibratingSolutionsCDM} it follows that 
$$
  \B{X}^* = \B{f}(\widebar{\B{X}}^{(\B{n})*},\B{E}^*) = \B{f}(\widebar{\B{\iota}}(\B{X}^*),\B{E}^*) = \B{f}^*(\B{X}^*,\B{E}^*) \quad\text{a.s.,}
$$
where we used in the second equality that $\widebar{\B{\iota}}(\B{X}^*)=\widebar{\B{X}}^{(\B{n})*}$, since for all $i\in\C{I}$ we have that $\widebar{X}_i^{(n_i)*}$ is a.s.\ equal to $(X^*_i,0,\dots,0)$ by Proposition~\ref{prop:EquilibriumSolution}.
\end{proof}

\begin{proof}[Proof of Proposition~\ref{prop:NoEquilibratedInitialConditionNoEquilibriumSolution}]
Suppose that the equilibrated \CDM\ $\C{M}_{\C{R}}$ has a solution $\B{X}^*$. Then the stochastic process $\B{X}:T\times\Omega\to\BC{X}$ defined by $\B{X}_t(\omega):= \B{X}^*(\omega)$ is a solution of $\C{R}$ that equilibrates to $\B{X}^*$.
\end{proof}

\begin{proof}[Proof of Proposition~\ref{prop:EquilibrationSubgraph}]
By definition, the graph of the equilibrated model $\C{M}_{\C{R}}$ has nodes $\C{I}\subseteq\widebar{\C{I}}^{(\B{n})}$ and the augmented graph of $\C{M}_{\C{R}}$ has nodes $\C{I}\cup\C{J}\subseteq\widebar{\C{I}}^{(\B{n})}\cup\C{J}$. For every $i\in\C{I}$, a functional parent of $i$ in $\C{M}_{\C{R}}$ is a functional parent in $\C{R}$, since for all $\B{e}\in\BC{E}$ and for all $\B{x}\in\BC{X}$ we have
$$
x_i = f^*_i(\B{x},\B{e})  \quad\implies\quad x_i = f_i(\widebar{\B{\iota}}(\B{x}),\B{e}) \,.
$$
  Note there are no integrated parents of $i$ in $\C{M}_{\C{R}}$ and there are no functional parents of $j\in\C{J}$.
\end{proof}


\begin{proof}[Proof of Theorem~\ref{thm:InterventionCommutesWithEquilibrationCDMSCM}]
This follows directly from Definitions~\ref{def:CDM}, \ref{def:InterventionsCDM} and \ref{def:EquilibrationCDM}. One can easily check that
$$
\begin{aligned}
  (\C{M}_{\C{R}})_{\intervene(I,\B{K}^*_I)}
  &= 
  \langle \C{I}, I\cup\C{J},\BC{X}, \BC{X}_I\times\BC{E}, \widetilde{\B{f}^*}, (\B{K}_I^*,\B{E}^*) \rangle \\
  &= 
  \langle \C{I}, I\cup\C{J},\BC{X}, \BC{X}_I\times\BC{E}, \tilde{\B{f}}^*, (\B{K}_I,\B{E})^* \rangle \\
  &=
  \C{M}_{\C{R}_{\intervene(I,\B{K}_I)}} \,,
  \end{aligned}
$$
where the intervened and equilibrated dynamic causal mechanism 
$$
  \widetilde{\B{f}^*} = \tilde{{\B{f}}}^*:\BC{X}\times (\BC{X}_I\times\BC{E})\to\BC{X}
$$
is given by
$$
  \tilde{f}_i^*(\B{x},(\B{e}_I,\B{e}_{\C{J}})) := \begin{cases}
    f_i(\widebar{\B{\iota}}(\B{x}),\B{e}_{\C{J}}) \quad & i\in \C{I}\setminus I \\
e_i \quad & i\in I \,.
\end{cases}
$$
\end{proof}

\begin{proof}[Proof of Corollary~\ref{corr:FisherSufficientCondition}]
The statement follows immediately from Theorem 1 of \citet{Fis58}
followed by application of Proposition~\ref{prop:StabilityResult}.

Theorem 1 of \citet{Fis58} states that under the stated condition, there exists an
invertible diagonal stabilization matrix $\Lambda \in \RN^{d\times d}$ such that 
$-\Lambda^{-1} A$ is Hurwitz.\footnote{A simple counterexample of a system that 
cannot be stabilized in this way is given by taking
the matrix 
  $$B = \begin{pmatrix} 0 & 1 & 1 \\ 1 & 0 & 1 \\ 1 & 1 & 0 \end{pmatrix} \,,$$
for which $\Lambda^{-1} (B-\mathbb{I})$ is not Hurwitz for any diagonal invertible matrix $\Lambda$.}

Note first that by construction, $\C{M}_{\C{R}_{\C{M},\Lambda}} = \C{M}$.
  The \CDM\ $\C{R}_{\C{M},\Lambda}$ satisfies Assumption~\asref{ass:ExplicitlySolvable}{(\C{I}\subseteq \C{I})}, that is, it can be written in the form of the equations in Proposition~\ref{prop:StabilityResult} with $I = \C{I}$, where $B_{II'} = -\Lambda$ and $B_{II} = B$, and hence
$B_{II'}^{-1}(-B_{II}+\mathbb{I}_I) = - \Lambda^{-1} A$
is Hurwitz. The statements now follow from Proposition~\ref{prop:StabilityResult}.
\end{proof}

\begin{proof}[Proof of Corollary~\ref{corr:RealizingLinearSCMAsCDM}]
The statement follows from Theorem~2.1 of \citet{LS12} followed by application of Proposition~\ref{prop:StabilityResult}
and Theorems~\ref{thm:EquilibratingSolutionsSCM} and \ref{thm:InterventionCommutesWithEquilibrationCDMSCM}.

Theorem~2.1 of \citet{LS12} states that for every matrix $A\in\RN^{d\times d}$ that satisfies for all subsets $I\subseteq \C{I}$ the condition $\det(A_{II})\det(\mathrm{diag}(A_{II}))>0$, there exists a diagonal matrix $D\in\RN^{d\times d}$ such that the matrix $D_{II} A_{II}$ is Hurwitz for all $I\subseteq\C{I}$. In particular, observe that this matrix $D$ is invertible, since $D_{II}$ is invertible for every $I\subseteq\C{I}$ (note $\det(D_{II})\neq 0$ due to $\det(D_{II} A_{II})\neq 0$). 

Let $\Lambda\in\RN^{d\times d}$ be an invertible diagonal matrix such that $-\Lambda_{II}^{-1}A_{II}$ is Hurwitz for every $I\subseteq\C{I}$.
Note first that by construction, $\C{M}_{\C{R}_{\C{M},\Lambda}} = \C{M}$.
Now let $\intervene(J,\B{K}_J)$ be a stochastic perfect intervention for some subset $J\subseteq\C{I}$ and $\B{K}_J$ some stochastic process that is constant in time. The intervened \CDM\ $(\C{R}_{\C{M},\Lambda})_{\intervene(J,\B{K}_J)}$ satisfies Assumption~\asref{ass:ExplicitlySolvable}{(I\subseteq \C{I})} for $I:=\C{I}\setminus J$, that is, it can be written in the form of the equations in Proposition~\ref{prop:StabilityResult}, where $B_{II'} = -\Lambda_{II}$, $B_{JJ} = -\mathbb{I}_{JJ}$, $B_{JI} =\B{0}_{JI}$ the zero matrix and $\Gamma_{J\C{J}}\B{e} = \B{K}_J$. Moreover,  
$$
B_{II'}^{-1}(B_{IJ}B_{JJ}^{-1}B_{JI}-B_{II}+\mathbb{I}_I) = - \Lambda_{II}^{-1}(\mathbb{I}_I-B_{II}) = - \Lambda_{II}^{-1}A_{II} \,,
$$
which is Hurwitz,
from which we conclude that every solution $\B{X}$ of $(\C{R}_{\C{M},\Lambda})_{\intervene(J,\B{K}_J)}$ is an equilibrating solution. Hence, from Theorem~\ref{thm:EquilibratingSolutionsSCM} it follows that for every solution $\B{X}$ of $(\C{R}_{\C{M},\Lambda})_{\intervene(J,\B{K}_J)}$, its limit $\B{X}^*$ is a solution of the equilibrated model 
$$\C{M}_{((\C{R}_{\C{M},\Lambda})_{\intervene(J,\B{K}_J)})} = (\C{M}_{\C{R}_{\C{M},\Lambda}})_{\intervene(J,\B{K}_J)} = \C{M}_{\intervene(J,\B{K}_J)} \,,$$
  where we made use of Theorem~\ref{thm:InterventionCommutesWithEquilibrationCDMSCM}. Note that $\B{E}$ is assumed constant (in time), and hence $\C{R}_{\C{M},\Lambda}$ is steady; in addition, $\B{K}_J$ is assumed to be constant.
 The solutions of $\C{M}_{\intervene(J,\B{K}_J)}$ are
 a.s.\ unique, because they satisfy the equations $\B{X}^*_I = A_{II}^{-1}(B_{IJ}\B{X}^*_J + \Gamma_{I\C{J}}\B{E})$ and $\B{X}^*_J=\B{K}_J$ almost surely.
\end{proof}

\bibliographystyle{abbrvnat}

\end{document}